\documentclass[conference]{IEEEtran}
\IEEEoverridecommandlockouts
\usepackage{cite}
\usepackage{booktabs}
\usepackage{makecell}
\usepackage{pifont} 
\usepackage{multirow}
\usepackage{tabularx}
\usepackage{graphicx} 
\usepackage{amssymb} 
\usepackage{amsmath,amssymb,amsfonts}
\usepackage{algorithmic}
\usepackage{graphicx}
\usepackage{subcaption}
\usepackage{caption}
\usepackage{textcomp}
\usepackage{xcolor}
\usepackage{soul}
\usepackage[linesnumbered,ruled,vlined]{algorithm2e}
\usepackage{float}
\usepackage{hyperref}
\usepackage{dsfont}
\usepackage{stfloats} 
\usepackage{tikz}
\def\BibTeX{{\rm B\kern-.05em{\sc i\kern-.025em b}\kern-.08em
    T\kern-.1667em\lower.7ex\hbox{E}\kern-.125emX}}

\usepackage{enumitem}
\setlist[itemize]{leftmargin=1em}

\newcommand{\projname}{K\textsuperscript{4}}

\newcommand*\circled[1]{\tikz[baseline=(char.base)]{             \node[shape=circle,fill,inner sep=1pt] (char) {\textcolor{white}{#1}};}}

\title{\projname{}: Online Log Anomaly Detection Via Unsupervised Typicality Learning
\thanks{This work was supported in part by the NSF research grant \#2137603, \#2112606, \#2117439, and \#2320952.}
}

\makeatletter
\renewcommand\IEEEauthorblockN[1]{}%
\renewcommand\IEEEauthorblockA[1]{}%
\makeatother

\author{%
  \begin{minipage}{\textwidth}
    \centering
    \normalsize
    Weicong Chen\textsuperscript{*}, Vikash Singh\textsuperscript{*}, Zahra Rahmani, Debargha Ganguly, Mohsen Hariri, Vipin Chaudhary\\
    Case Western Reserve University, Cleveland, OH, USA\\
    \{weicong, vikash, zahra, debargha, mxh1029, vipin\}@case.edu
  \end{minipage}%
}

\begin{document}

\maketitle

\begingroup
  \renewcommand\thefootnote{\fnsymbol{footnote}}
  \setcounter{footnote}{1}
  \footnotetext{These authors contributed equally to this work.}
\endgroup

\begin{abstract}
Log anomaly detection (LogAD) is crucial for identifying failures and threats in large-scale computing and cyberinfrastructure systems. However, most existing LogAD approaches suffer from key limitations: they depend on slow and error-prone log parsing, employ tightly coupled end-to-end pipelines, often require supervision for improved detection performance, and rely on flawed single-pass evaluation protocols that fail to reflect the temporal dynamics of real-world online detection. These issues significantly hinder scalability, adaptability, and the practical deployment of solutions.

To address these limitations, we introduce \projname{} (\ul{K}nowing the Un\ul{k}nown by \ul{K}nowing only the \ul{K}nown), a fully unsupervised, parser-independent, and representation-agnostic LogAD framework designed for high-performance online detection. At its core, \projname{} is grounded in a novel formulation based on \textit{representation-level typicality estimation}, which transforms arbitrary log embeddings into compact and interpretable four-dimensional descriptors: Precision, Recall, Density, and Coverage (PRDC), which are swiftly computed via GPU-acceleration into geometric $k$-nearest neighbor statistics. These descriptors inform lightweight, modular detectors, including KDE, GMM, OCSVM, and a new adaptation of DeepSVDD, which enables efficient and accurate anomaly scoring without relying on structured formats or log representation retraining.

To support realistic deployment scenarios, we also propose a principled chunk-based evaluation protocol that mimics online log ingestion, alleviates the performance overestimation and dataset undercoverage issues of prior single-pass evaluations, and enables reproducible benchmarking across datasets with varying anomaly densities. Using this setup, we conduct over 125,000 experiments across three real-world datasets (HDFS, BGL, Thunderbird), six pre-trained embedding models, four detectors, and multiple training and log sampling configurations. Compared to six representative baseline methods spanning supervised, semi-supervised, self-supervised, and unsupervised paradigms, \projname{} consistently sets new state-of-the-art results (AUROC: 0.995–0.999, F1: 0.989–0.992) and outperforms all baselines by large margins, while keeping detector training under 4 seconds and per-sample inference latency as low as 4\,$\mu$s, which are orders of magnitude faster than the most competitive alternatives.
\end{abstract}


\section{Introduction}
\label{sec:introduction}

Logs are essential artifacts in computing systems, recording the operational behavior of applications, kernels, and user activities. In large-scale environments such as high-performance computing (HPC) systems, cloud infrastructures, and edge platforms, automated log anomaly detection (LogAD) is vital for prompt error recovery, root-cause diagnosis, and proactive system management~\cite{oliner2007supercomputers,he2016experience, xu2009detecting}. 

With the recent surge in language models and generative AI, a growing body of work~\cite{du2017deeplog, meng2019loganomaly, guo2021logbert, zhang2019robust, jin2024large, lin2024fastlog} has begun leveraging AI techniques to capture semantic patterns in log sequences, aiming to enable more effective LogAD. 
These methods typically follow a structured pipeline consisting of four stages:
(1) Log parsing, where raw logs are converted into structured templates by masking variable fields (e.g., timestamps, node IDs);
(2) Log grouping, where templates are assembled into fixed-size or session-based sequences;
(3) Representation generation, which uses learned or pre-trained encoding models (e.g., LSTM~\cite{du2017deeplog}, FastText~\cite{zhang2019robust}, Transformers~\cite{guo2021logbert, jin2024large}) to encode log sequences into either sequential, quantitative, or semantic embeddings; and
(4) Classification model(s), which model representations and discriminate anomalous logs or log sequences from normal ones through supervised, semi-supervised, or unsupervised objectives.

\vspace{1ex}
\noindent \textbf{Real-World LogAD -- Needs and Gaps:}
Real-world LogAD systems must satisfy several operational requirements. First, they should not merely excel in post hoc analyses of static datasets, but instead deliver accurate detection in \textit{online}, \textit{continuously streaming} log environments. Second, they should not wait for large volumes of historical data to accumulate, but rather support \textit{fast cold-start} using minimal initial logs. Third, they should not require labeled data, but operate in a fully \textit{autonomous} manner. Fourth, they must not impose heavy runtime burdens, but remain \textit{computationally efficient}, as LogAD is expected to be a ubiquitous and persistent task across system modules.

Despite extensive research, existing LogAD approaches fall short of these practical requirements. As illustrated in the dashed enclosure of Figure~\ref{fig:overview}, we identify four critical gaps:

\ul{\textit{(1) Better Detection Only Comes With Supervision:}}
As highlighted in prior studies~\cite{le2022log, jin2024large, yang2021semi}, supervised and semi-supervised LogAD methods consistently outperform self-supervised and unsupervised approaches due to their ability to learn discriminative features from labeled anomaly examples. However, their reliance on labeled data misaligns with real-world deployment needs. Achieving, or even surpassing, supervised-level accuracy without labeled data remains a key challenge for practical, online LogAD.

\ul{\textit{(2) Widespread Reliance on Log Parsers:}} A large body of LogAD studies~\cite{du2017deeplog, meng2019loganomaly, guo2021logbert, zhang2019robust, le2022log, jin2024large, yang2021semi, lin2024fastlog} necessitates structured templates generated by log parsers such as Drain~\cite{he2017drain}, Spell~\cite{du2017spell}, or FT-Tree~\cite{zhang2017syslog}. These tools require global knowledge of log formats and thus typically rely on access to large offline corpora. In online detection environments, parser errors and format drift inject structural noise that degrades downstream detection quality~\cite{le2022log}.

\ul{\textit{(3) Rigid and Inefficient Coupling of Representation Models and Detection Architectures:}}  
Many deep learning-based LogAD systems~\cite{guo2021logbert, du2017deeplog, meng2019loganomaly, zhang2019robust, lin2024fastlog} rigidly couple encoders (e.g., LSTM, BERT) with specific detectors via joint training, limiting modularity and generalization. These pipelines thus force detectors to infer from high-dimensional embeddings, resulting in high computational cost. More recent works~\cite{jin2024large, guan2024logllm} frame detection as supervised classification with large decoder-only LLMs, further increasing inference overhead and abandoning modular detection altogether.

\ul{\textit{(4) Flawed Single-pass Evaluation Protocols:}}
Most LogAD studies evaluate public log datasets in a single pass, where each dataset typically contains millions to hundreds of millions of log lines~\cite{zhu2023loghub}. Some studies~\cite{meng2019loganomaly, zhang2019robust, le2022log, yang2021semi, jin2024large} use the majority portion of the dataset (up to 80\%) for training, which contradicts the objective of fast cold-start and leads to overly optimistic estimates of detection performance. Others~\cite{du2017deeplog, guo2021logbert, lin2024fastlog} use a smaller chronological fraction (1–10\%), which, despite still involving a substantial volume of logs, fails to capture the inherent variability within each dataset.

\vspace{1ex}
\noindent \textbf{Contributions:}
Motivated by these significant gaps in existing LogAD solutions, we introduce \textit{\textbf{\projname{} (\ul{K}nowing the Un\ul{k}nown by \ul{K}nowing only the \ul{K}nown)}}, a novel framework that offers high-performance, modular, and robust anomaly detection that meets stringent real-world deployment needs. As shown following the red arrows in Figure~\ref{fig:overview}, our contributions include:

\begin{figure*}[htbp]
    \centering
    \includegraphics[width=\textwidth]{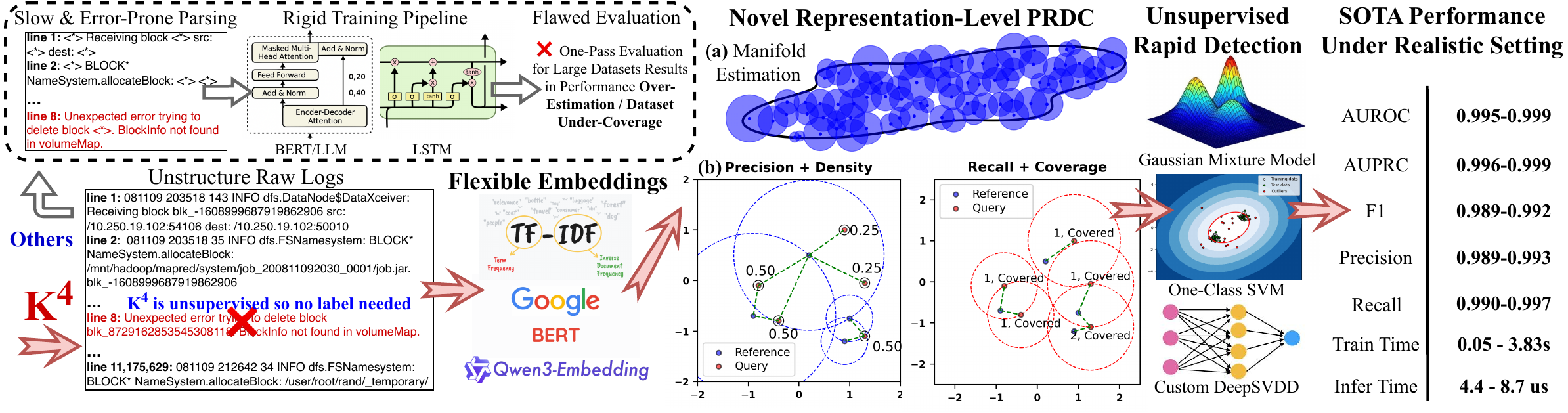}
    \caption{\projname{}'s Novelty, Architecture, and Performance Highlights. \small{\projname{} addresses key limitations in prior LogAD systems: supervision vs. performance trade-offs, parser reliance, rigid pipelines, and unrealistic evaluation. It introduces a novel representation-level anomaly scoring framework using geometric typicality statistics (PRDC), computed via kNN manifold estimation (Detailed in Section~\ref{sec:methodology}). The modular design enables flexible embeddings (e.g., TF-IDF, Qwen3-Embedding) and lightweight detectors (e.g., GMM, DeepSVDD), achieving real-time performance and SOTA accuracy across HDFS, BGL, and Thunderbird under a realistic chunk and sliding-window-based evaluation protocol (Detailed in Section~\ref{ss:experimental_setup}).}}
    \vspace{-2ex}
    \label{fig:overview}
\end{figure*}

\vspace{1ex}
\noindent \circled{1} \textbf{Unsupervised LogAD via Representation-Level Typicality Estimation:}
We propose a transformative LogAD formulation based on \textit{representation-level typicality estimation}~\cite{ganguly2025forte}, which maps arbitrary log embeddings into compact and interpretable four-dimensional statistical descriptors: precision, recall, density, and coverage (PRDC). These metrics are computed through geometric analysis over local $k$-nearest neighbor (kNN) manifolds constructed from normal-only training data, with each PRDC dimension encoding a distinct geometric intuition about a sample’s “typicality” within the normal representation space. This new PRDC representation enables efficient, unsupervised, and format-agnostic anomaly scoring. To the best of our knowledge, \projname{} is the first to introduce such manifold-derived typicality features in the LogAD domain, offering a generalizable and interpretable foundation for robust and scalable detection across diverse log sources.

\vspace{1ex}
\noindent \circled{2} \textbf{Modular, Real-Time, and High-Performance Pipeline:}
Built on PRDC features, \projname{} cleanly separates representation models and detection architectures, allowing seamless pairing of six off-the-shelf embedding models (TF-IDF~\cite{sparck1972statistical}, Word2Vec~\cite{mikolov2013efficient}, SBERT~\cite{reimers2019sentence}, and three Qwen3 variants~\cite{qwen3embedding}) with four efficient detectors: Gaussian Mixture Models (GMM)~\cite{dempster1977maximum}, Kernel Density Estimation (KDE)~\cite{rosenblatt1956kde}, One-Class SVM (OCSVM)~\cite{scholkopf2001estimating}, and our newly adapted Deep Support Vector Data Description (DeepSVDD)~\cite{ruff2018deep}. This modular design supports rapid adaptation to evolving LogAD task dynamics by enabling effortless integration of new embedding or detection components. The compactness of PRDC representations significantly reduces computational overhead, allowing \projname{} to achieve real-time performance with inference latency as low as 4\,$\mu$s per test sample and detector training times as low as 50\,ms. These capabilities, along with parser-free design, make \projname{} highly scalable and well-suited for continuous online deployment.

\vspace{1ex}
\noindent \circled{3} \textbf{Realistic Evaluation Protocol for Online LogAD.}
We introduce a chunk-based evaluation protocol that partitions each large-scale log dataset into disjoint, fixed-size chunks. Within each chunk, we apply sliding window sampling with configurable strides to emulate realistic log arrival while ensuring semantic preservation through overlapping context. This setup enables comprehensive coverage of the dataset, balances training and testing across varying anomaly densities, and corrects the detection overestimation and dataset undercoverage dilemma inherent in prior single-pass evaluations. Our protocol establishes a faithful and reproducible benchmark for online log anomaly detection.

\vspace{1ex}
\noindent \circled{4} \textbf{Comprehensive Evaluation and New State-of-the-Art Results:}
We conduct an extensive evaluation of \projname{} involving $120,912$ experiments across variations in embeddings, detectors, window sizes, strides, and training sample sizes. In parallel, we benchmark $4,122$ experiments covering six prior representative LogAD studies, co-designed under our proposed evaluation protocol. All experiments are executed across three real-world public datasets~\cite{zhu2023loghub} using two HPC clusters equipped with 40 NVIDIA A100 and 40 NVIDIA H200 GPUs. The results show that \projname{} consistently outperforms leading supervised, semi-supervised, self-supervised, and unsupervised approaches. Our best configuration achieves AUROC/F1 scores of $0.996$/$0.989$ (HDFS), $0.999$/$0.992$ (BGL), and $0.999$/$0.992$ (Thunderbird), establishing a new and challenging-to-match standard for practical, scalable, and accurate log anomaly detection.

\section{Background and Related Work}
\label{sec:background_relatedwork}

\subsection{Original Formulation of  Typicality Estimation}
\label{ss:background_typicality_estimation}


Typicality estimation offers a principled way to detect out-of-distribution (OOD) inputs by assessing how well samples conform to the known data distribution, using typical sets from information theory~\cite{nalisnick2019detecting, morningstar2021density}.
In parallel, Precision, Recall, Density, and Coverage (PRDC) metrics have been proposed for evaluating image-domain generative models by comparing distributions of generated and real samples. Originating from the distribution-level precision and recall formulation by Kynkäänniemi et al.\cite{kynkaanniemi2019improved} and extended with density and coverage by Naeem et al.\cite{naeem2020reliable}, PRDC frames quality estimation as a set-to-set comparison between reference samples $X$ and query samples $Q$, using $k$-nearest neighbor (kNN) radii to estimate set-level geometric manifold (visualized in Figure~\ref{fig:overview}(a) with a 2D space and $k$ = 1) and further derive global geometric statistics as follows:

\begin{itemize}[leftmargin=*, itemsep=0.3ex]
    \item \textbf{Precision:} The fraction of query samples $q \in Q$ that lie within the $k$-NN neighborhoods of any reference point $x \in X$. This metric assesses the extent to which the query set falls into high-density regions of the reference distribution.

    \item \textbf{Recall:} The fraction of reference samples $x \in X$ that are contained within the $k$-NN neighborhoods of any query sample $q \in Q$. This reflects the degree to which the query set covers the support of the reference distribution.

    \item \textbf{Density:} The normalized count of reference points $x \in X$ whose $k$-NN neighborhood contains a given query point $q \in Q$, serving as a proxy for the local likelihood of $q$ under the reference distribution.

    \item \textbf{Coverage:} The proportion of reference samples $x \in X$ whose nearest query point lies within their own $k$-NN radius. This statistic indicates how well the query distribution covers the local neighborhoods of the reference distribution.
\end{itemize}

While PRDC was originally designed for evaluating generative models, its geometric formulation naturally lends itself to LogAD by quantifying deviations from the known data distribution. However, prior formulations focus on comparing entire distributions in aggregate, making them unsuitable for assessing the conformity of individual samples. This instance-level capability is essential in online LogAD, where streaming logs must be evaluated sample by sample to detect localized anomalies as (or even before) they occur. We address this gap in Section~\ref{ss:representation_level_statistics} by adapting PRDC for pointwise statistics over embedding representations.

\subsection{Related Work}
\label{ss:related-work}

Table~\ref{tab:framework-comparison} summarizes representative related LogAD studies to be compared with \projname{} and highlights key distinctions from our approach.

\renewcommand{\arraystretch}{1.5}
\begin{table}[htbp]
\centering
\caption{Comparison of recent LogAD frameworks across key design dimensions. 
\projname{} is unique in being parser-free, embedding-agnostic, and fully unsupervised, while supporting efficient anomaly modeling through statistical detectors.}
\label{tab:framework-comparison}
\scalebox{0.78}{%
\begin{tabular}{|l|c|c|c|c|}
\hline
\textbf{\normalsize Framework} & \textbf{\normalsize Parser} & \textbf{\normalsize Representation} & \textbf{\normalsize Architecture} & \textbf{\normalsize Paradigm} \\
\hline
DeepLog~\cite{du2017deeplog} & Spell~\cite{du2017spell} & Template ID & LSTM & Semi-supervised \\
\hline
LogAnomaly~\cite{meng2019loganomaly} & FT-Tree~\cite{zhang2017syslog} & Template2Vec & LSTM & Unsupervised \\
\hline
LogRobust~\cite{zhang2019robust} & Drain~\cite{he2017drain} & FastText~\cite{joulin2016fasttext} & Bi-LSTM & Supervised \\
\hline
FastLogAD~\cite{lin2024fastlog} & Drain & BERT & Transformer & Self-supervised \\
\hline
LLM-AD~\cite{jin2024large} & Custom & N/A & Transformer & Supervised \\
\hline
Ladle~\cite{myllari2025ladle} & Custom & Sentence BERT & KNN & Unsupervised \\
\hline
\textbf{\projname{} (Ours)} & None & Any→PRDC & \makecell[l]{GMM, OCSVM,\\ KDE, DeepSVDD} & Unsupervised \\
\hline
\end{tabular}}  
\end{table}

\textit{DeepLog}\cite{du2017deeplog} adopts a semi-supervised paradigm by training an LSTM~\cite{graves2012long} to predict the next event in normal sequences, flagging anomalies when actual events fall outside the top-$k$ predictions.
\textit{LogAnomaly}\cite{meng2019loganomaly} employs unsupervised dual LSTM encoders to jointly model log templates and quantitative features using FT-Tree parser~\cite{zhang2017syslog} and in-house Template2Vec embeddings, followed by predicting the next log and its attributes, with high predictive errors signaling anomalies.
\textit{LogRobust} enhances robustness to parser noise by applying a supervised Bi-LSTM classifier to FastText~\cite{joulin2016fasttext} embeddings of Drain-parsed~\cite{he2017drain} logs. 
\textit{FastLogAD}~\cite{lin2024fastlog} is a self-supervised framework that employs a masked language model generator to produce pseudo-anomalies from normal logs. It then trains a LogBERT-based~\cite{guo2021logbert} discriminator to distinguish between normal and corrupted logs. 
Jin et al. \cite{jin2024large} fine-tune large language models (LLMs) and leverage in-context learning for supervised classification (we abbreviate as LLM-AD). 
\textit{Ladle}~\cite{myllari2025ladle} performs unsupervised LogAD by computing kNN distances between sentence-BERT embeddings of test log windows and a reference set of normal training windows. Anomaly scores are derived from the average distance to the $k$ nearest neighbors.

\section{Methodology of \projname{}}
\label{sec:methodology}

Since Figure~\ref{fig:overview} has clearly demonstrated \projname{}'s LogAD pipeline, this section focuses on articulating its novel unsupervised learning methodology, which, for the first time, incorporates typicality learning in the LogAD domain.

\subsection{Representation-Level Typicality Estimation}
\label{ss:representation_level_statistics}

As discussed in Section~\ref{ss:background_typicality_estimation}, the original typicality estimation formulation operates at the distribution level and is unsuitable for instance-level anomaly detection. To overcome this limitation, we leverage the \textit{per-point} PRDC formulation inspired by \textsc{Forte}~\cite{ganguly2025forte}, adapted to text-based log data to compute geometric statistics independently for each test representation with respect to a reference set comprising only normal samples.

Formally, let $E_{\text{ref}} = \{x_i^{(r)}\}_{i=1}^n \subset \mathbb{R}^d$ be a reference set of embeddings (generated from normal logs), and let $E_{\text{query}} = \{x_j^{(g)}\}_{j=1}^m \subset \mathbb{R}^d$ be a set of query embeddings (from held-out normal or test logs). Define $B(x, r)$ as a Euclidean ball centered at $x$ with radius $r$, and let $\mathrm{NND}_k(x)$ denote the Euclidean distance from $x$ to its $k$-th nearest neighbor in its own set, excluding itself. 
Let $\mathds{1}(\cdot)$ denote the indicator function that returns $1$ if its argument is true and $0$ otherwise. The PRDC statistics for a query point $x_j^{(g)}$ are then defined as follows:

\begin{itemize}[leftmargin=*, itemsep=0.5ex]
    \item \textbf{Precision Per Point ($P_j$):} Indicates whether the query lies within the $k$-NN ball of any reference point.
    \[
    P_j = \mathds{1} \left[ x_j^{(g)} \in \bigcup_{i=1}^{n} B(x_i^{(r)}, \mathrm{NND}_k(x_i^{(r)})) \right]
    \]

    \item \textbf{Recall Per Point ($R_j$):} Measures the fraction of reference points that lie within the $k$-NN ball around the query.
    \[
    R_j = \frac{1}{n} \sum_{i=1}^{n} \mathds{1} \left[ x_i^{(r)} \in B(x_j^{(g)}, \mathrm{NND}_k(x_j^{(g)})) \right]
    \]

    \item \textbf{Density Per Point ($D_j$):} Measures how frequently the query lies inside the neighborhoods of reference points, normalized by $k$ and the number of reference samples. It captures the local data density surrounding the query.
    \[
    D_j = \frac{1}{k n} \sum_{i=1}^{n} \mathds{1} \left[ x_j^{(g)} \in B(x_i^{(r)}, \mathrm{NND}_k(x_i^{(r)})) \right]
    \]

    \item \textbf{Coverage Per Point ($C_j$):} Indicates whether the closest reference point to the query lies within the query’s $k$-NN ball. It reflects the support coverage of the normal data.
    \[
    C_j = \mathds{1} \left[ \min_{i} \| x_j^{(g)} - x_i^{(r)} \| < \mathrm{NND}_k(x_j^{(g)}) \right]
    \]
\end{itemize}

Each query sample $x_j^{(g)}$ is thereby mapped into a four-dimensional vector: 

\[
\mathbf{r}_j = [P_j, R_j, D_j, C_j]^\top \in \mathbb{R}^4
\]


Figure~\ref{fig:overview}(b) visualizes PRDC in 2D with $k$ = 1. In Precision + Density, a query point (red) satisfies precision if it lies within at least one reference (blue) 1-NN ball, marked by a black outline. Green dashed lines connect each query to the reference points whose radii contain it, and the adjacent number indicates density. In Recall + Coverage, each green dashed line shows a reference lying within a query’s 1-NN ball. The text of each query indicates the recall value and whether coverage is satisfied.

The algorithmic implementation of representation-level PRDC is described in Algorithm~\ref{alg:prdc}. It leverages GPU-accelerated distance computation and $k$-NN search using \texttt{torch.cdist} and \texttt{torch.topk}, enabling efficient PRDC computation at scale.

\begin{algorithm}[htbp]
\small
\caption{Representation-Level PRDC}
\label{alg:prdc}
\KwIn{Reference embeddings $E_{\text{ref}} \in \mathbb{R}^{n \times d}$, query embeddings $E_{\text{query}} \in \mathbb{R}^{m \times d}$, number of neighbors $k$}
\KwOut{PRDC matrix $V \in \mathbb{R}^{m \times 4}$}

$d_{\text{ref}} \gets \texttt{cdist}(E_{\text{ref}}, E_{\text{ref}})$ \hfill\# intra-ref distances

$d_{\text{query}} \gets \texttt{cdist}(E_{\text{query}}, E_{\text{query}})$ \hfill\# intra-query distances

$d_{\text{cross}} \gets \texttt{cdist}(E_{\text{query}}, E_{\text{ref}})$ \hfill\# query-to-ref distances

$r_{\text{ref}} \gets \texttt{topk}(d_{\text{ref}}, k)[0][:, -1]$ \hfill\# $k$-NN radius for each reference point

$r_{\text{query}} \gets \texttt{topk}(d_{\text{query}}, k)[0][:, -1]$ \hfill\# $k$-NN radius for each query point

\For{$j \leftarrow 1$ \KwTo $m$}{
    $x_j^{(g)} \gets E_{\text{query}}[j]$

    $P_j \gets \mathds{1}\left[ \exists i: d_{\text{cross}}[j, i] < r_{\text{ref}}[i] \right]$ \hfill\# Precision

    $R_j \gets \frac{1}{n} \sum_{i=1}^n \mathds{1} \left[ d_{\text{cross}}[j, i] < r_{\text{query}}[j] \right]$ \hfill\# Recall

    $D_j \gets \frac{1}{k n} \sum_{i=1}^n \mathds{1} \left[ d_{\text{cross}}[j, i] < r_{\text{ref}}[i] \right]$ \hfill\# Density

    $C_j \gets \mathds{1} \left[ \min_i d_{\text{cross}}[j, i] < r_{\text{query}}[j] \right]$ \hfill\# Coverage
}

\Return $V = \{(P_j, R_j, D_j, C_j)\}_{j=1}^m$
\end{algorithm}

This representation-level PRDC formulation serves as the foundation of \projname{}'s unsupervised LogAD methodology by offering several advantages:
\textbf{(1)} It computes fine-grained, per-sample measures of conformity to the normal data distribution, making it well-suited for real-time and streaming log anomaly detection where localized decisions are essential;
\textbf{(2)} It decouples the detection process from any specific embedding model by operating on geometric relationships rather than raw embeddings. This streamlines cross-model integration;
and \textbf{(3)} It distills high-dimensional embeddings into compact 4-dimensional feature vectors that retain essential distributional information, enabling efficient and interpretable anomaly detection downstream.

\subsection{Unsupervised Typicality Learning}
\label{ss:unsupervised_typicality_learning}

Building on the reference-level PRDC formulation, we now describe how these per-sample vectors $\mathbf{r} \in \mathbb{R}^4$ enable unsupervised anomaly modeling. In \projname{}, we cast the problem as learning the statistical ``normality'' of PRDC vectors derived purely from normal log's embeddings.

\vspace{1ex}
\noindent\textbf{Unsupervised Training:}
As outlined in Lines 1 to 3 of Algorithm~\ref{alg:workflow}, we randomly divide the set of normal log embeddings into two equal subsets: a reference set ${E_{\text{ref}} = x_i^{(r)}}{i=1}^{N/2}$ and a query set ${E_{\text{qry}} = x_j^{(q)}}_{j=1}^{N/2}$. Each query embedding $x_j^{(q)}$ is then converted into a four-dimensional PRDC vector $\mathbf{r}_j \in R_{\text{train}}$ using Algorithm~\ref{alg:prdc}. These PRDC descriptors quantify how typical each query embedding is with respect to the normal data manifold, based on geometric relationships within the reference set. The resulting PRDC vectors serve as training inputs for the downstream anomaly detector $\mathcal M$.

\begin{algorithm}[htbp]
\small
\caption{Unsup. Typicality Learning \& Inferencing}
\label{alg:workflow}
\KwIn{Normal log embeddings $E_{\mathrm{train}}\in\mathbb R^{N\times d}$, test log embeddings $E_{\mathrm{test}}\in\mathbb R^{T\times d}$, number of neighbors $k$, detector $\mathcal M$}
\KwOut{Trained detector $\mathcal M$, anomaly scores $\mathbf s\in\mathbb R^T$}

$(E_{\mathrm{ref}}, E_{\mathrm{qry}}) \gets \mathrm{random\_split}(E_{\mathrm{train}}, 0.5)$ \hfill\# reference vs.\ query  

$R_{\mathrm{train}} \gets \text{PRDC}(E_{\mathrm{ref}}, E_{\mathrm{qry}}, k)$ \hfill \# cf.\ Algorithm~\ref{alg:prdc}

$\mathcal M.\text{fit}(R_{\mathrm{train}})$

$R_{\mathrm{test}} \gets \text{PRDC}(E_{\mathrm{ref}}, E_{\mathrm{test}}, k)$ \hfill \# reuse reference

$\mathbf s \gets \mathcal M.\text{score}(R_{\mathrm{test}})$

\Return $\mathcal M,\,\mathbf s$
\end{algorithm}

\vspace{1ex}
\noindent\textbf{Inference:}
In Lines 4 to 6 of Algorithm~\ref{alg:workflow}, each test embedding is evaluated by computing its PRDC vector with respect to the same reference set $E_{\text{ref}}$ used during training. This transformation quantifies how closely the test sample conforms to the normal representation manifold. The resulting PRDC vector is then passed to the anomaly detector $\mathcal M.\texttt{score()}$, which assigns an anomaly score, where higher values indicate greater likelihood of abnormality.

\vspace{1ex}
\noindent\textbf{Choice of $\mathcal M$:}
We support four unsupervised detectors. \textbf{GMM}~\cite{dempster1977maximum} models the training distribution as a weighted sum of Gaussian components, assigning anomaly scores based on log-likelihood under the fitted mixture. \textbf{KDE}~\cite{rosenblatt1956kde} is a non-parametric estimator that computes local data density by averaging kernel responses from nearby points. \textbf{OCSVM}~\cite{scholkopf2001estimating} learns a high-dimensional decision boundary that maximally separates normal instances from the origin, treating outliers as support vectors. \textbf{DeepSVDD}~\cite{ruff2018deep} maps inputs into a latent space via a neural network and minimizes the radius of a hypersphere that encloses the mapped representations, with anomalies lying outside the learned boundary.

GMM, KDE, and OCSVM are implemented using both Scikit-Learn (CPU-only) and custom GPU-accelerated PyTorch variants (\texttt{TorchGMM}, \texttt{TorchKDE}, \texttt{TorchOCSVM}), which offer 4-10x training and inference speedups to our experiments.

DeepSVDD is newly integrated beyond the detector suite employed in \textsc{Forte}~\cite{ganguly2025forte}. We adapt the original DeepSVDD by replacing the joint optimization of the hypersphere radius R with a non-parametric (1-$\nu$)-quantile update after each epoch. This decouples the radius from the loss gradient, stabilizing training on low-dimensional PRDC features. To further improve generalization in shallow architectures, we incorporate BatchNorm and Dropout layers. These modifications result in more reliable anomaly boundaries for compact representations. 

\section{Evaluation}
\label{sec:evaluation}

\subsection{Experimental Setup For Realistic Online LogAD}
\label{ss:experimental_setup}

\noindent\textbf{Baselines:}
As discussed in Section~\ref{ss:related-work}, we conduct fair, consistent, and extensive experiments on six representative baselines spanning statistical, deep-neural, and language-model-based LogAD studies: 
DeepLog~\cite{du2017deeplog}, 
LogAnomaly~\cite{meng2019loganomaly}, 
LogRobust~\cite{zhang2019robust}, 
FastLogAD~\cite{lin2024fastlog},
Ladle~\cite{myllari2025ladle},
and LLM-AD~\cite{jin2024large}. Note that some of these baselines have included comparisons with earlier prominent studies like LogBERT~\cite{guo2021logbert} and PLELog~\cite{yang2021semi}, we thus omitted evaluating them.

\vspace{1ex}
\noindent\textbf{Datasets:}
We evaluate \projname{} and related works on three publicly available datasets: \textit{HDFS}, \textit{BGL}, and \textit{Thunderbird}. These datasets~\cite{zhu2023loghub} are widely adopted in related work and originate from large-scale distributed and supercomputing systems.

\begin{itemize}
  \item \textit{HDFS (Hadoop Distributed File System)}: Collected from over 200 Amazon EC2 nodes, this dataset contains approximately 11.2 million log messages. Logs are grouped by \texttt{block\_id} into 575,061 blocks, with a total of 16,838 blocks (2.93\%) are labeled as anomalous.
  
  \item \textit{BGL (Blue Gene/L)}: Sourced from the Blue Gene/L supercomputer at Lawrence Livermore National Laboratory (LLNL), this dataset includes 4.75 million system log messages, with 348,460 (7.34\%) labeled as anomalous.
  
  \item \textit{Thunderbird}: Collected from the Thunderbird supercomputer at Sandia National Laboratories (SNL), this dataset contains over 211 million log messages, including approximately 3.24 million anomalous entries (1.54\%).
\end{itemize}

\vspace{1ex}
\noindent\textbf{Dataset Coverage:}
As discussed in ``Needs and Gaps'' of Section~\ref{sec:introduction}, prior LogAD studies typically evaluate public log datasets using a single experiment. This practice fails to reflect real-world online LogAD scenarios, as it will result in either excessive training data, limited coverage of dataset variability, or both, depending on the train/test splitting ratios.

To address this flawed experimental practice, we propose a more realistic evaluation protocol by partitioning each dataset into disjoint chunks and conducting one detection experiment per chunk. For this paper, each chunk contains one million logs, which is empirically chosen to ensure a sufficient and sensible number of training samples, normal test samples, and anomaly test samples across the evaluated datasets. As a result, \projname{} performs $12$, $5$, and $212$ experiments on the HDFS, BGL, and Thunderbird datasets, respectively. All baseline frameworks are adapted to our chunk-based protocol for fair and consistent comparisons.

\vspace{1ex}
\noindent\textbf{Log Sampling and Evaluation Dimensions:}
Within each dataset chunk, we apply a sliding window to segment logs into fixed-length samples. A sample is labeled anomalous if it contains at least one anomalous log; otherwise, it is labeled normal. The window size controls temporal context, while the stride determines sampling density. We perform a grid search over window sizes ${40, 80, 160, 320}$ and strides ${5, 10, 20, 40}$ to identify effective configurations. With the derived configuration, we further vary the training sample sizes from ${500, 1000, 2000, 5000, 10000, 20000}$ to evaluate \projname{}'s robustness under different log data availability conditions. This enables us to assess the trade-offs between accuracy and data efficiency. Baselines are evaluated with the same selected window size and stride, using ${500, 5000, 20000}$ training samples for consistency.

\vspace{1ex}
\noindent\textbf{Embeddings:}
We evaluate six pre-trained embeddings spanning classical statistical, shallow neural, and large language model (LLM)-based approaches. \texttt{TF-IDF}~\cite{sparck1972statistical} (term frequency-inverse document frequency) is a sparse, count-based representation that encodes the relative importance of tokens within documents. \texttt{Word2Vec}~\cite{mikolov2013efficient} is a shallow, context-based neural model that learns token embeddings by predicting neighboring words via skip-gram. SBERT~\cite{reimers2019sentence} produces dense vector representations for sentences by fine-tuning a Siamese network on top of BERT. We specifically use the lightweight variant \texttt{all-MiniLM-L6-v2}, which balances performance and efficiency.
\texttt{Qwen3-Embedding}~\cite{qwen3embedding} are multilingual, instruction-tuned encoder-only models tailored for downstream embedding and ranking tasks. We employ three Qwen3-Embedding variants: 0.6B, 4B, and 8B parameters. Note that in our evaluation, the 0.6B and 4B variants perform very similarly to the 8B model across all datasets and detector configurations. For clarity, we report only results from Qwen3-Embedding-8B.

Because SBERT and Qwen3 have context length limits of 512 and 32k tokens, respectively, and a large window sample can potentially exceed these limits, we thus apply a token length-aware embedding strategy: Each log window is split into multiple consecutive shards, where each shard fits within the model’s token budget by packing in as many log lines as possible without exceeding the limit. Each shard is embedded independently using the target model, and the final window representation is obtained by mean pooling of the embeddings across all shards.

\vspace{1ex}
\noindent\textbf{Experiment Counts and Testbeds:}
Using our proposed online LogAD experimental and evaluation setup, we conduct $(12~\text{HDFS}$ $+ 5~\text{BGL} + 212~\text{Thunderbird}) \times 6~\text{embeddings} \times 4~\text{window sizes} \times 4~\text{strides} \times 4~\text{detectors} = 87,936$ experiments to study \projname{}’s performance trends across temporal configurations. We further run $[12 + 5 + 212] \times 6 \times 4 \times 6 = 32,976$ experiments to evaluate the impact of training sample size. In total, we execute $120,912$ experiments for \projname{}, and $[12 + 5 + 212] \times 6~\text{baselines} \times 3~\text{train sizes} = 4,122$ experiments for baseline comparisons.

To facilitate this intricate experiment dimension, Table~\ref{tab:testbeds} depicts the clusters used in this study. 

\begin{table}[htbp]
\footnotesize
\caption{Specifications of Two HPC Clusters}
\label{tab:testbeds}
\centering
\renewcommand{\arraystretch}{0.95} 
\setlength{\tabcolsep}{5pt}        
\begin{tabular}{@{}lcc@{}}
    \toprule
    \textbf{Cluster} & \textbf{CWRU AISC} & \textbf{CWRU AISC II} \\
    \midrule
    \textbf{Processor}      & AMD EPYC 7742 $\times$ 2          & Intel Xeon Platinum 8468 $\times$ 2 \\
    \textbf{RAM}      & 2048\,GB            & 2048\,GB \\
    \textbf{GPU}            & NVIDIA A100 $\times$ 8            & NVIDIA H200 $\times$ 8 \\
    \textbf{VRAM}            & 80\,GB HBM2e             & 141\,GB HBM3e \\
    \textbf{Scale}          & 5 nodes (40 A100)       & 5 nodes (40 H200) \\
    \bottomrule
\end{tabular}
\end{table}

\noindent\textbf{Detection Performance Metrics:}
We employ a suite of metrics to evaluate the quality of anomaly detection. \textit{AUROC} (Area Under the Receiver Operating Characteristic Curve) captures the trade-off between true positive and false positive rates across varying thresholds, offering a threshold-independent measure of how well a model discriminates normal vs. anomaly. \textit{AUPRC} (Area Under the Precision-Recall Curve) is particularly informative for typical log datasets where anomalies are rare, as it emphasizes the trade-off between precision and recall. We also report traditional threshold-dependent metrics: \textit{F1}, \textit{Precision}, and \textit{Recall}, with their values based on the decision boundary that maximizes the F1 score. Additionally, we report \textit{FPR@95TPR}, the false positive rate at 95\% recall, to capture practical utility in LogAD, where detecting most anomalies is critical, but excessive false alarms can overwhelm operators. Among these, we prioritize \textbf{AUROC} as our primary metric, as it provides a fair, threshold-agnostic basis for comparing detection quality.

\vspace{1ex}
\noindent\textbf{Runtime Performance Metrics:}
Purposed as an online LogAD solution, we also evaluate \projname{}'s runtime efficiency along two axes: (1) \textit{Training times}, defined as the time required to generate embeddings for all training samples, and the time for training an anomaly detector with the generated embeddings, and (2) \textit{Inference times}, which includes the embedding generation time for one incoming test sample and the time for a detector to score the embedding of this sample.

\subsection{Evaluation on \projname{}'s Detection Capability}
\label{sec:eval:stride-window-sample}

\begin{figure*}[h]
    \centering
    \subfloat[HDFS - TF-IDF]{\includegraphics[width=0.24\textwidth]{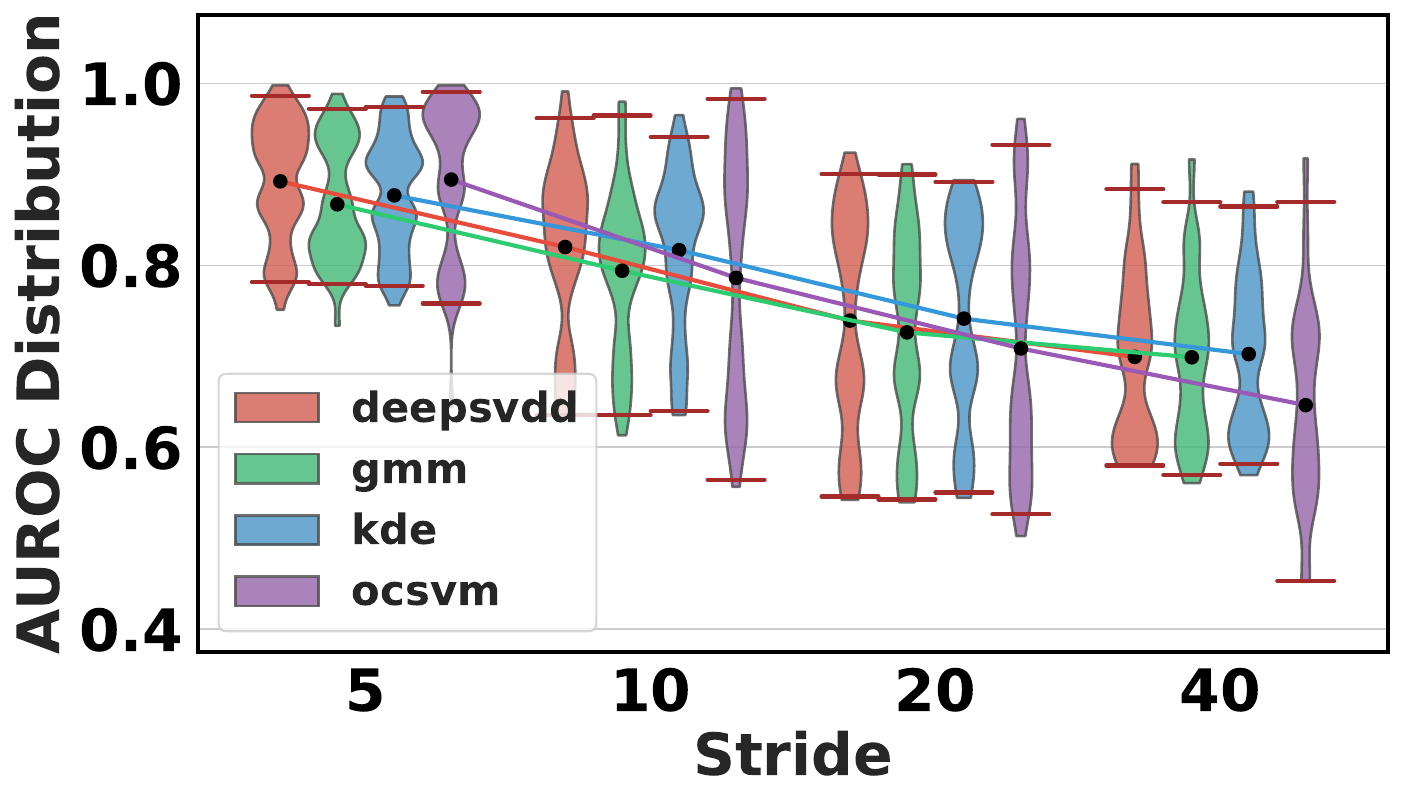}}
    \subfloat[HDFS - Word2Vec]{\includegraphics[width=0.24\textwidth]{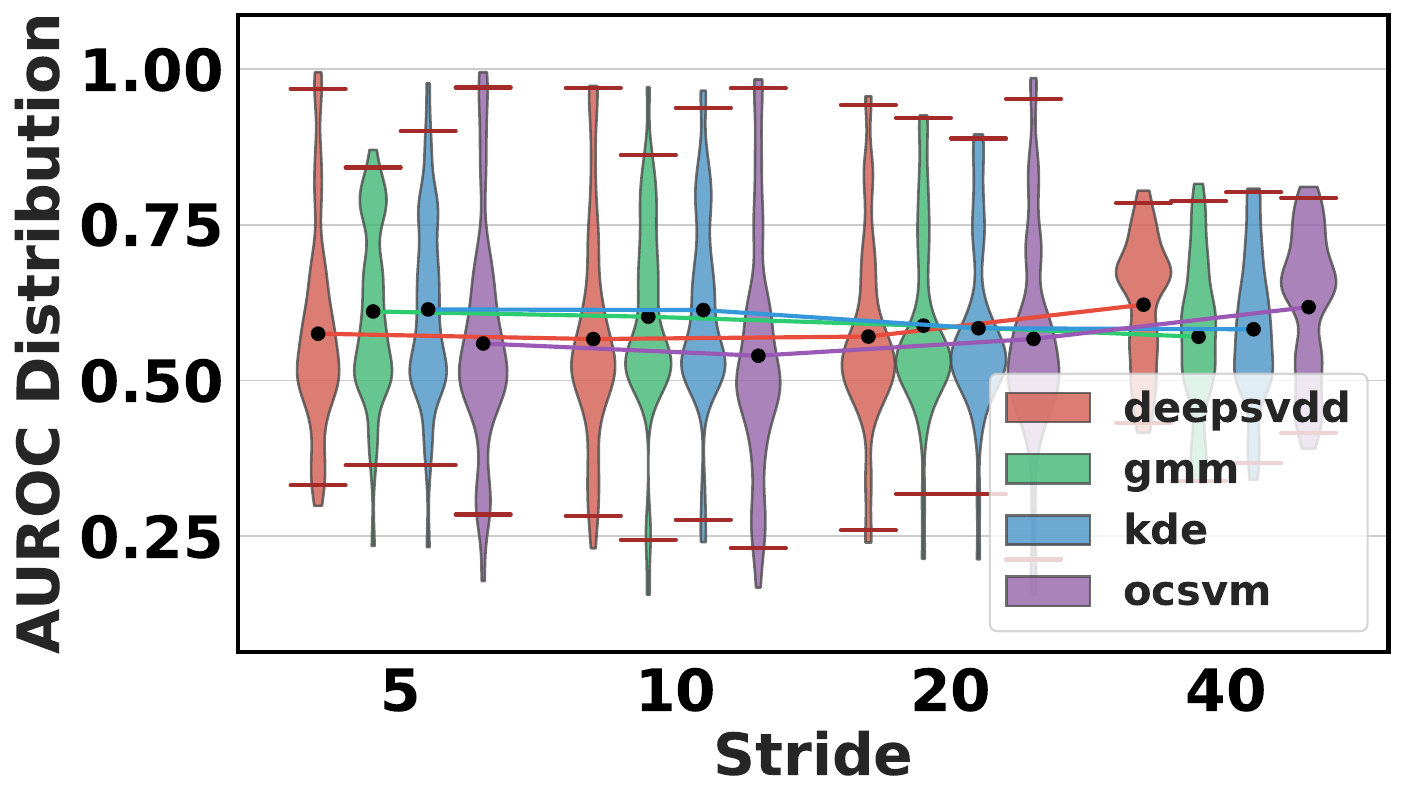}}
    \subfloat[HDFS - SBERT]{\includegraphics[width=0.24\textwidth]{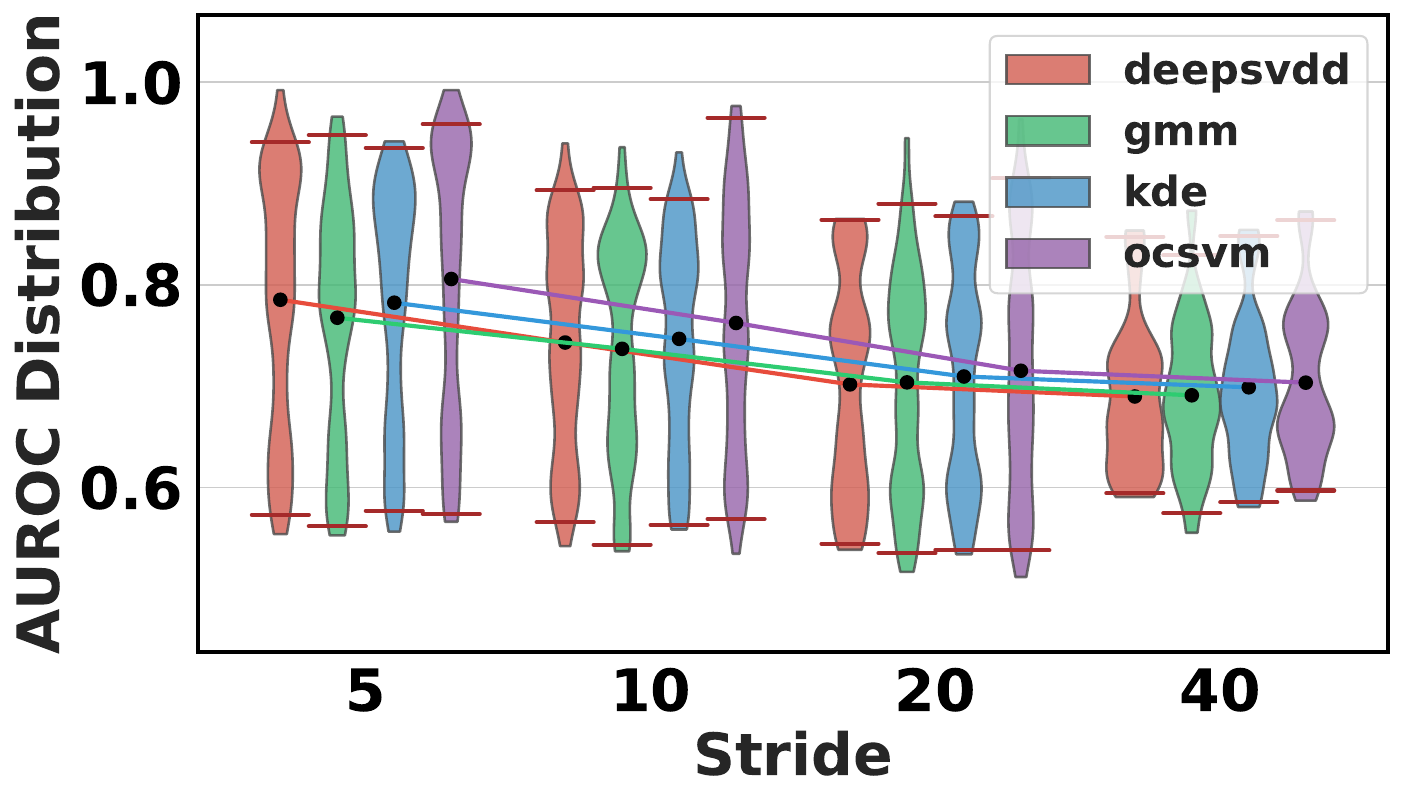}}
    \subfloat[HDFS - Qwen-8B]{\includegraphics[width=0.24\textwidth]{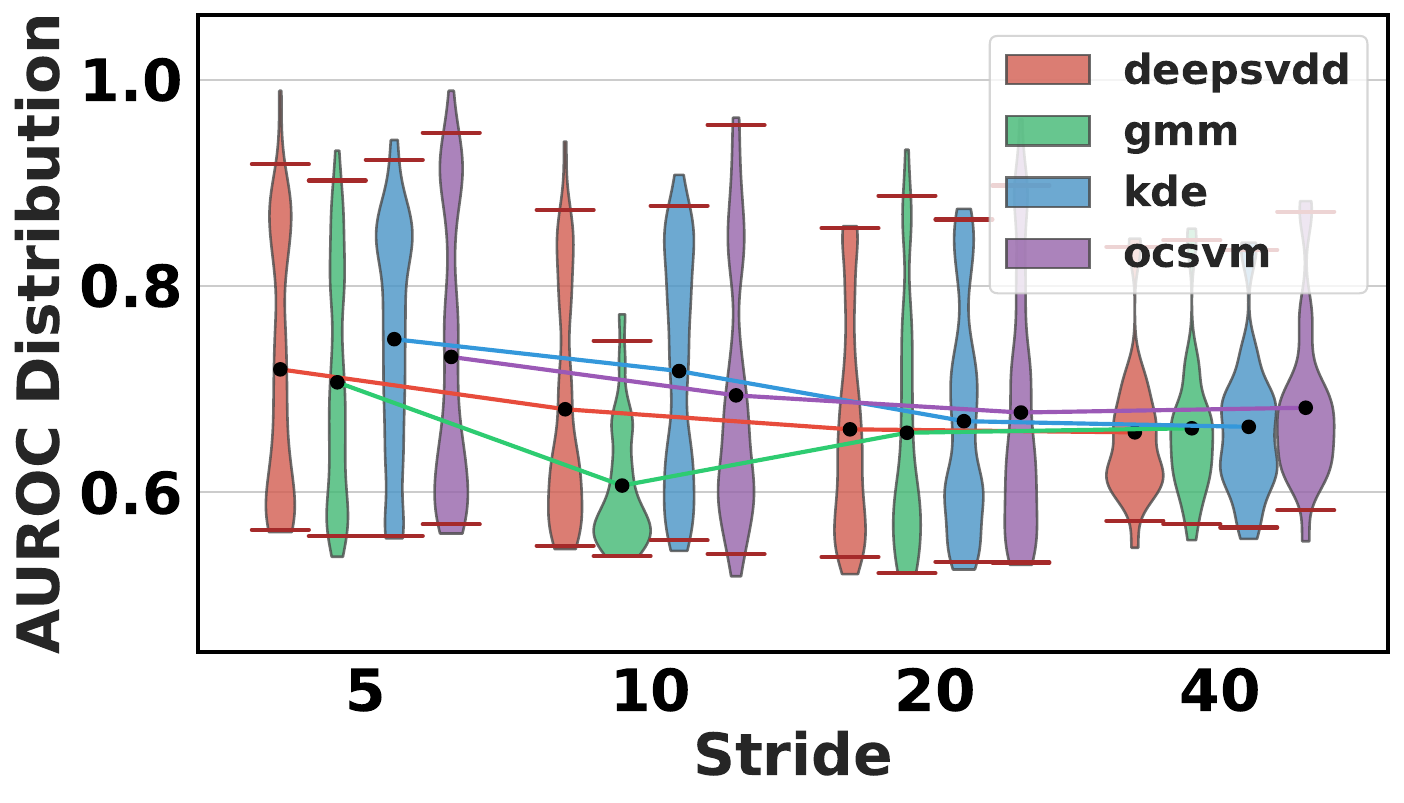}} \\
    \subfloat[BGL - TF-IDF]{\includegraphics[width=0.24\textwidth]{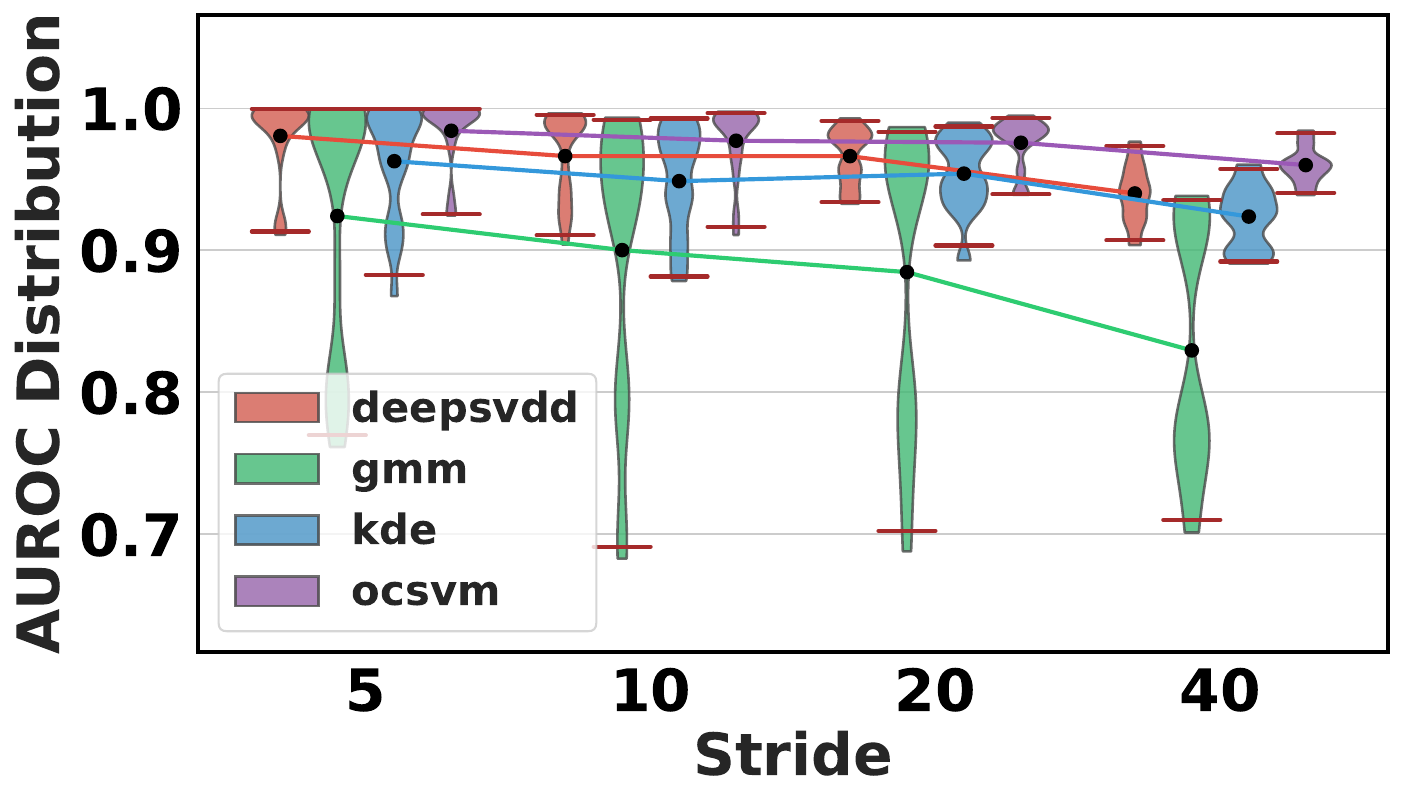}}
    \subfloat[BGL - Word2Vec]{\includegraphics[width=0.24\textwidth]{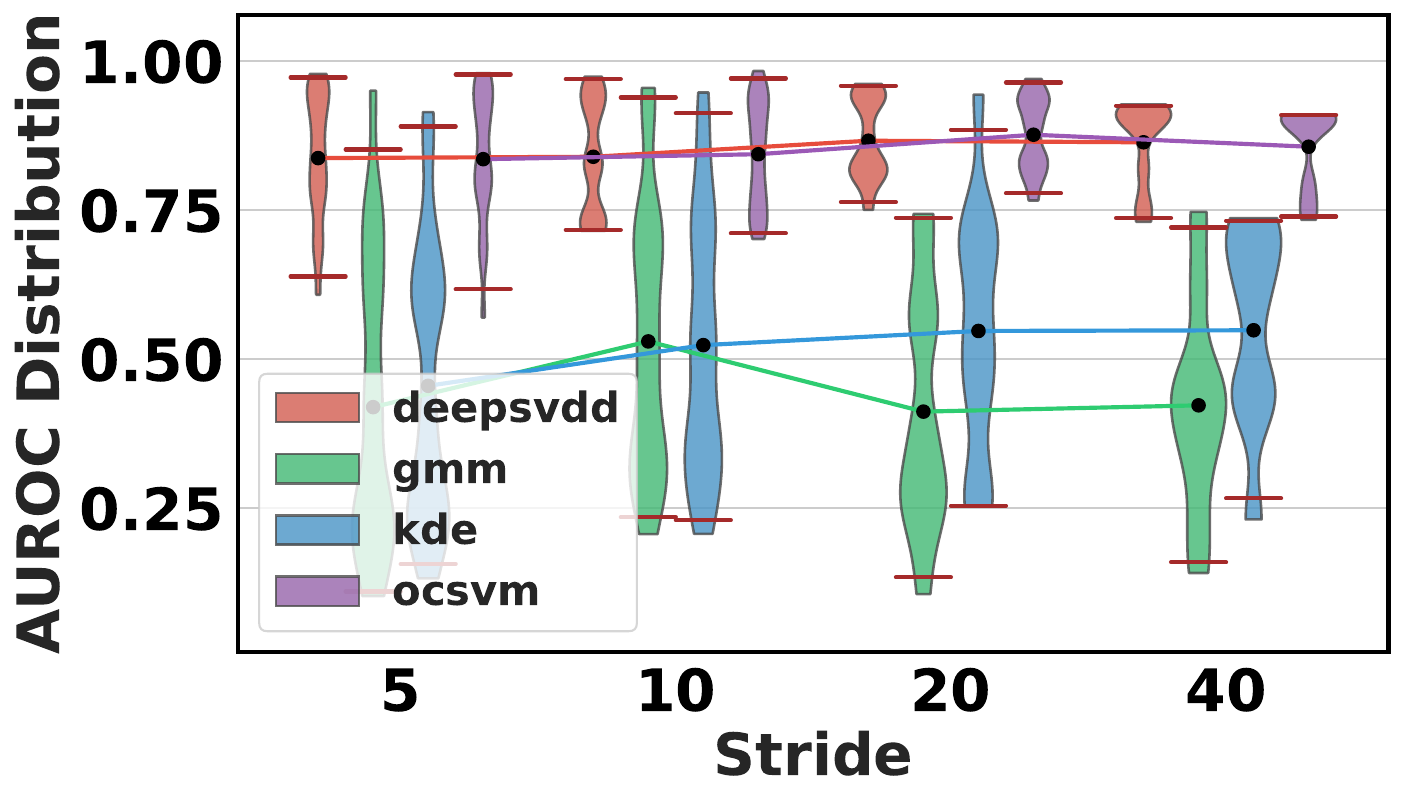}}
    \subfloat[BGL - SBERT]{\includegraphics[width=0.24\textwidth]{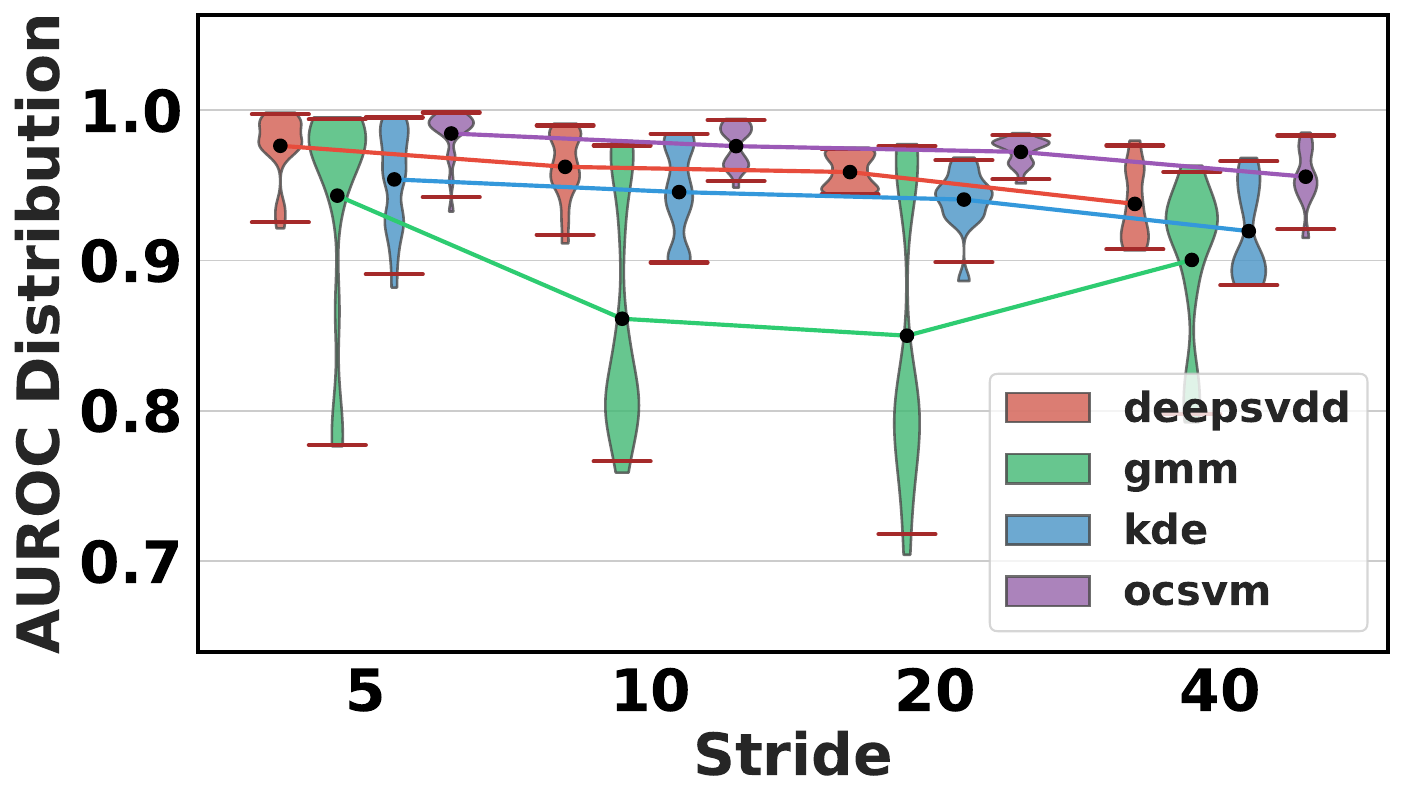}}
    \subfloat[BGL - Qwen-8B]{\includegraphics[width=0.24\textwidth]{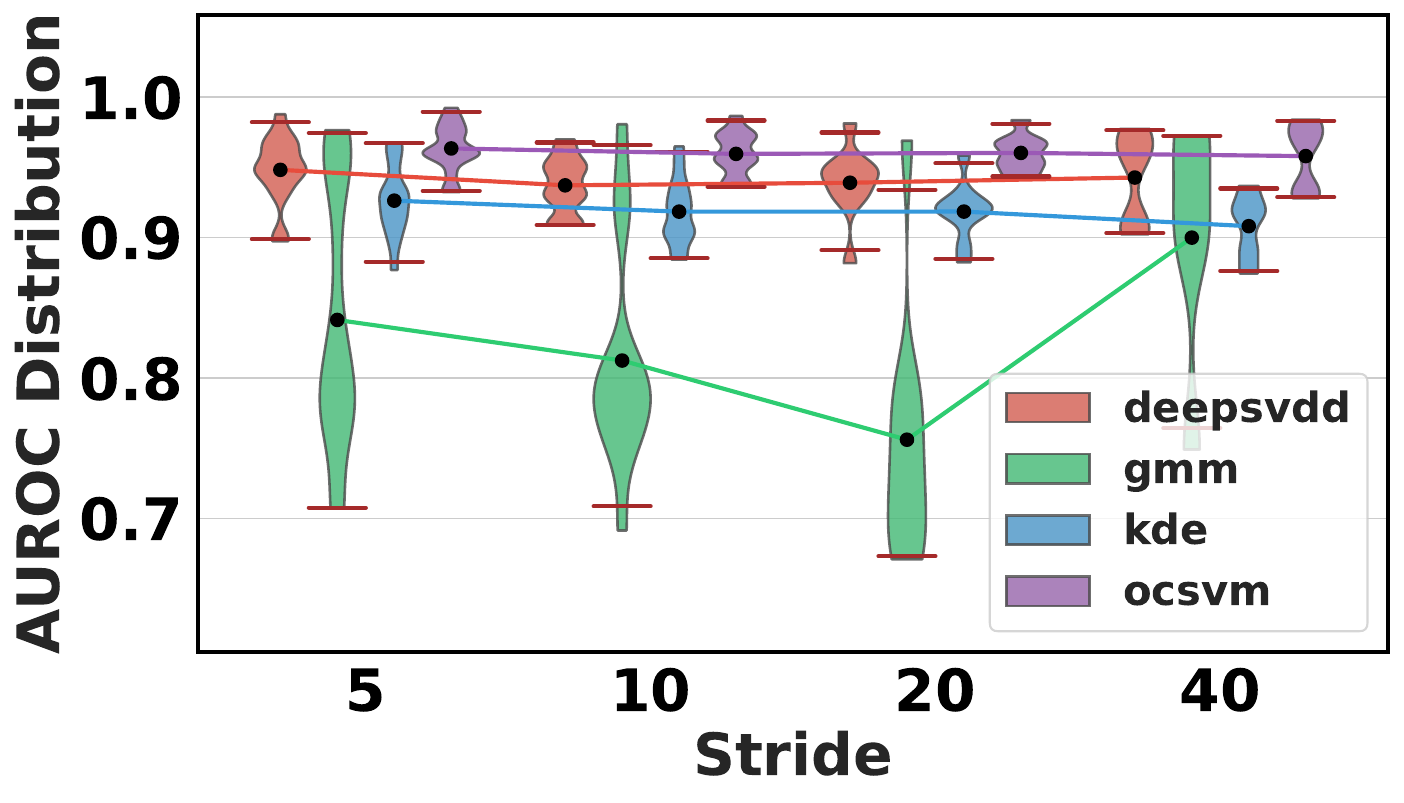}} \\
    \subfloat[Thunderbird - TF-IDF]{\includegraphics[width=0.24\textwidth]{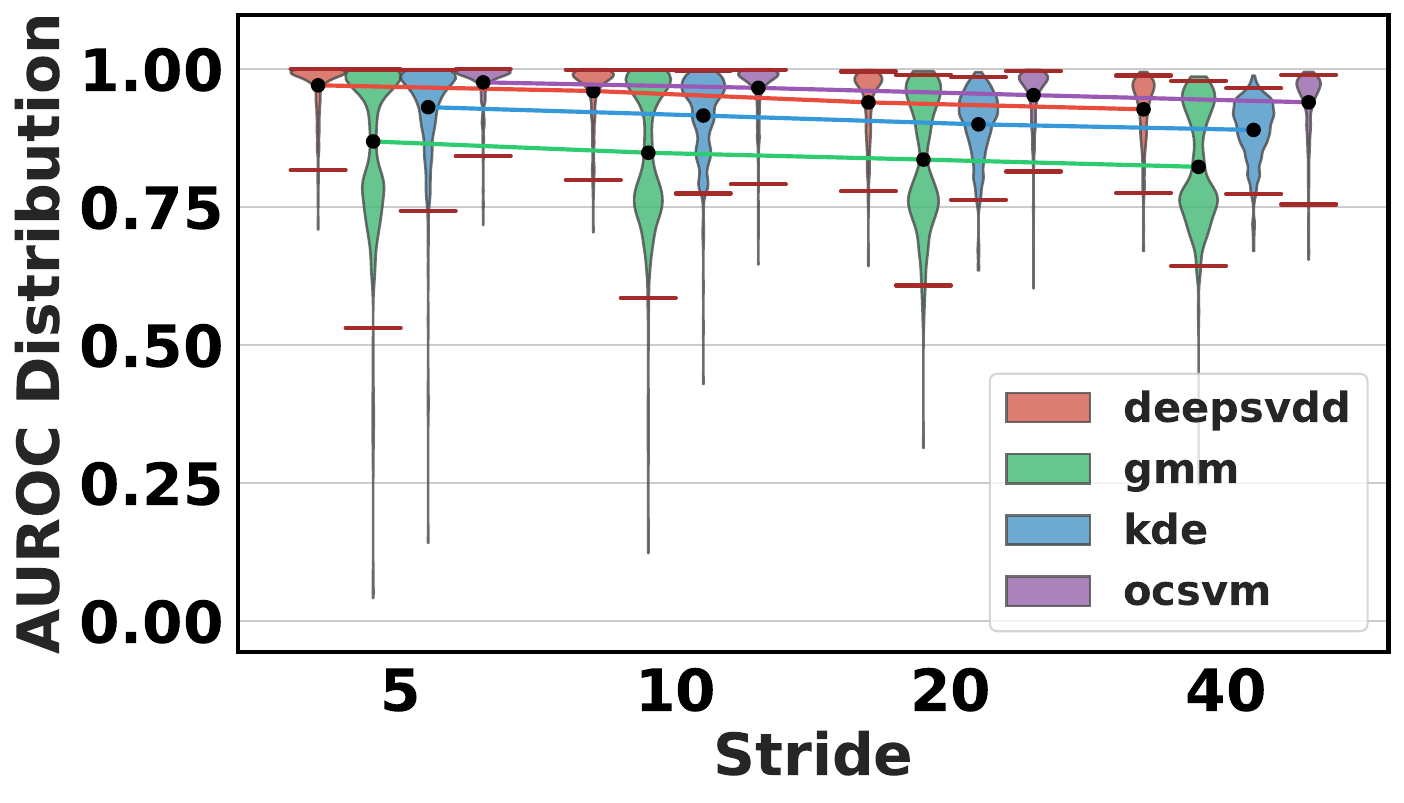}}
    \subfloat[Thunderbird - Word2Vec]{\includegraphics[width=0.24\textwidth]{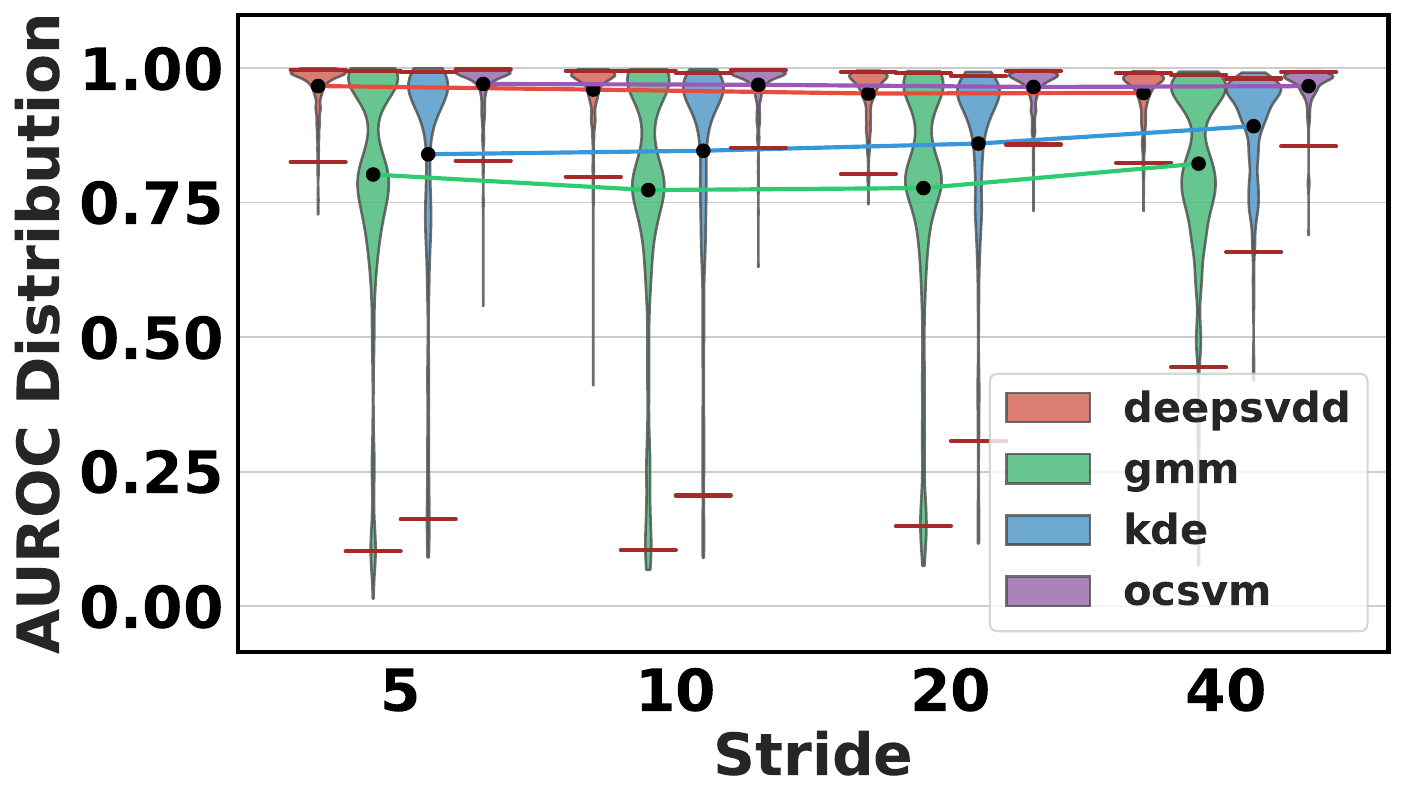}}
    \subfloat[Thunderbird - SBERT]{\includegraphics[width=0.24\textwidth]{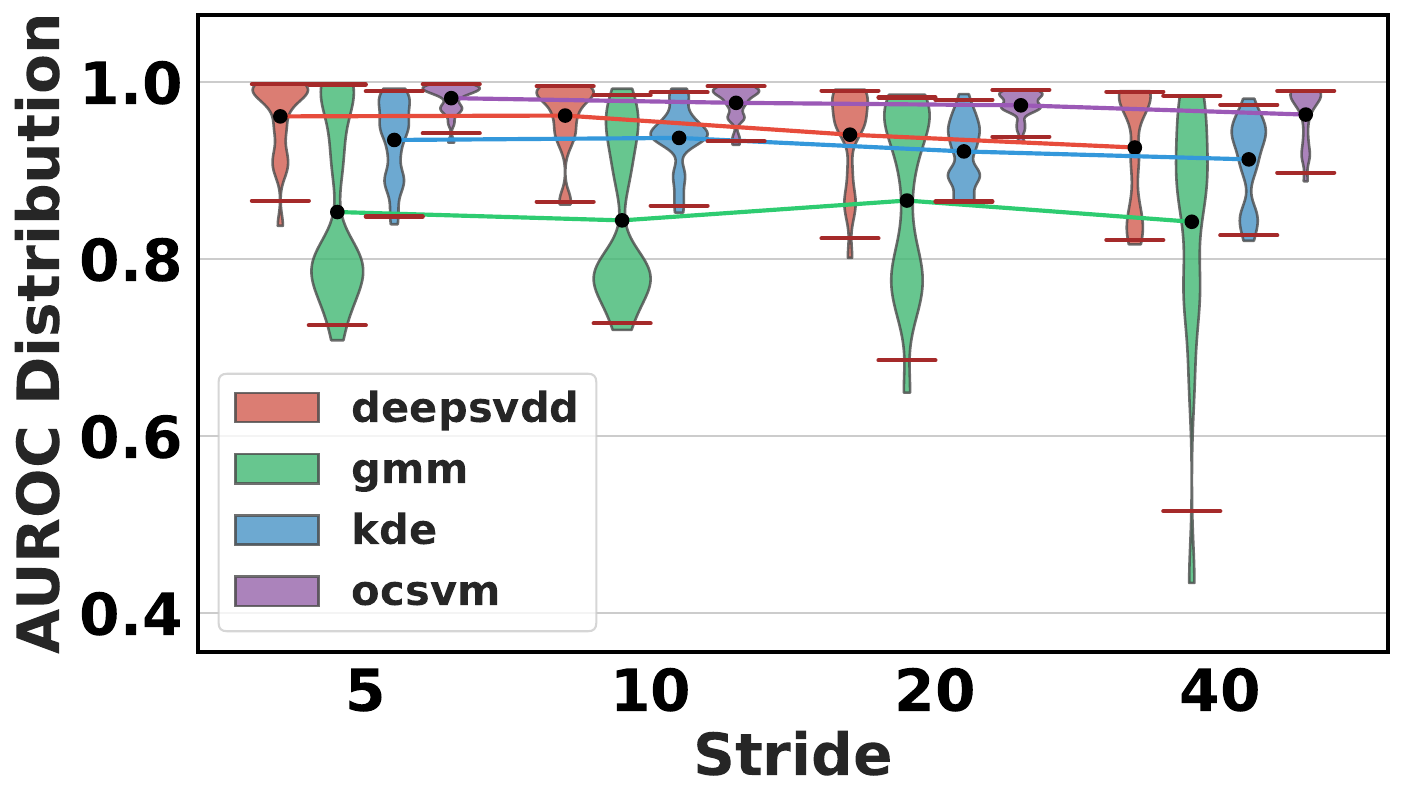}}
    \subfloat[Thunderbird - Qwen-8B]{\includegraphics[width=0.24\textwidth]{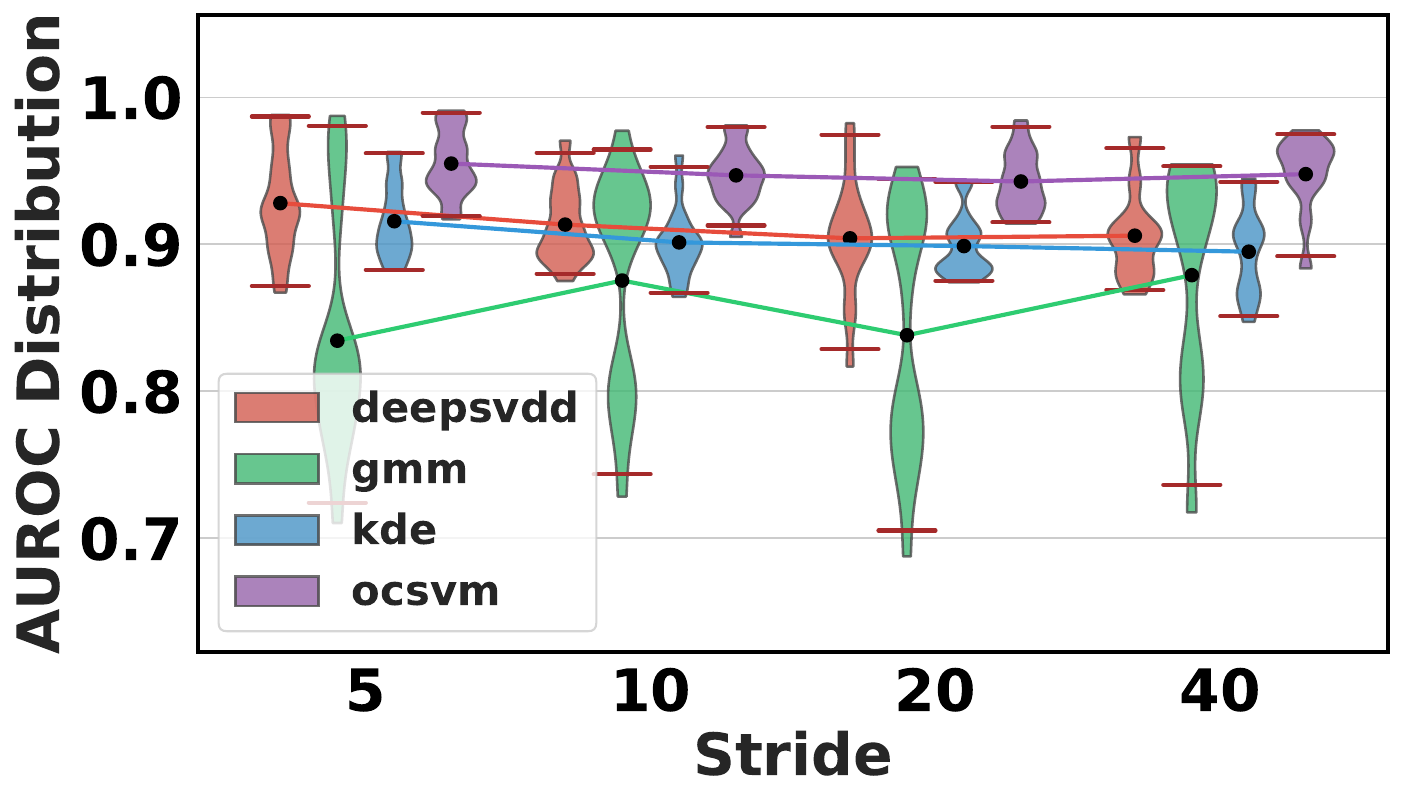}}
    \caption{AUROC scores for each detector as a function of \textbf{stride size}, shown across datasets and embedding methods. Each experiment utilized 5K training samples, 5K normal test samples, and 5K anomalous test samples. Results for Qwen3-embedding-4B and Qwen3-embedding-0.6B embeddings are omitted due to their performance parity with the 8B model.}
    \label{fig:stride_results}
\end{figure*}

\noindent\textbf{Stride Size Analysis: }
Fig.~\ref{fig:stride_results} shows AUROC distributions at different stride values. Smaller strides ($S$=5) consistently yield higher performance across datasets, embeddings, and detectors. With TF-IDF, AUROC drops notably as the stride increases. For example, on HDFS, mean AUROC declines from $0.812$ at $S$=5 to $0.687$ at $S$=40. SBERT and Qwen show slightly less degradation: on BGL, Qwen falls from $0.974$ to $0.918$ between $S$=5 and $S$=40, and SBERT from $0.961$ to $0.912$. On HDFS, Qwen drops from $0.926$ to $0.831$ and SBERT from $0.905$ to $0.812$. In contrast, Word2Vec is more stride-resilient. For instance, on Thunderbird, AUROC increases from $0.841$ to $0.908$ as the stride grows.

From detectors, DeepSVDD and OCSVM are the most accurate overall, with average AUROCs of $0.836$ and $0.835$, respectively. GMM performs the worst overall, with an average AUROC of $0.703$, and shows pronounced degradation with increasing stride. KDE is only slightly better than GMM, with an average of $0.752$

Taken together, \ul{$S$=5 is the most reliable setting. The best-performing detector is DeepSVDD, with OCSVM a close second.}

\begin{figure*}[h]
    \centering
    \subfloat[HDFS - TF-IDF]{\includegraphics[width=0.24\textwidth]{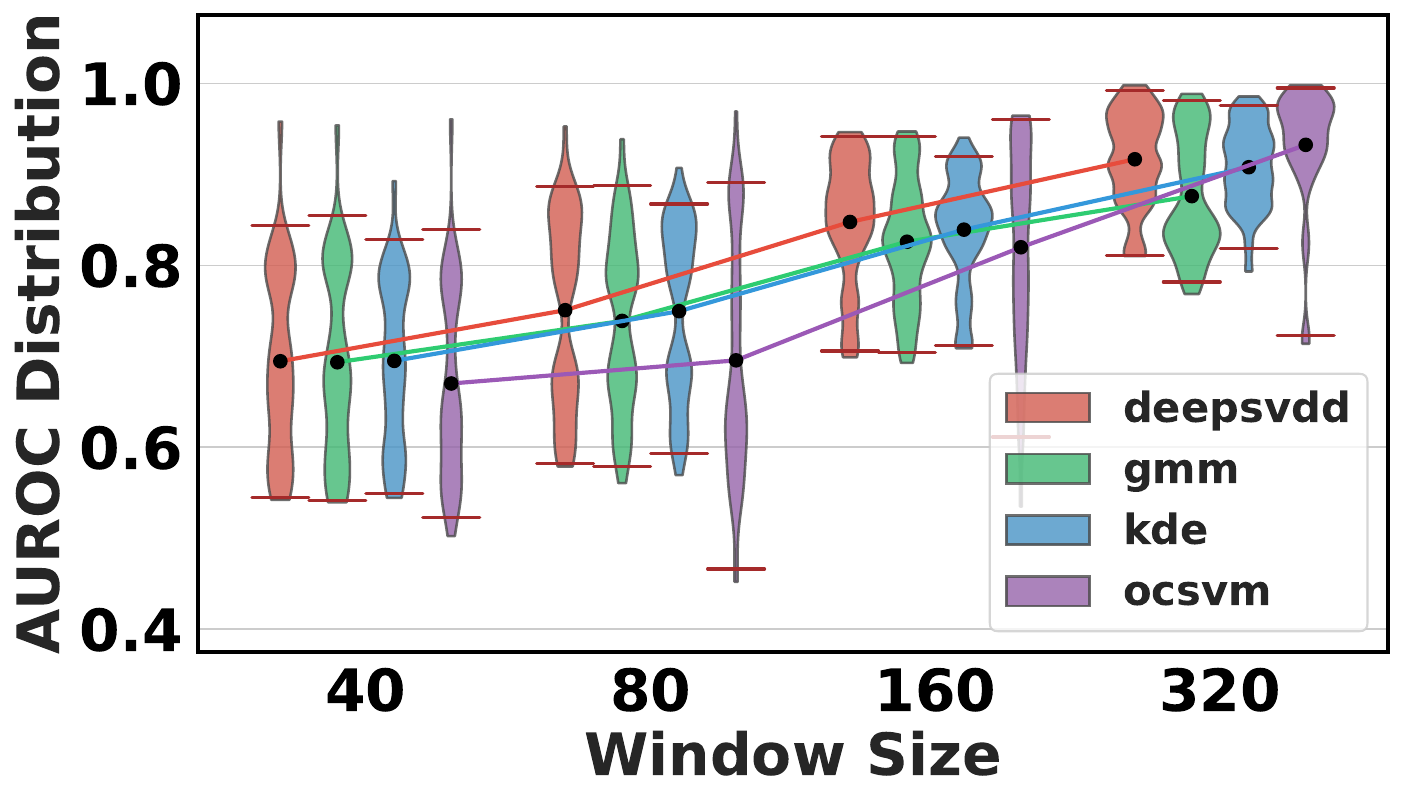}}
    \subfloat[HDFS - Word2Vec]{\includegraphics[width=0.24\textwidth]{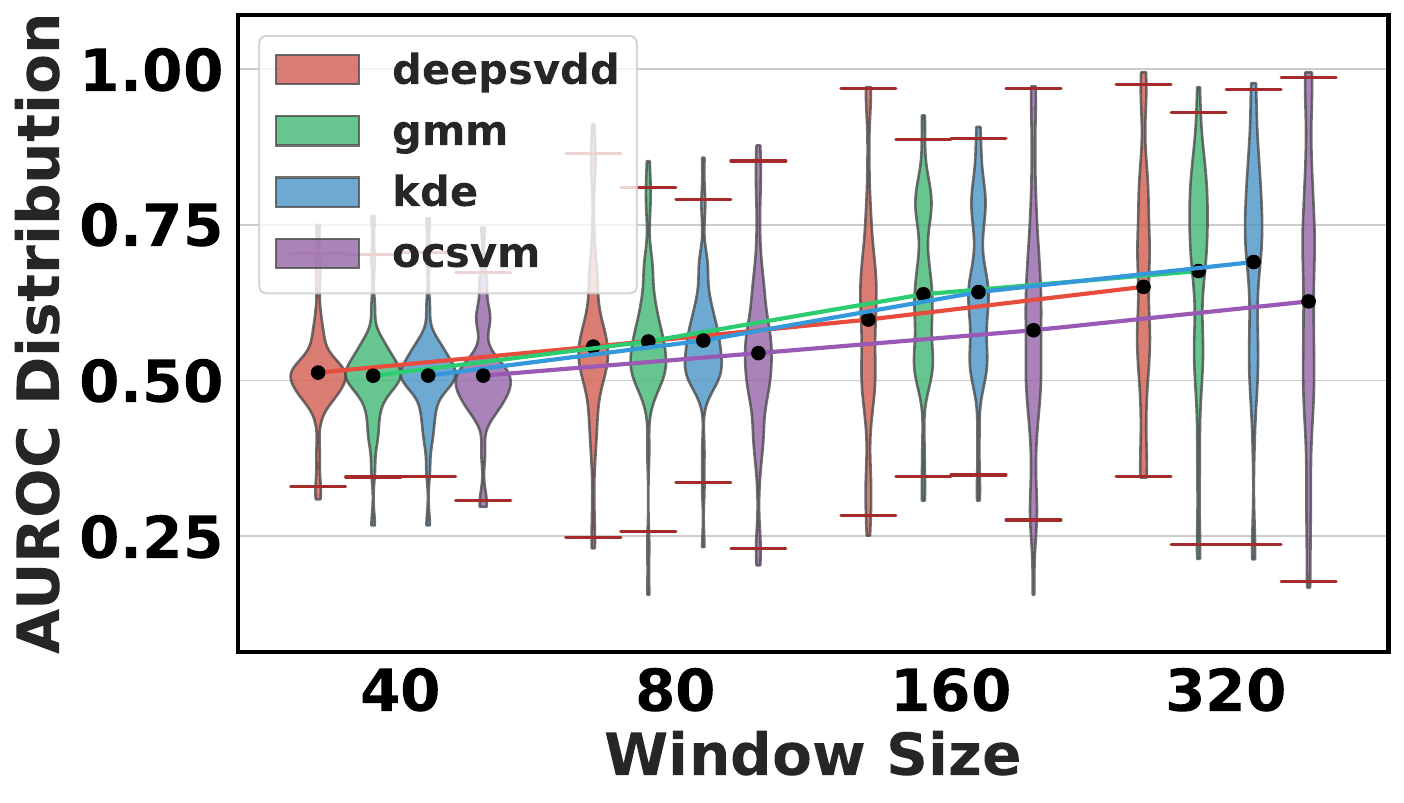}}
    \subfloat[HDFS - SBERT]{\includegraphics[width=0.24\textwidth]{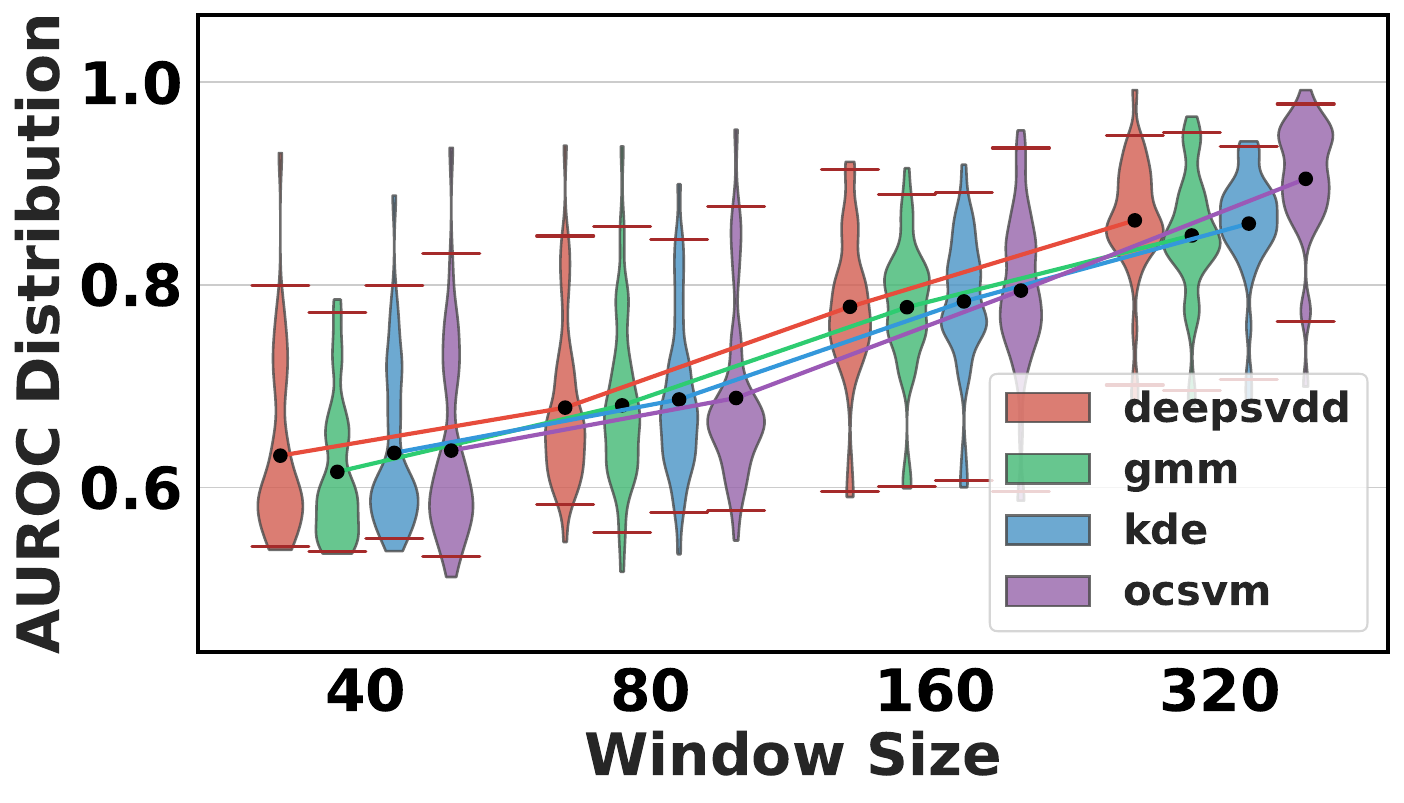}}
    \subfloat[HDFS - Qwen-8B]{\includegraphics[width=0.24\textwidth]{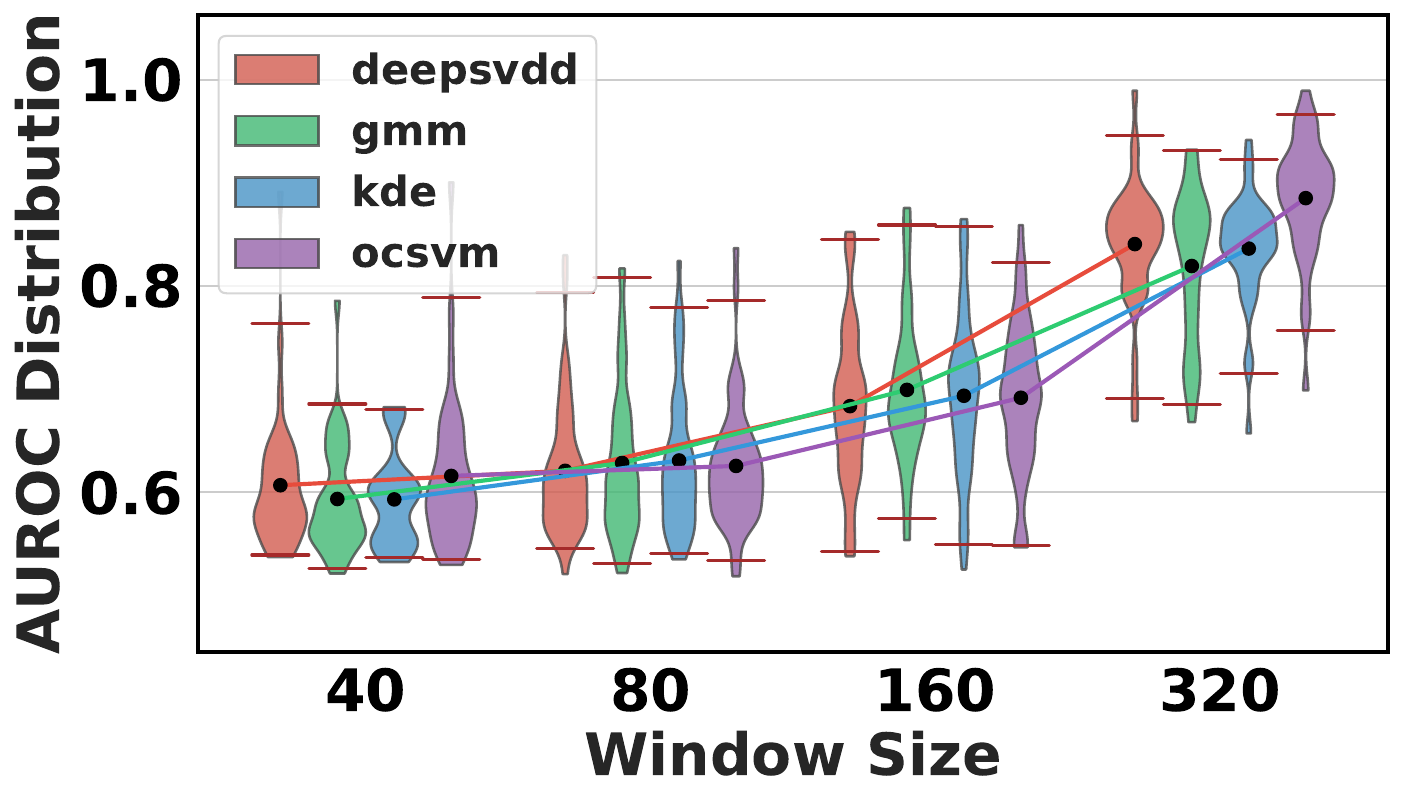}} \\
    \subfloat[BGL - TF-IDF]{\includegraphics[width=0.24\textwidth]{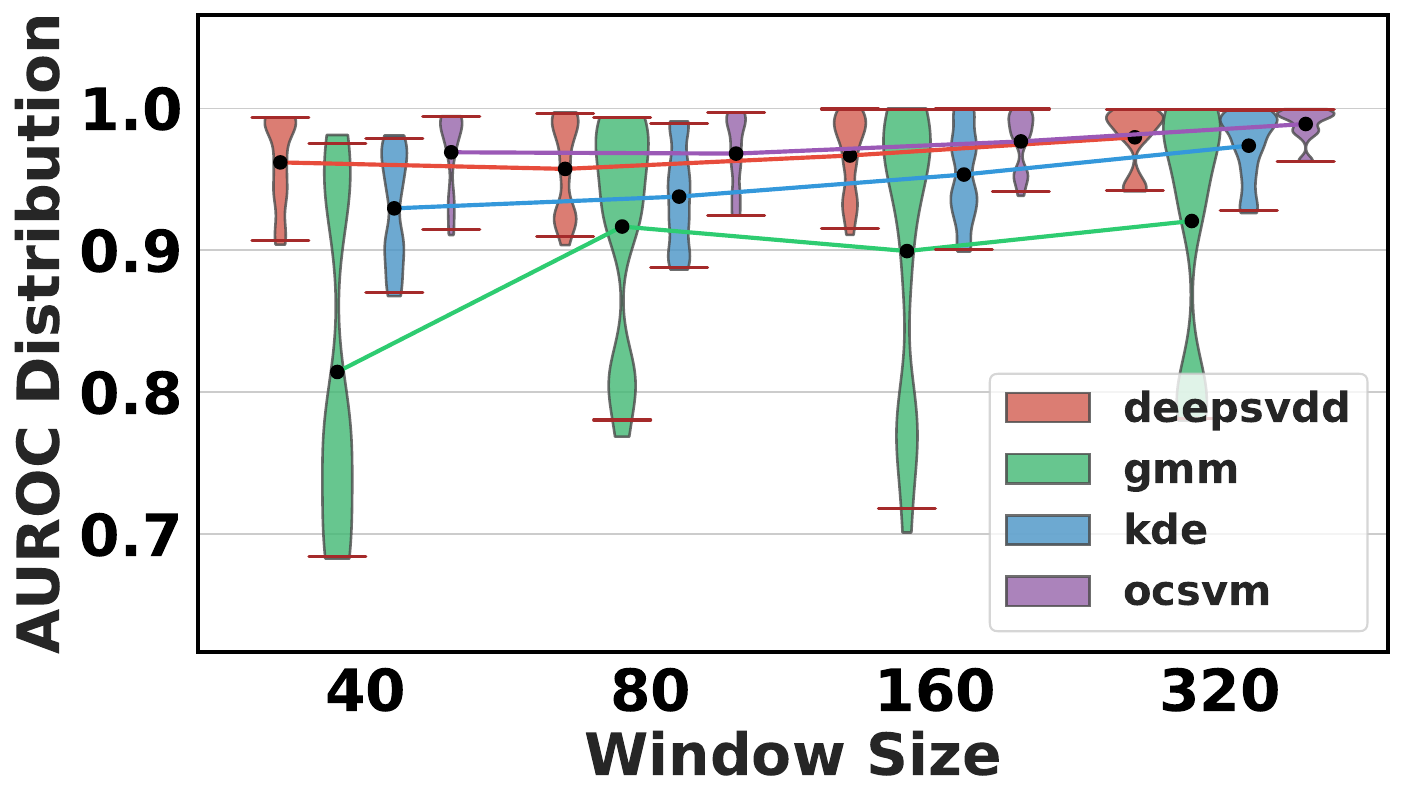}}
    \subfloat[BGL - Word2Vec]{\includegraphics[width=0.24\textwidth]{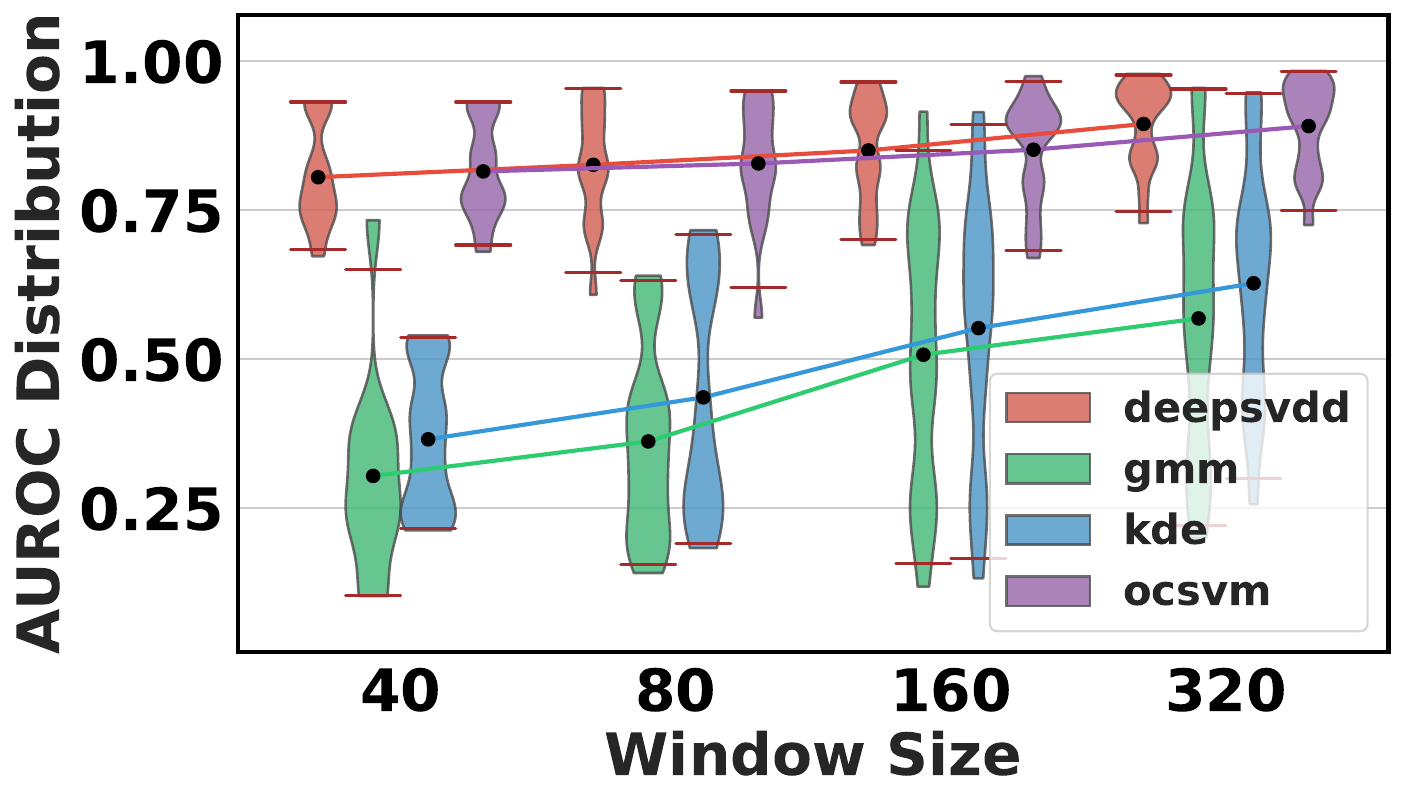}}
    \subfloat[BGL - SBERT]{\includegraphics[width=0.24\textwidth]{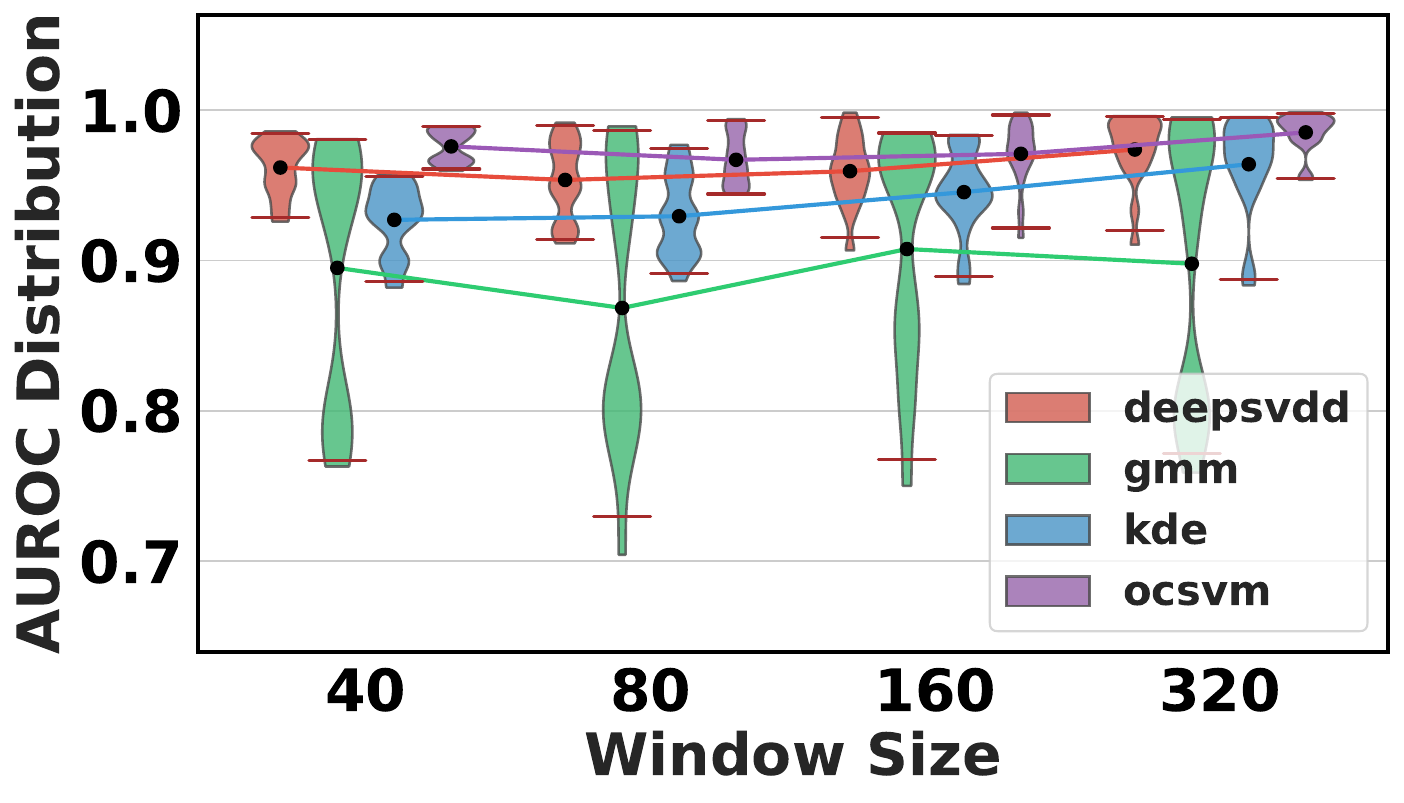}}
    \subfloat[BGL - Qwen-8B]{\includegraphics[width=0.24\textwidth]{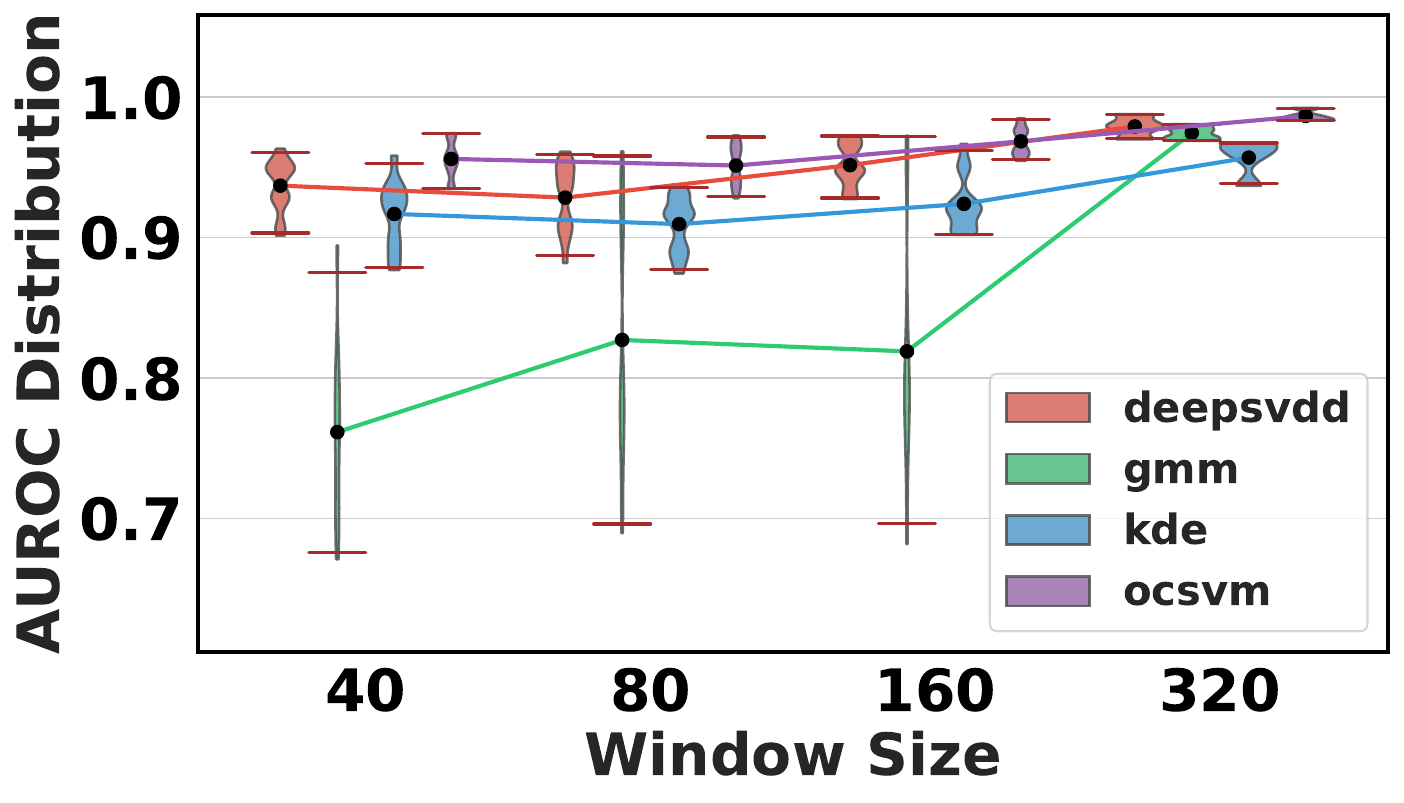}} \\
    \subfloat[Thunderbird - TF-IDF]{\includegraphics[width=0.24\textwidth]{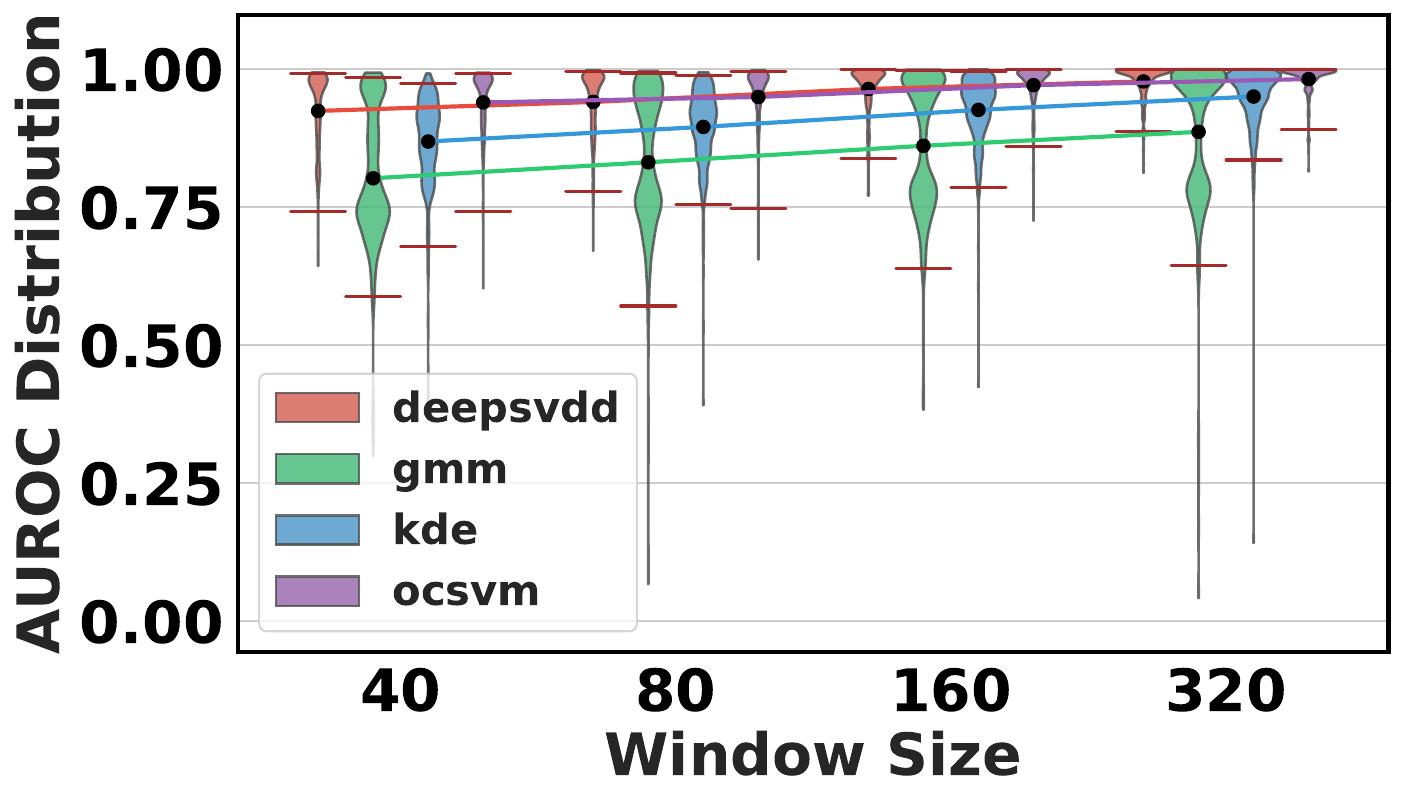}}
    \subfloat[Thunderbird - Word2Vec]{\includegraphics[width=0.24\textwidth]{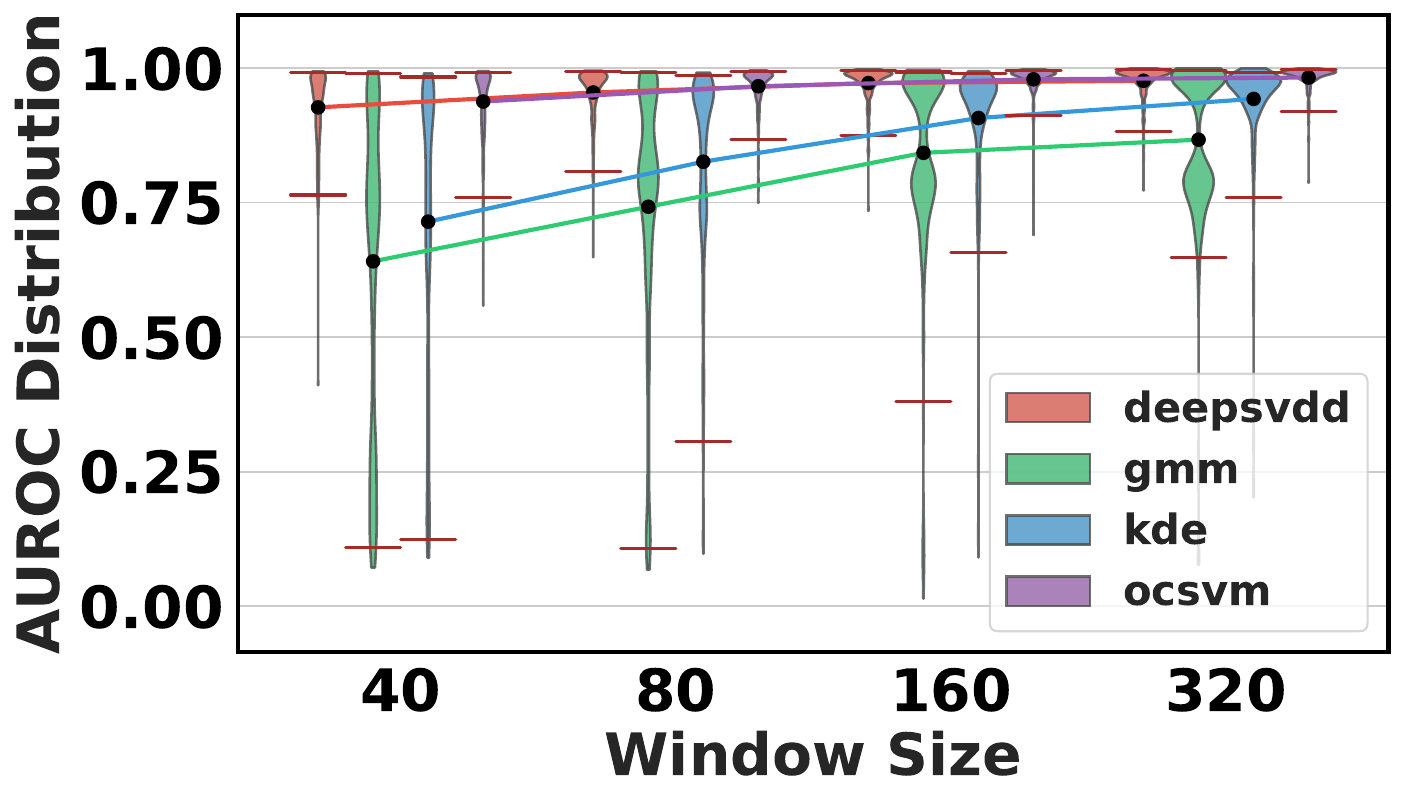}}
    \subfloat[Thunderbird - SBER]{\includegraphics[width=0.24\textwidth]{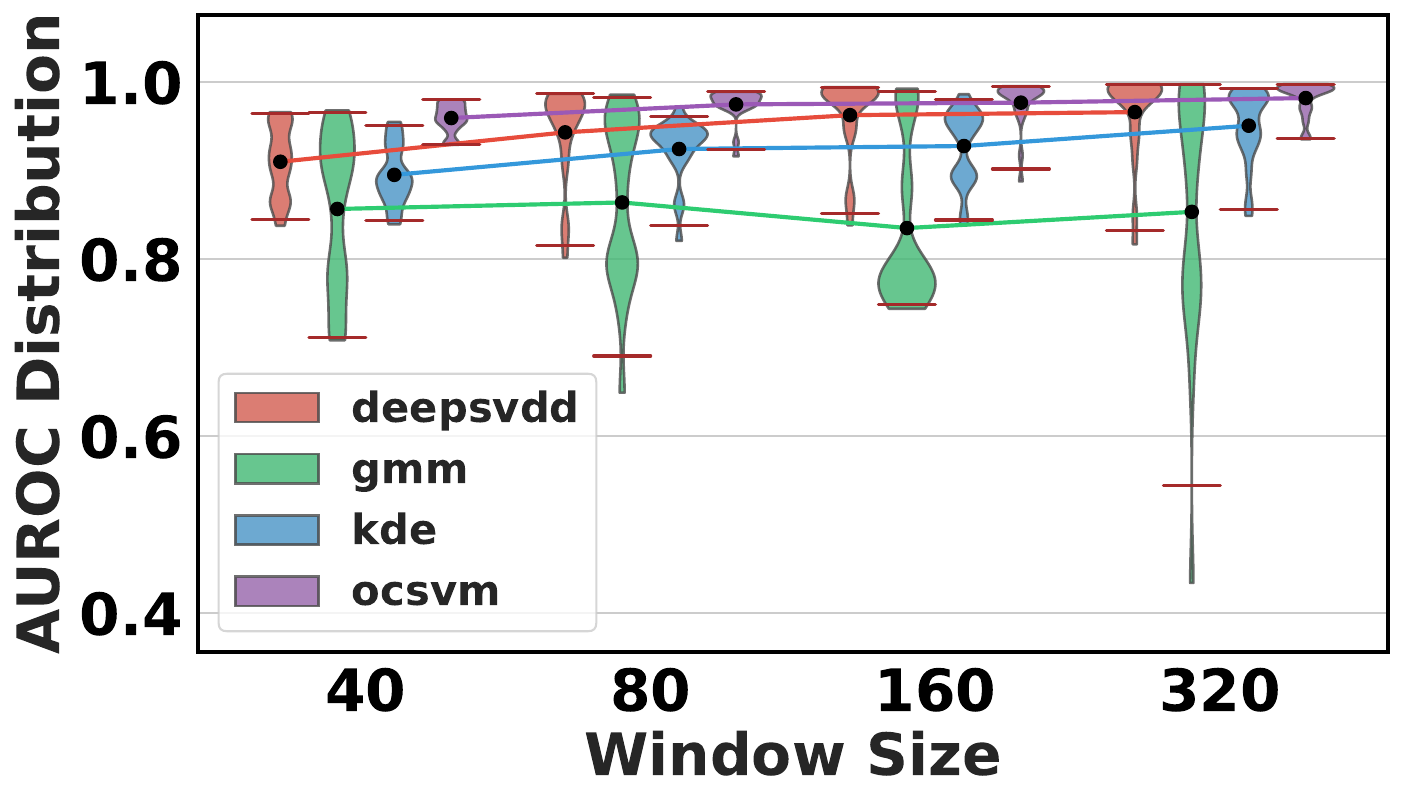}}
    \subfloat[Thunderbird - Qwen-8B]{\includegraphics[width=0.24\textwidth]{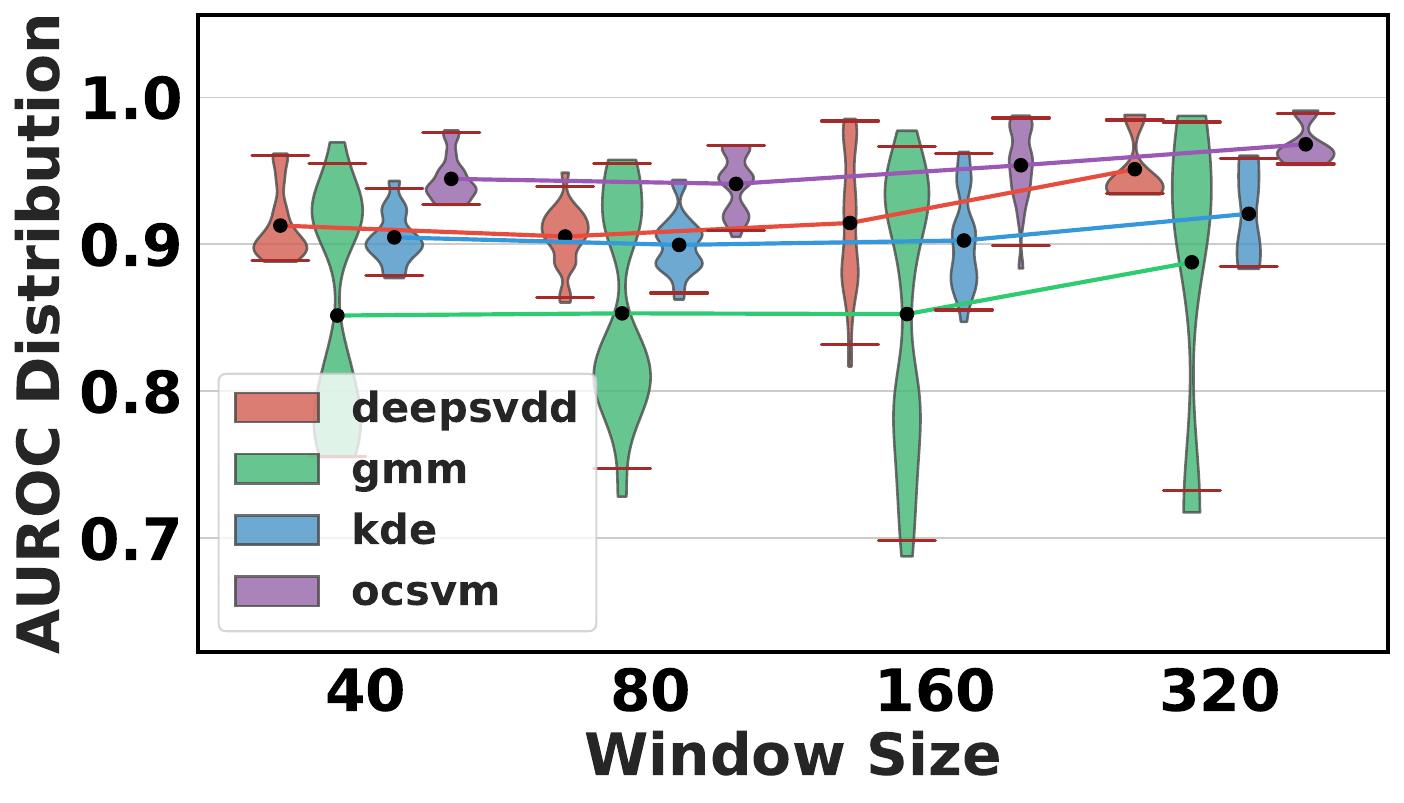}}
    \caption{AUROC scores for each detector as a function of \textbf{window size}, shown across datasets and embedding methods. Other settings remain the same as in Figure~\ref{fig:stride_results}}
    \label{fig:window_size_results}
\end{figure*}

\vspace{1ex}
\noindent\textbf{Window Size Analysis:}
Fig.~\ref{fig:window_size_results} presents AUROC distributions across window sizes $\{40, 80, 160, 320\}$. Overall, larger windows lead to higher accuracy. With TF-IDF, mean AUROC increases from $0.688$ at $W$=40 to $0.903$ at $W$=320 on HDFS, and from $0.919$ to $0.954$ on BGL as window size increases. With Word2Vec, AUROC rises from $0.805$ to $0.940$ on Thunderbird, and from $0.572$ to $0.732$ on BGL.

Moving to the detector, DeepSVDD ($0.8470$ average) and OCSVM ($0.8465$) are again the top-performing detectors. In contrast, KDE ($0.7632$) and GMM ($0.7152$) not only perform significantly worse but also exhibit much larger variability.

Overall, \ul{$W$=320 delivers the best AUROC in most configurations, particularly when paired with DeepSVDD and OCSVM.}

\begin{figure*}[htbp]
    \centering
    \subfloat[HDFS - TF-IDF]{\includegraphics[width=0.24\textwidth]{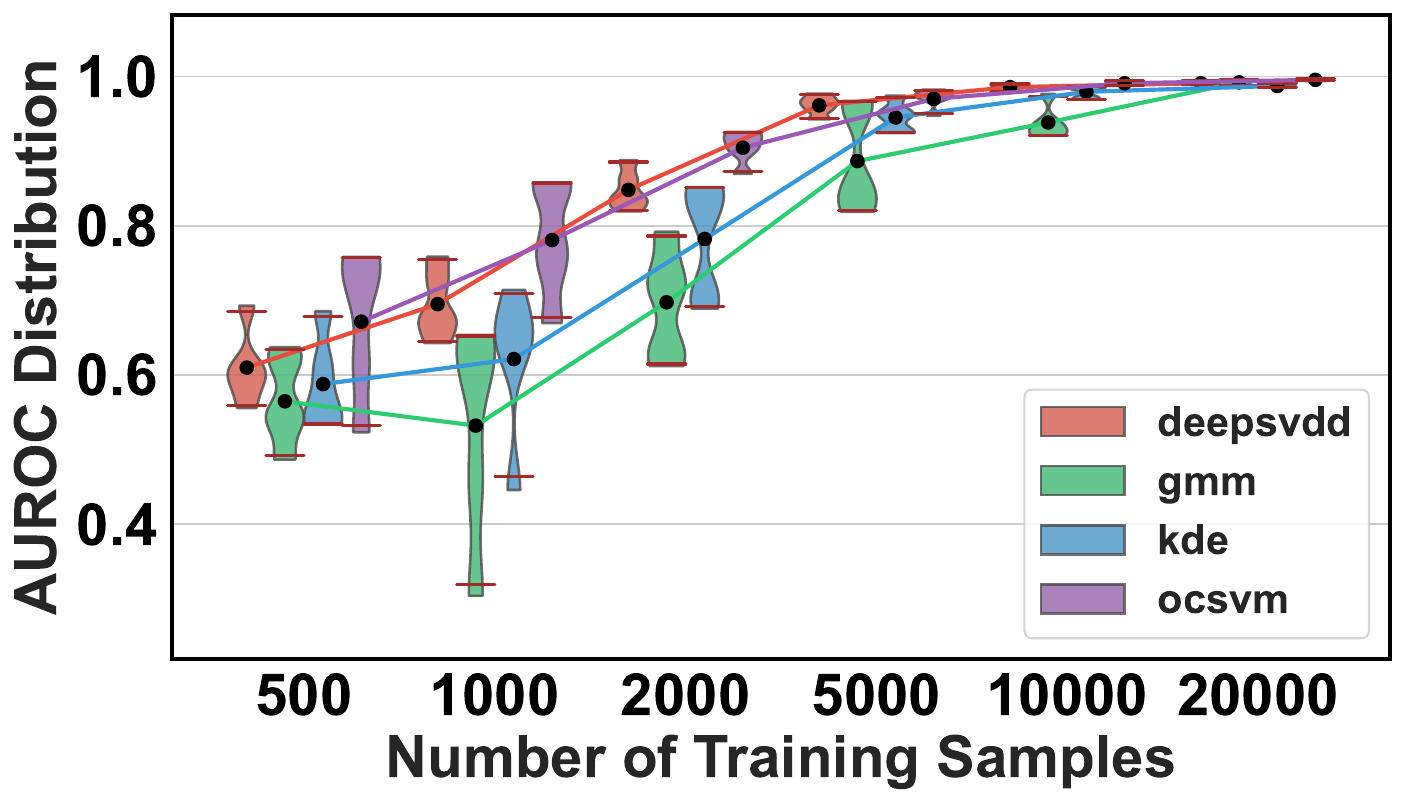}}
    \subfloat[HDFS - Word2Vec]{\includegraphics[width=0.24\textwidth]{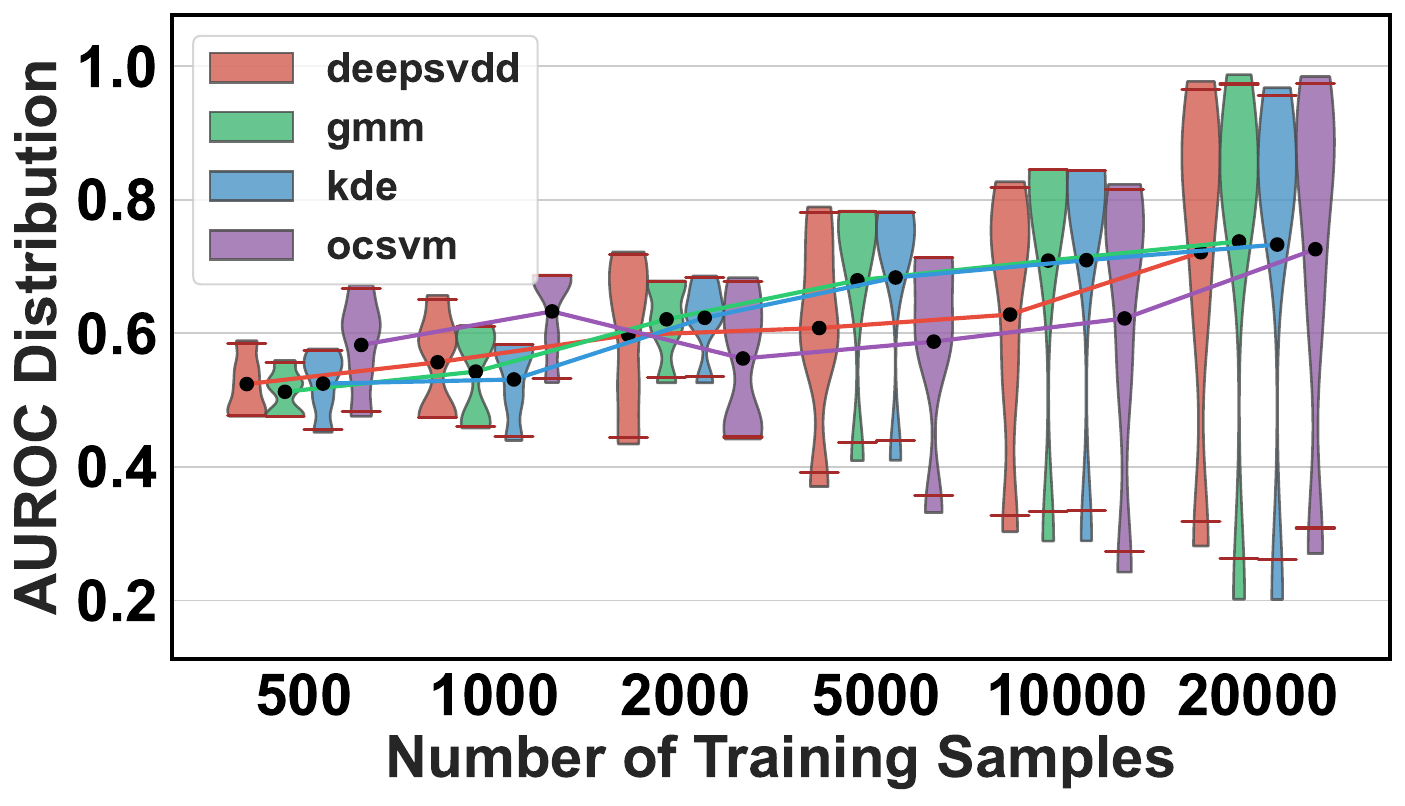}}
    \subfloat[HDFS - SBERT]{\includegraphics[width=0.24\textwidth]{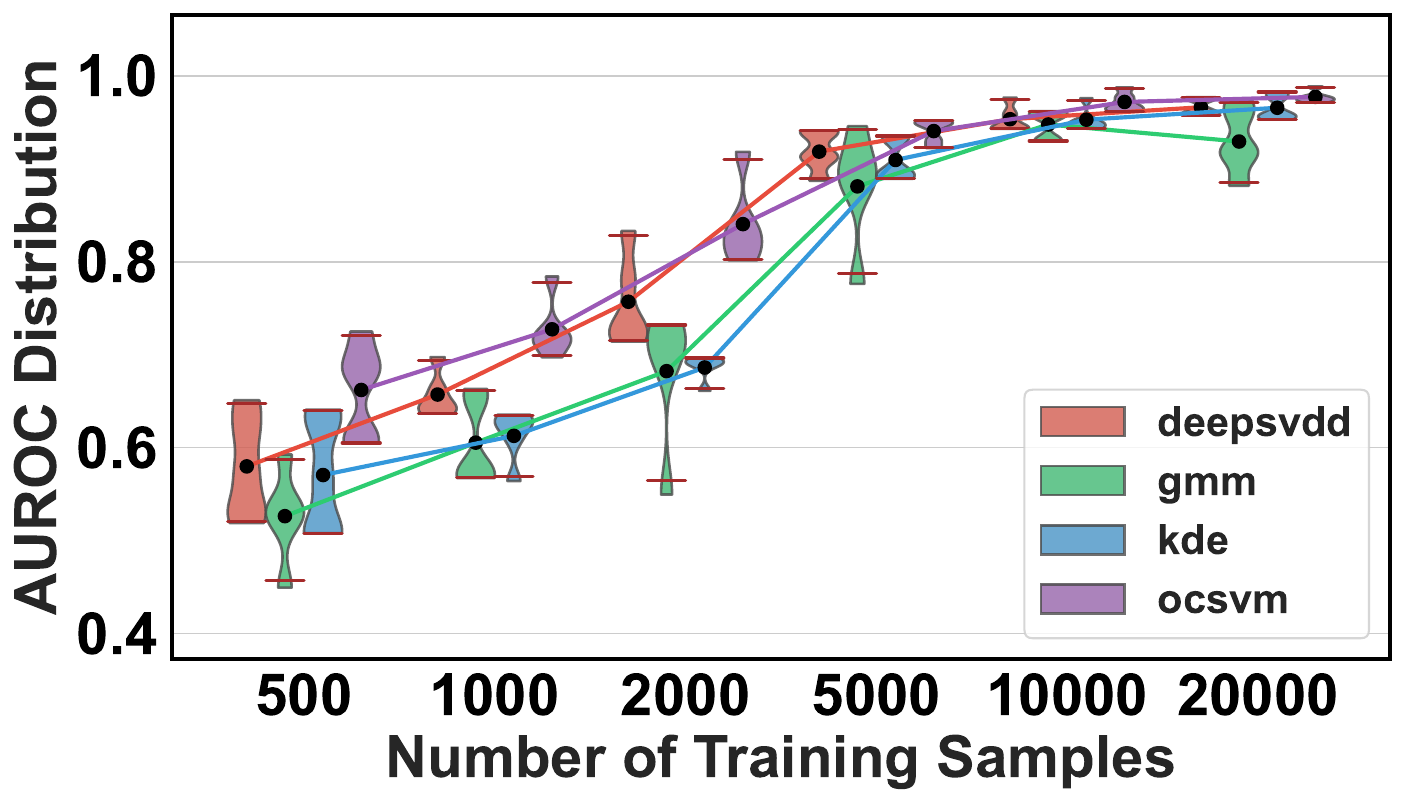}}
    \subfloat[HDFS - Qwen-8B]{\includegraphics[width=0.24\textwidth]{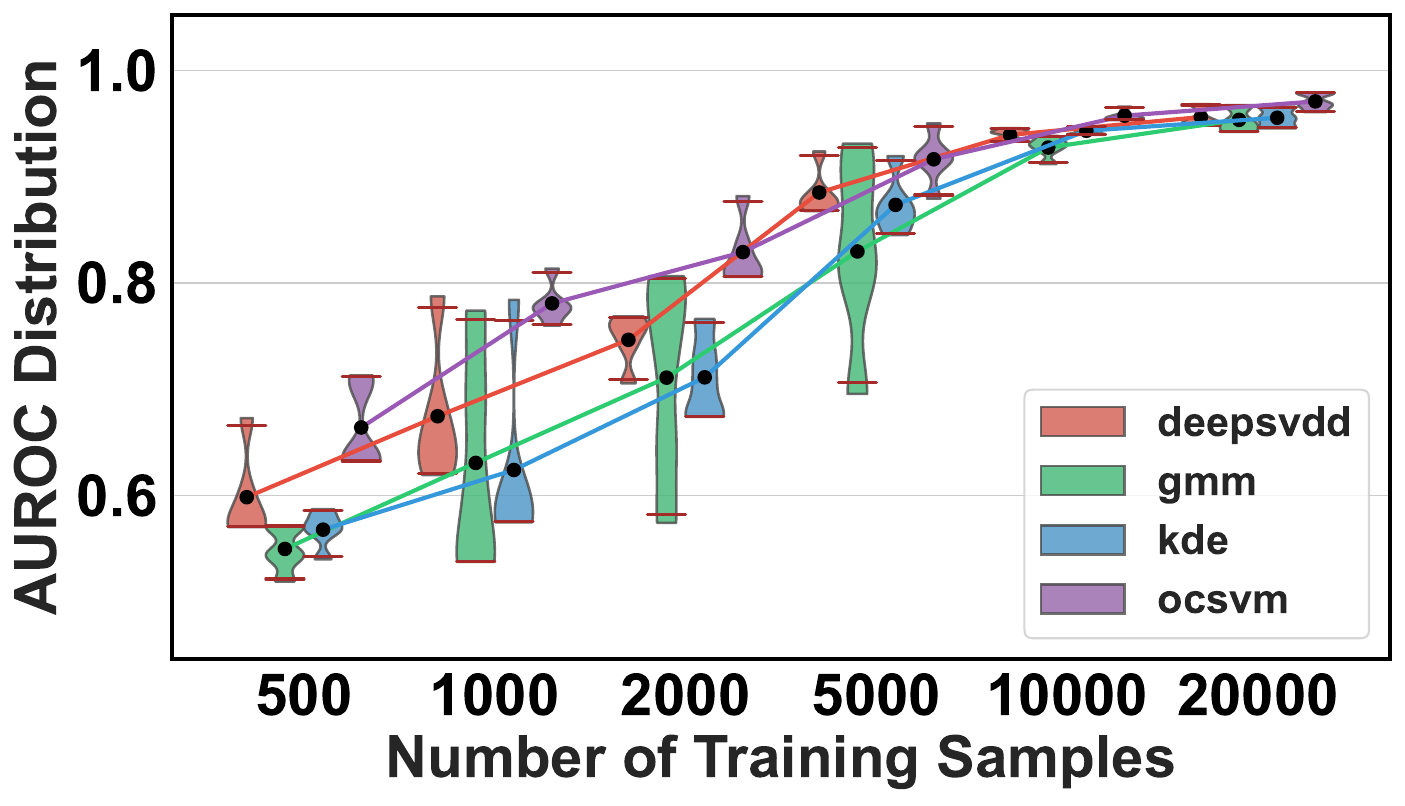}} \\
    \subfloat[BGL - TF-IDF]{\includegraphics[width=0.24\textwidth]{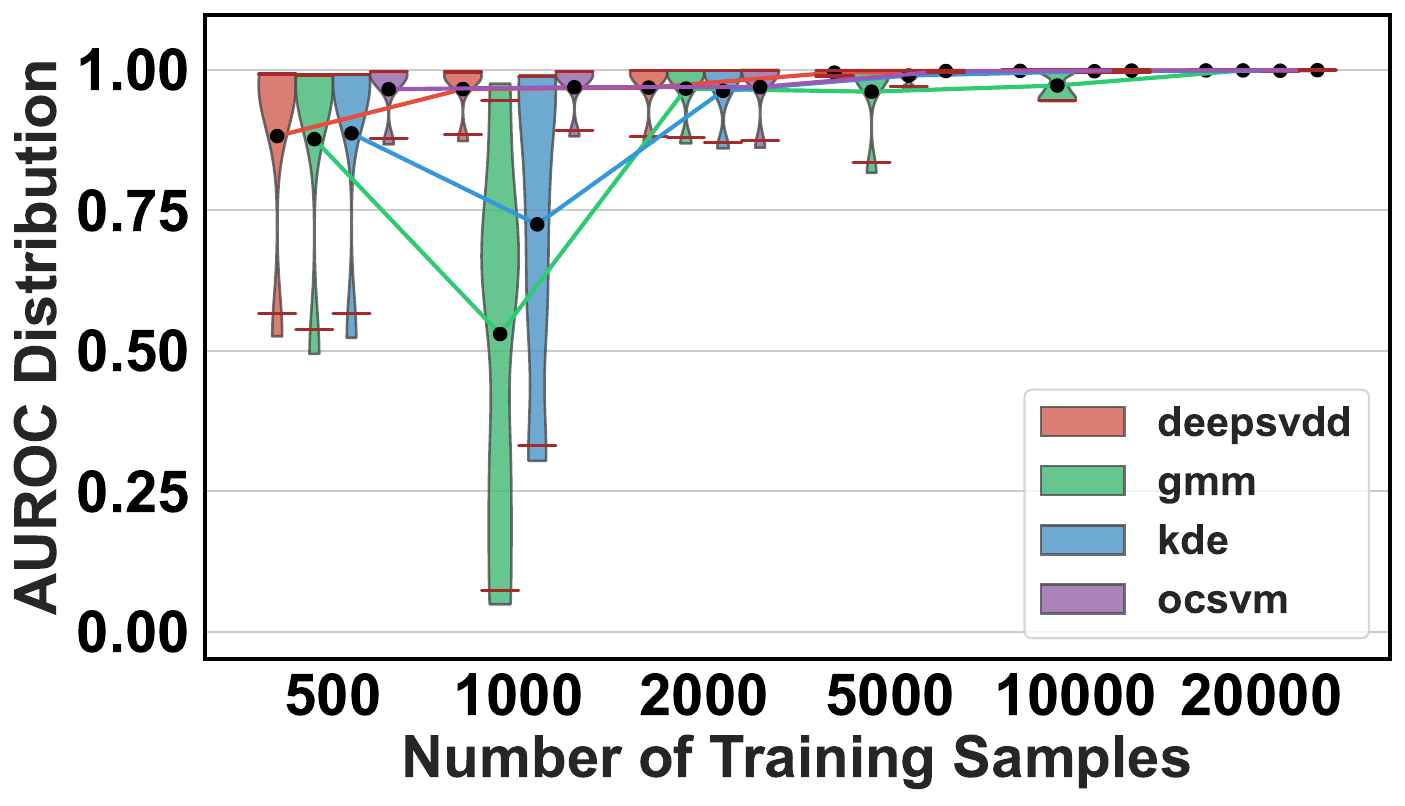}}
    \subfloat[BGL - Word2Vec]{\includegraphics[width=0.24\textwidth]{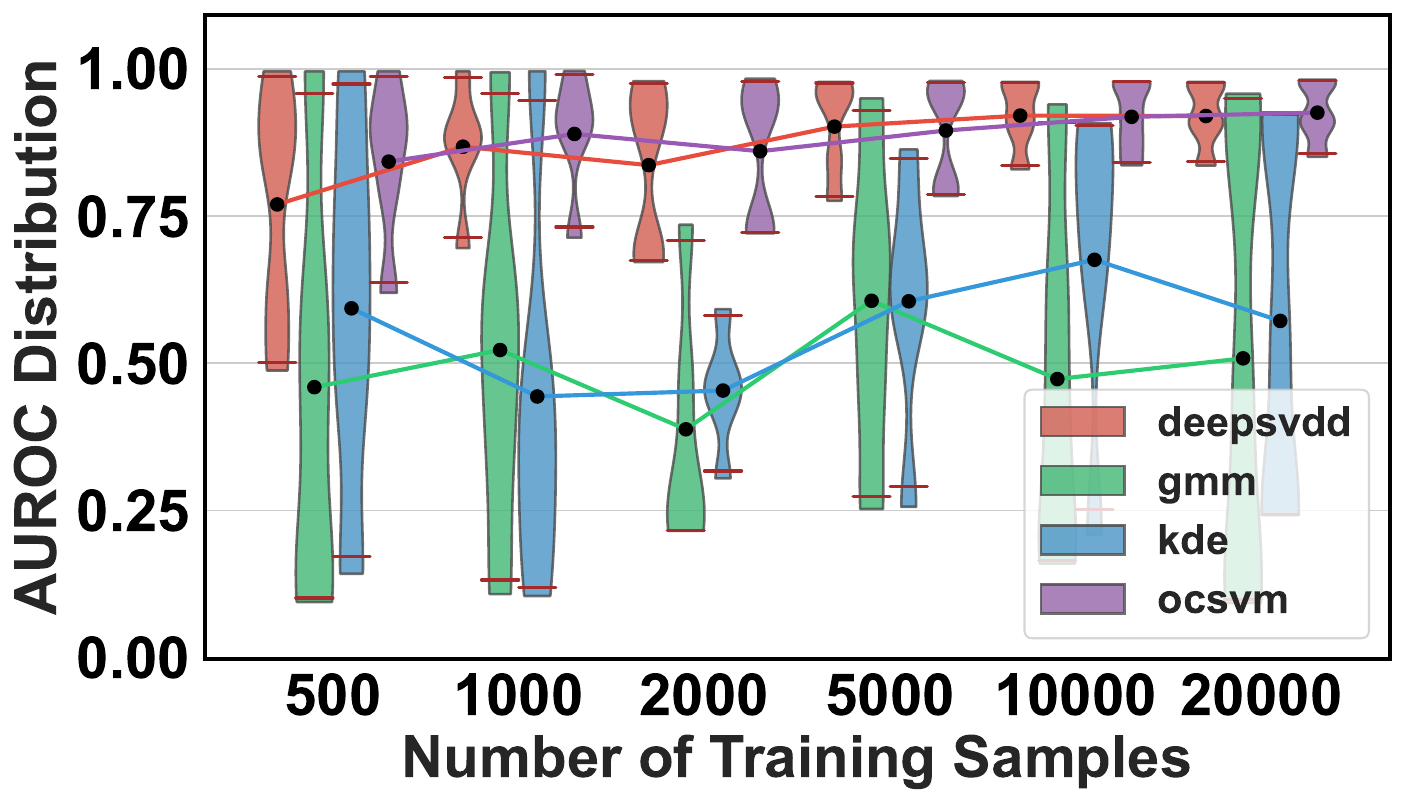}}
    \subfloat[BGL - SBERT]{\includegraphics[width=0.24\textwidth]{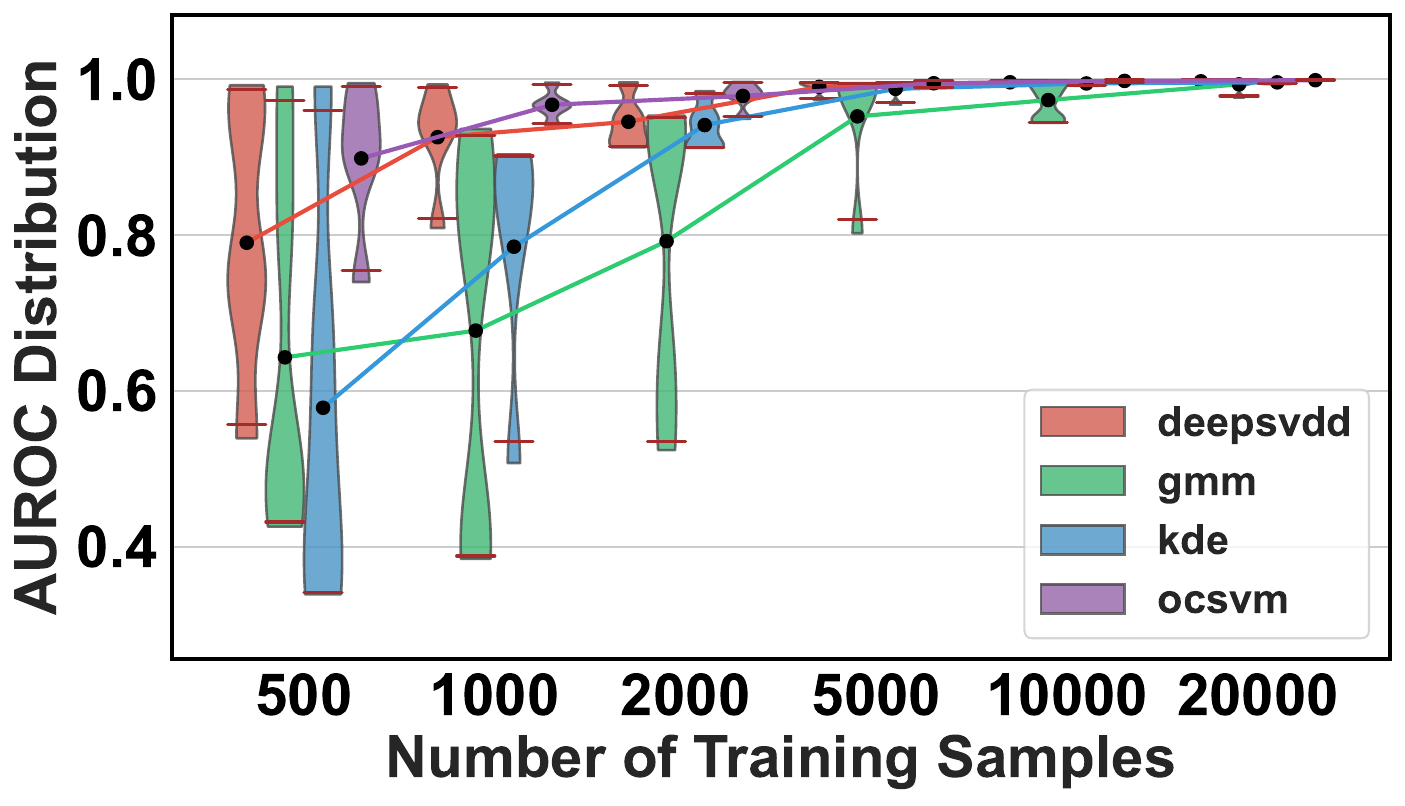}}
    \subfloat[BGL - Qwen-8B]{\includegraphics[width=0.24\textwidth]{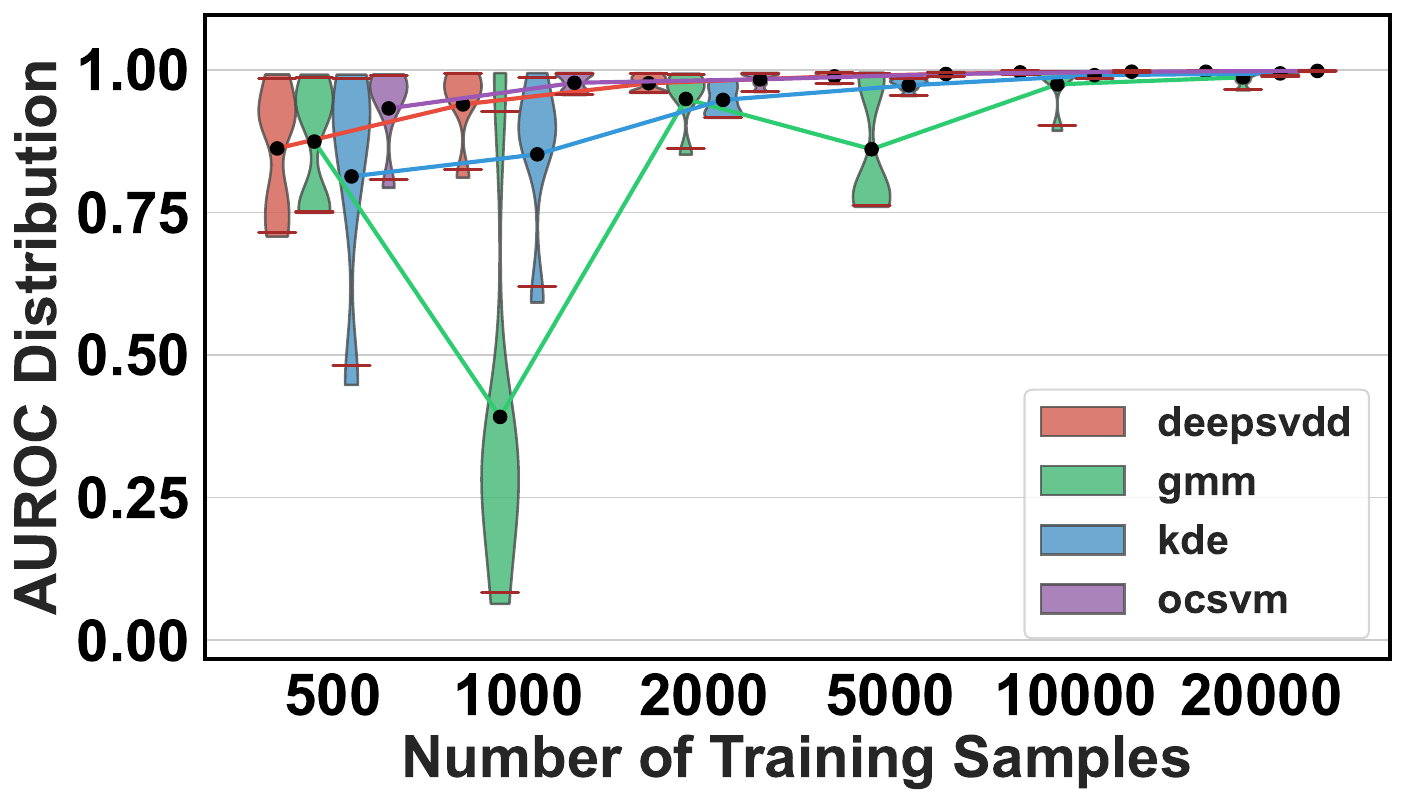}} \\
    \subfloat[Thunderbird - TF-IDF]{\includegraphics[width=0.24\textwidth]{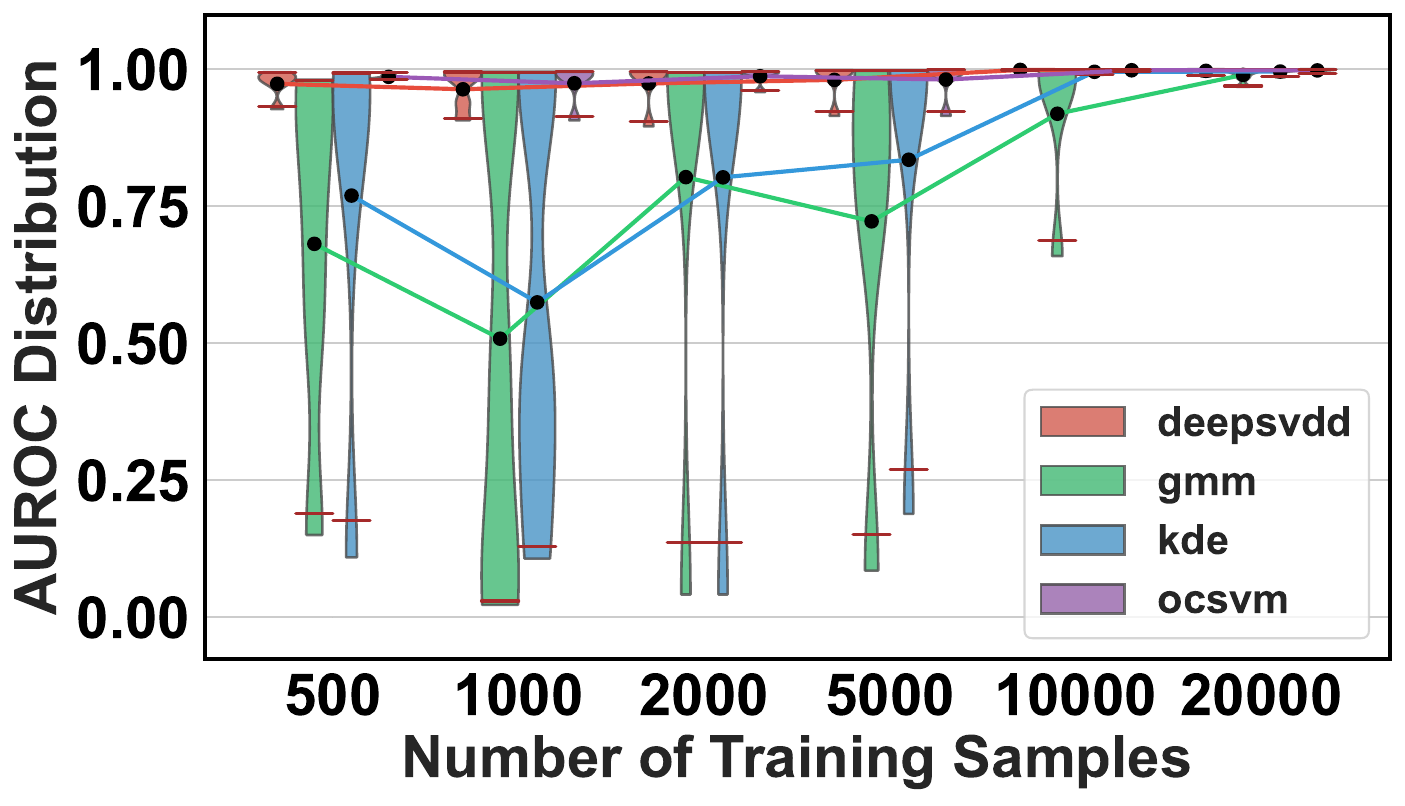}}
    \subfloat[Thunderbird - Word2Vec]{\includegraphics[width=0.24\textwidth]{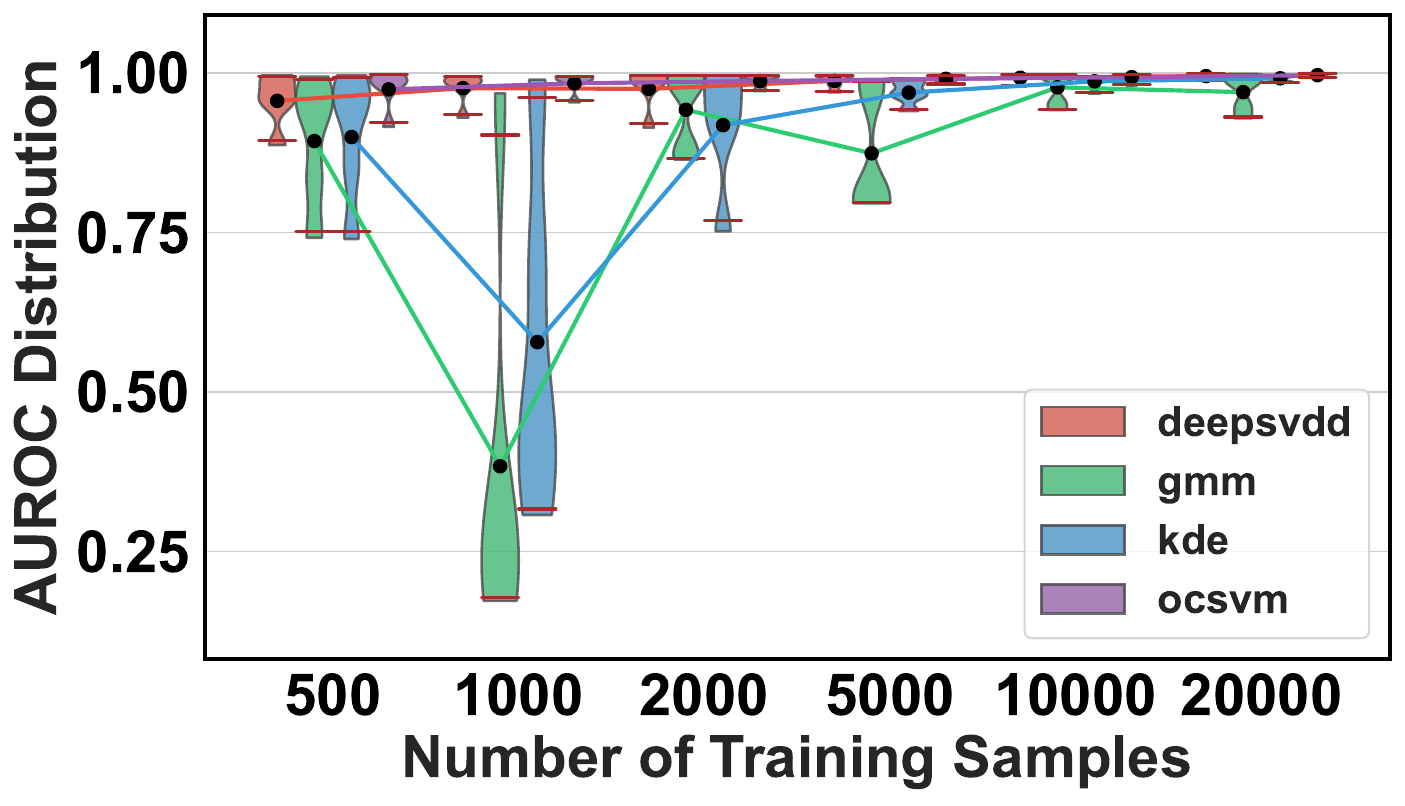}}
    \subfloat[Thunderbird - SBER]{\includegraphics[width=0.24\textwidth]{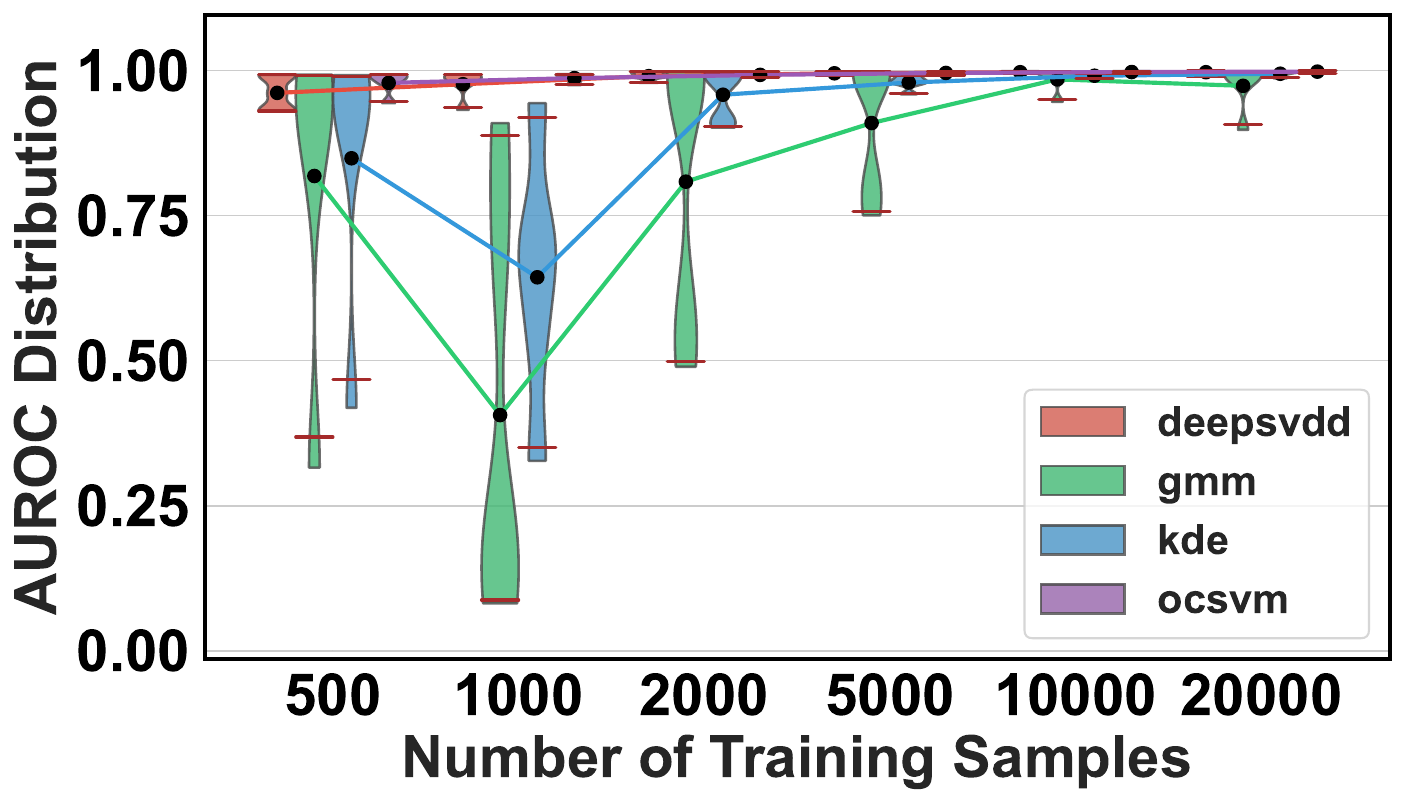}}
    \subfloat[Thunderbird - Qwen-8B]{\includegraphics[width=0.24\textwidth]{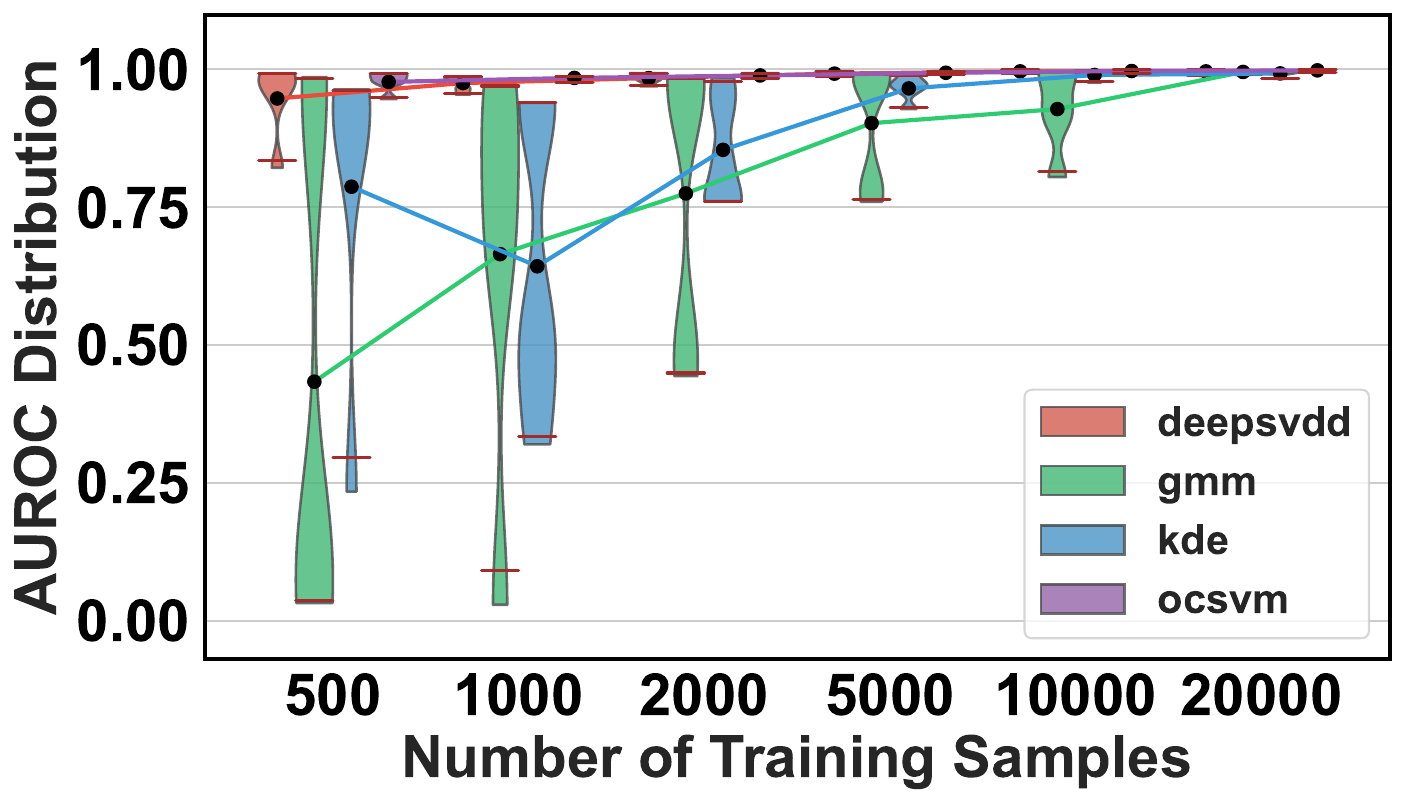}}
    \caption{AUROC scores for each detector as a function of \textbf{training sample size}, shown across datasets and embedding methods. All experiments use a window size of 320 and a stride of 5, which are parameters identified as \textbf{optimal} in Figure \ref{fig:stride_results} and Figure \ref{fig:window_size_results} and \textbf{reflective of real-world LogAD deployment}. Results for Qwen3-embedding-4B and Qwen3-embedding-0.6B embeddings are omitted due to their performance parity with the 8B model.}
    \label{fig:num_train_samples_results}
\end{figure*}

\vspace{1ex}
\noindent\textbf{Training Sample Size Analysis:}
Fig.~\ref{fig:num_train_samples_results} illustrates how AUROC varies with training sample size, ranging from $500$ to $20,000$, under the most effective configuration ($S$=5, $W$=320). We also exclude GMM and KDE numbers from the text (although we still plot them in the figure for completeness), as they underperform compared to DeepSVDD and OCSVM.

TF-IDF and Word2Vec both benefit significantly from additional training data. On \textsc{BGL}, TF-IDF consistently yields near-perfect AUROC at $10,000$ samples and beyond, while Word2Vec exhibits steady improvement across the range, despite having lower absolute accuracy. On \textsc{HDFS}, TF-IDF achieves strong performance once training size exceeds $5,000$, while Word2Vec remains lower but follows a similar trend. On \textsc{Thunderbird}, both embeddings attain high accuracy at moderate data scales, with TF-IDF stabilizing above $0.99$ beyond $10,000$ samples.
SBERT surpasses $0.96$ AUROC on \textsc{HDFS} and approaches $0.998$ on \textsc{BGL} with $20,000$ samples, while also showing strong results on \textsc{Thunderbird}. Qwen performs similarly, attaining high AUROC values across all sample sizes, and converging rapidly beyond $5,000$ samples. 


\begin{figure}[htbp]
  \centering
  \begin{subfigure}[t]{0.32\columnwidth}
    \centering
    \includegraphics[width=\linewidth]{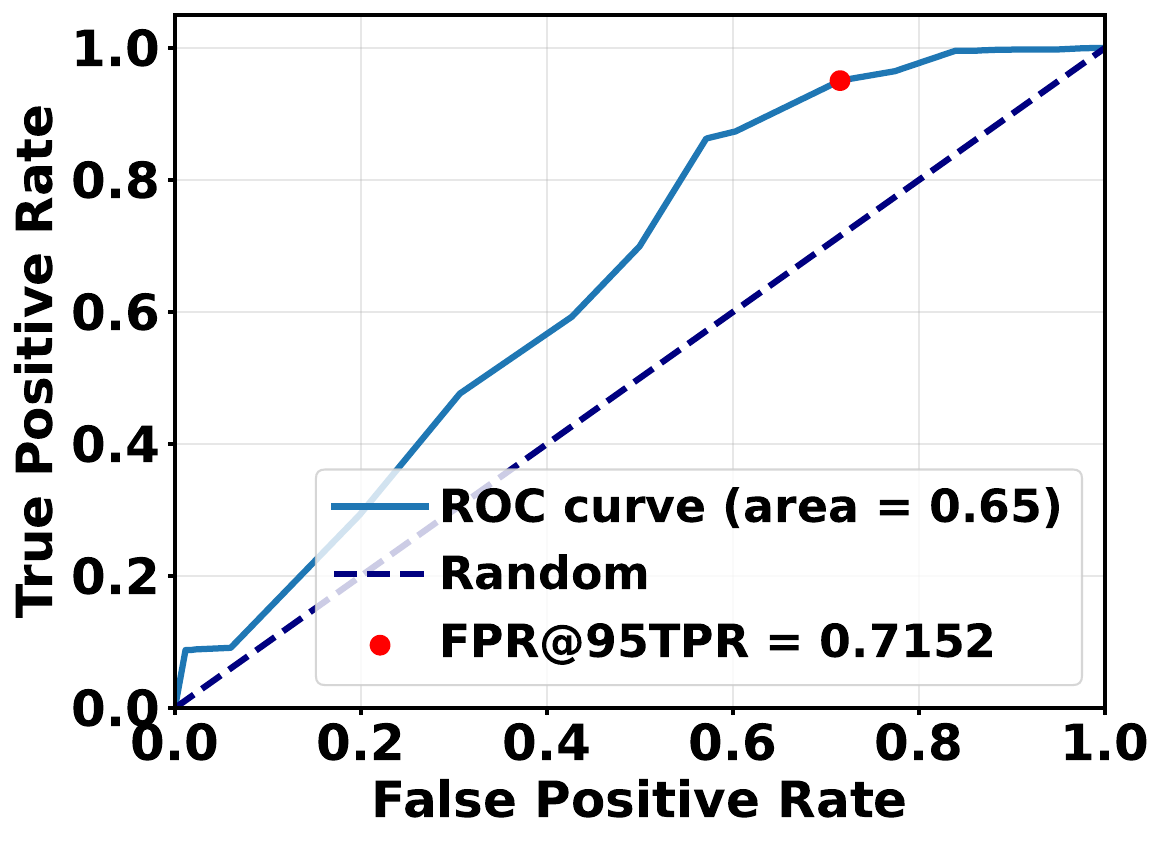}
    \subcaption{ROC curve @ 500 training samples}\label{fig:500-roc}
  \end{subfigure}\hfill
  \begin{subfigure}[t]{0.32\columnwidth}
    \centering
    \includegraphics[width=\linewidth]{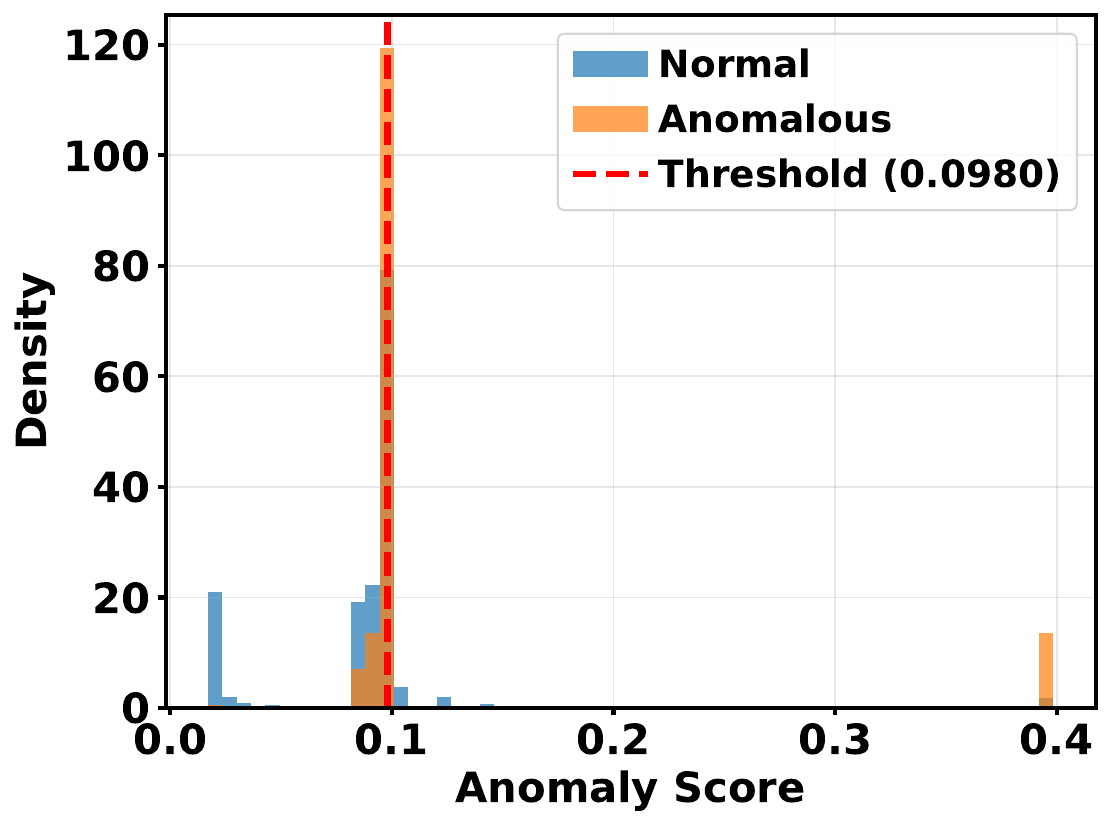}
    \subcaption{Score Distribution @ 500 training samples}\label{fig:500-score}
  \end{subfigure}\hfill
  \begin{subfigure}[t]{0.32\columnwidth}
    \centering
    \includegraphics[width=\linewidth]{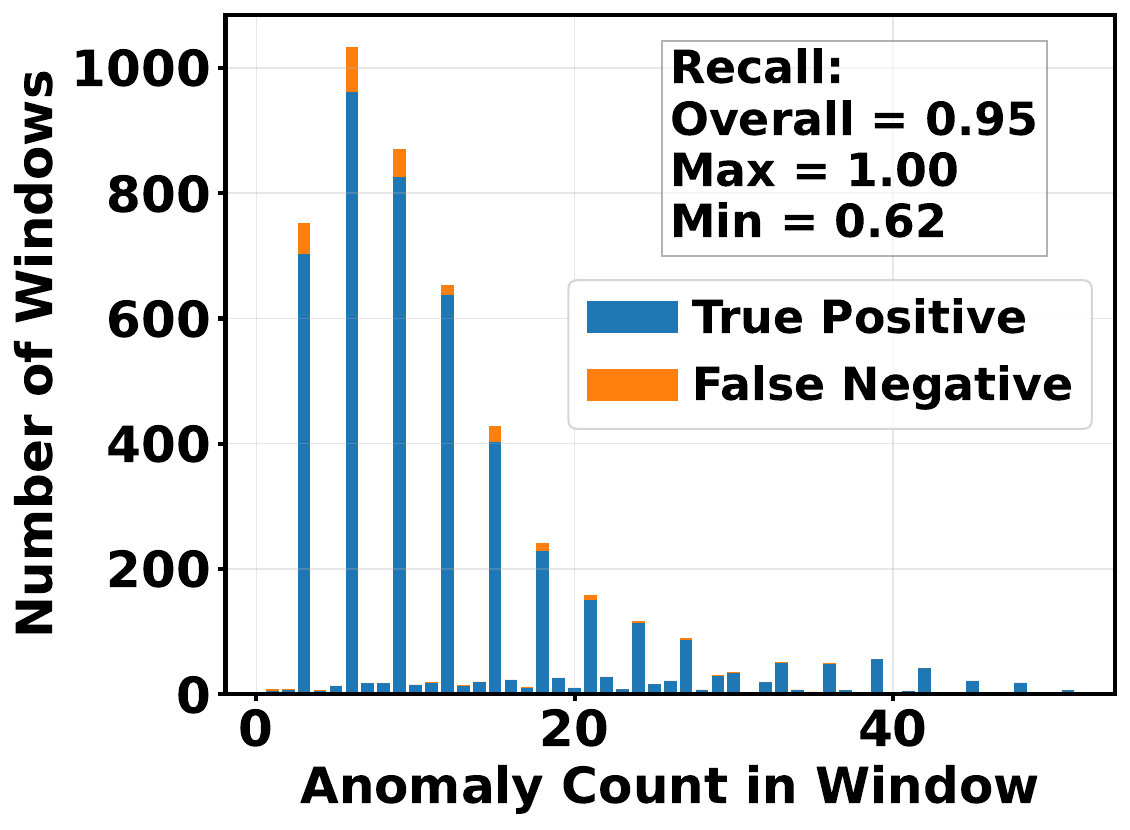}
    \subcaption{Anomaly Density \& Recall @ 500 samples}\label{fig:anom500}
  \end{subfigure}

  \begin{subfigure}[t]{0.32\columnwidth}
    \centering
    \includegraphics[width=\linewidth]{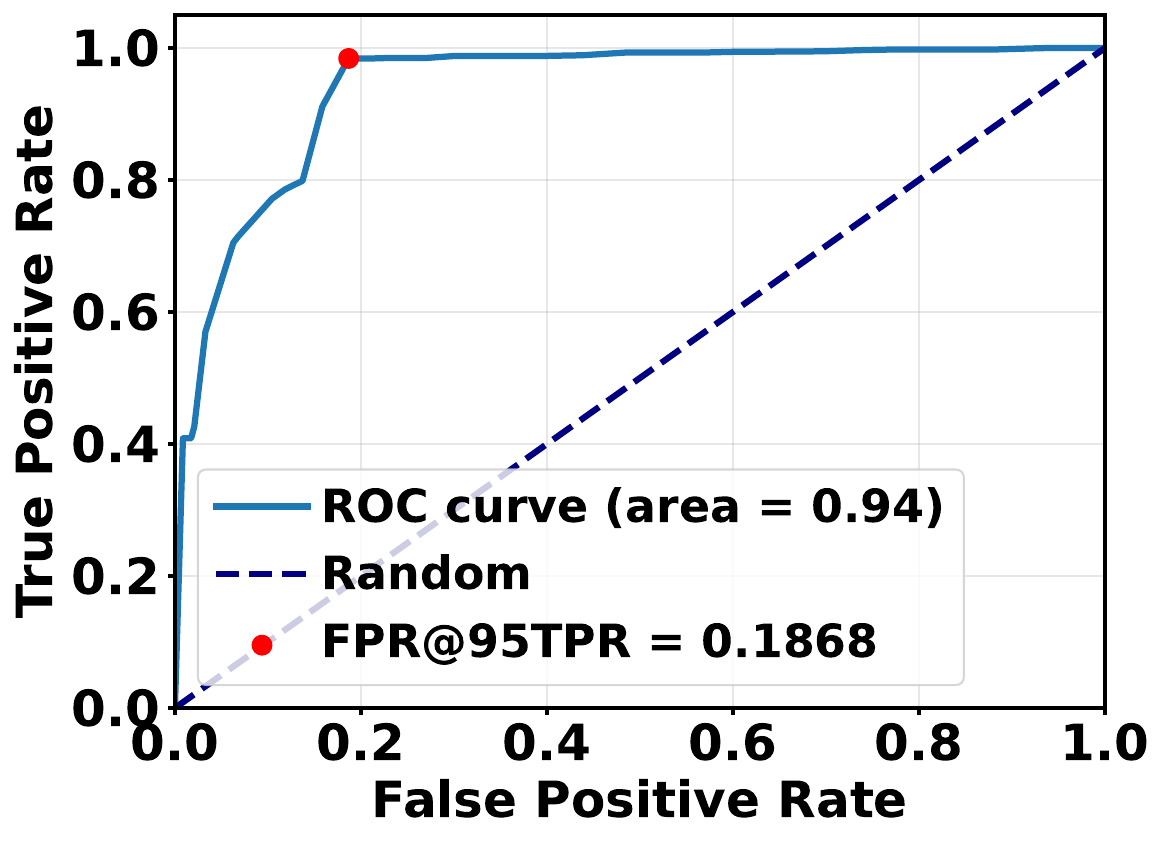}
    \subcaption{ROC curve @ 5K training samples}\label{fig:5,000-roc}
  \end{subfigure}\hfill
  \begin{subfigure}[t]{0.32\columnwidth}
    \centering
    \includegraphics[width=\linewidth]{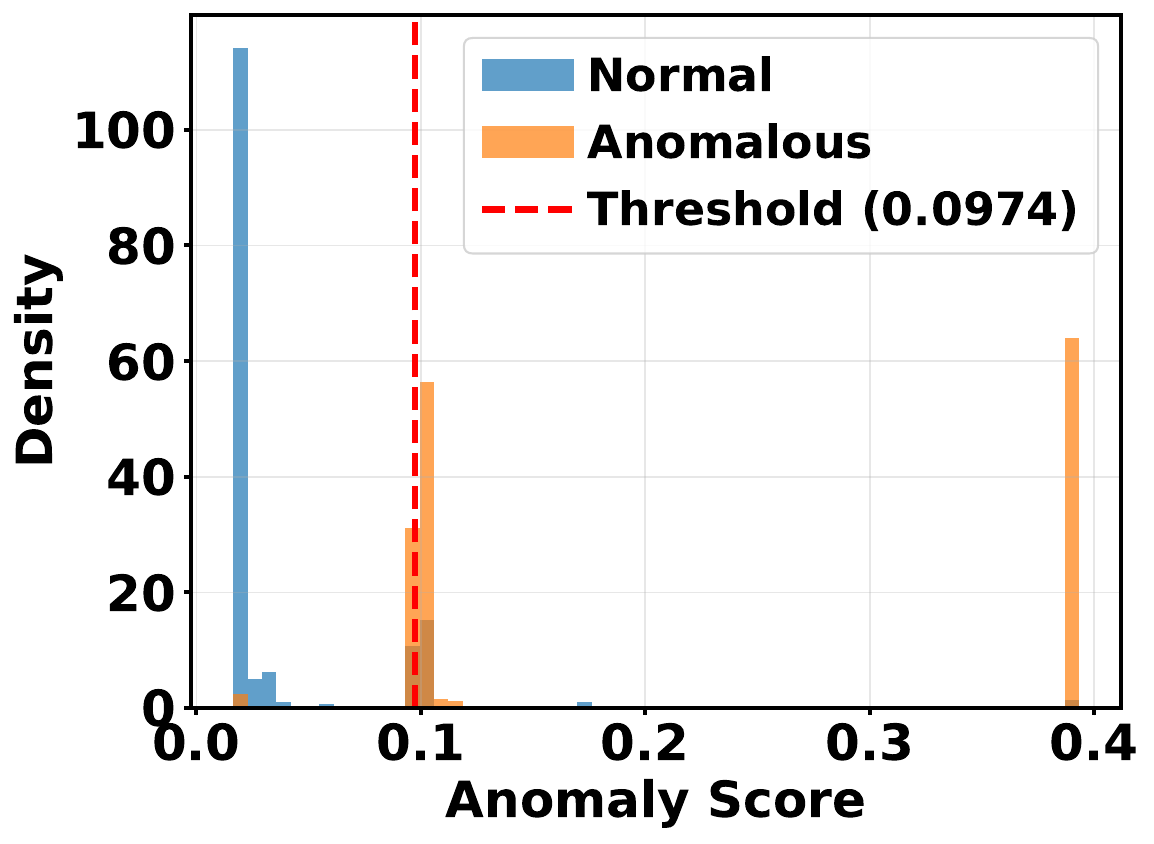}
    \subcaption{Score Distribution @ 5K training samples}\label{fig:5000-score}
  \end{subfigure}\hfill
  \begin{subfigure}[t]{0.32\columnwidth}
    \centering
    \includegraphics[width=\linewidth]{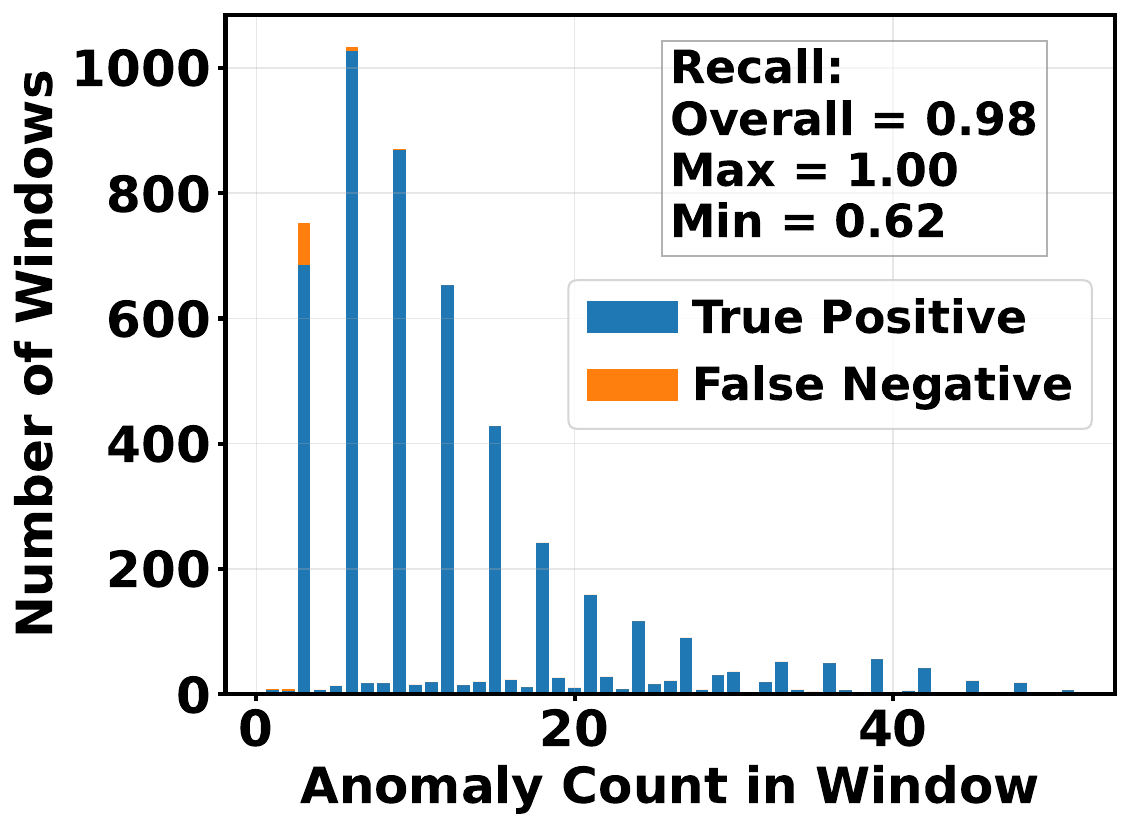}
    \subcaption{Anomaly Density \& Recall @ 5K samples}\label{fig:anom5000}
  \end{subfigure}

  \begin{subfigure}[t]{0.32\columnwidth}
    \centering
    \includegraphics[width=\linewidth]{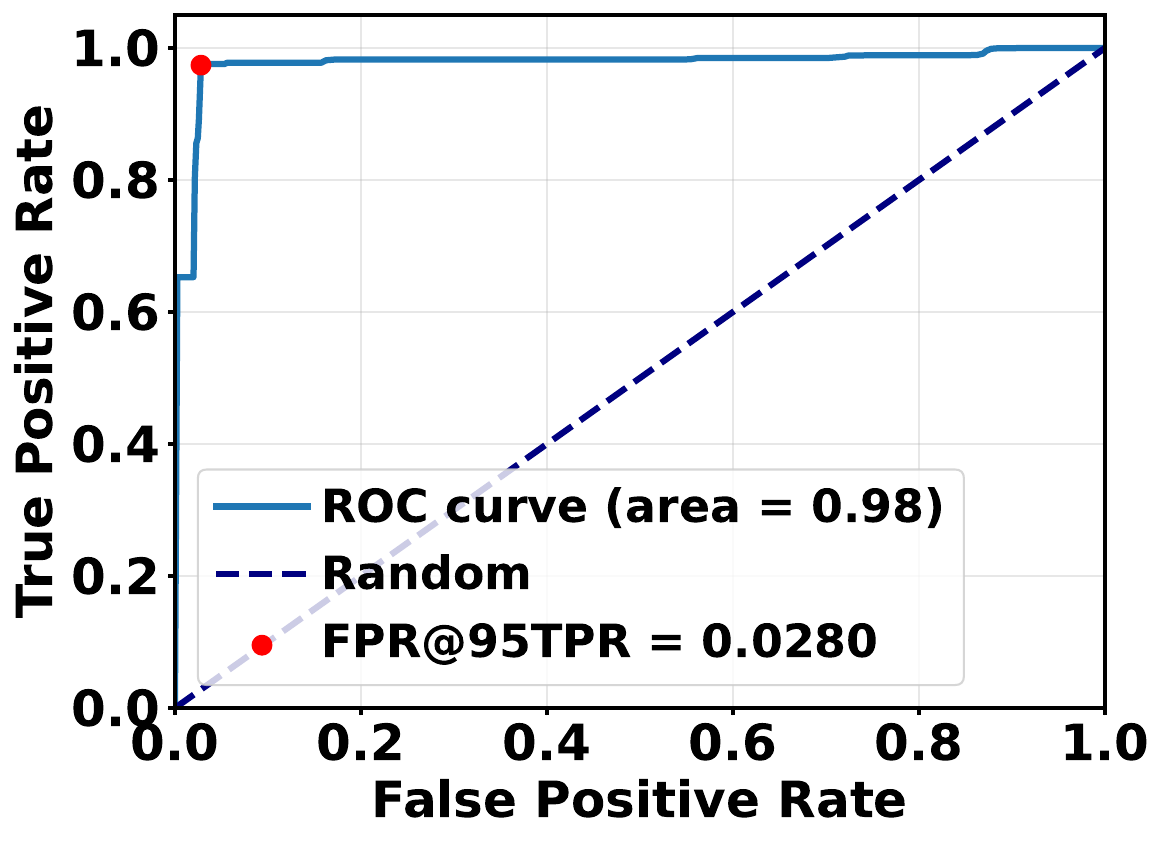}
    \subcaption{ROC curve @ 20K training samples}\label{fig:20000-roc}
  \end{subfigure}\hfill
  \begin{subfigure}[t]{0.32\columnwidth}
    \centering
    \includegraphics[width=\linewidth]{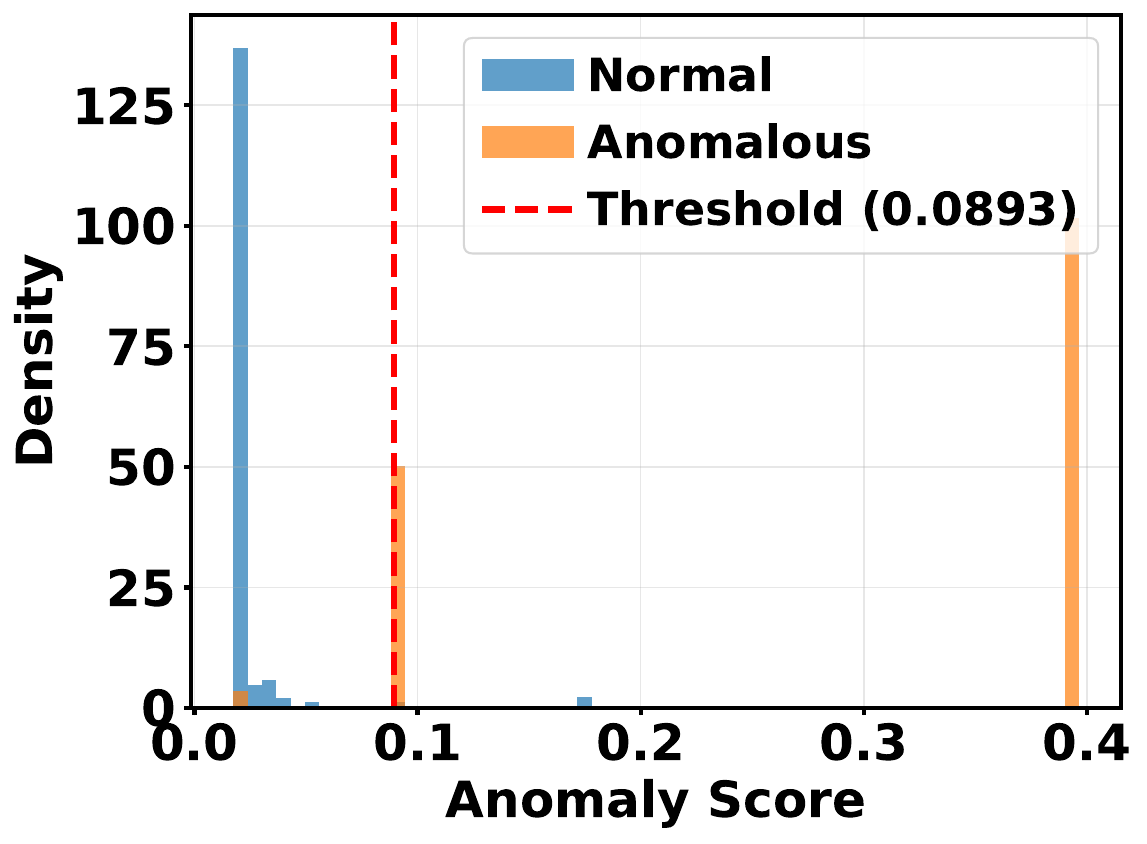}
    \subcaption{Score Distribution @ 20K training samples}\label{fig:20000-score}
  \end{subfigure}\hfill
  \begin{subfigure}[t]{0.32\columnwidth}
    \centering
    \includegraphics[width=\linewidth]{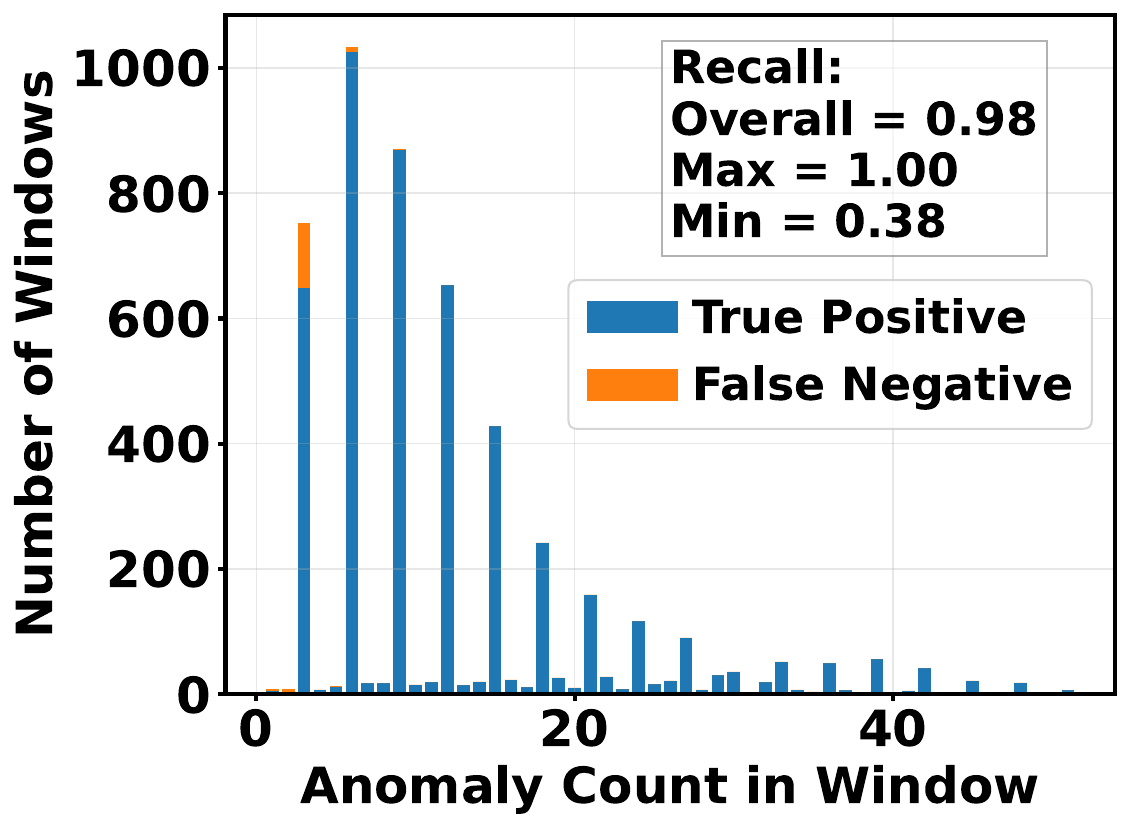}
    \subcaption{Anomaly Density \& Recall @ 20K samples}\label{fig:anom20000}
  \end{subfigure}

  \caption{Zoom-in analysis of an experiment set (HDFS chunk 3 + SBERT + DeepSVDD) visualizing the progression of ROC (left), score distribution (middle), and anomaly density (right) triplet with growing training sample size. \projname{} generates such visual diagnostics for every experiment, enabling nuanced understanding of model behavior across settings, thereby facilitating actionable insights for optimizing real-world deployments.}
  \label{fig:roc-score-anom}
\end{figure}

\vspace{1ex}
\noindent\textbf{Visual Diagnostic:}
Fig.~\ref{fig:roc-score-anom} provides a detailed progression of \projname{}'s detection behavior as training data increases. First, the ROC curves clearly indicate improved discrimination with more data: AUROC rises from $0.651$ at $500$ samples to $0.941$ at $5,000$ and $0.980$ at $20,000$. Correspondingly, FPR@95TPR drops significantly from $0.715$ to $0.187$ to $0.028$, highlighting reduced false alarms at high recall thresholds.
Second, score distributions sharpen with training size. At $500$ samples, normal and anomalous scores overlap heavily around the decision threshold (e.g., $0.098$), making separation difficult. As sample size increases, distributions diverge: by $20,000$ samples, anomalous scores dominate above the threshold ($0.089$), yielding much cleaner separation and more confident detection.
Third, anomaly density analysis (i.e., the number of anomaly logs within a window) reveals that model recall improves with denser anomalies. From $500$ to $5,000$ samples, recall increases markedly (from $0.951$ to $0.984$), especially in windows with moderate anomaly counts. Beyond $5,000$, recall gains taper, but false positives continue to decline from $0.187$ to $0.028$, indicating higher-quality precision-recall tradeoffs.

Importantly, false negatives (FN) persist primarily in windows with very low anomaly density. This reflects a classic ``dilution effect'', i.e., sparse anomalies are harder to detect because their signals are overwhelmed by normal logs. Such subtle anomalies are often the most dangerous (e.g., indicative of stealthy system breaches or early-stage faults), so a more permissive threshold may be justified to ensure such rare but critical anomalies are not missed.

\begin{figure}[htbp]
  \centering
  \begin{subfigure}{0.32\columnwidth}
    \includegraphics[width=\linewidth]{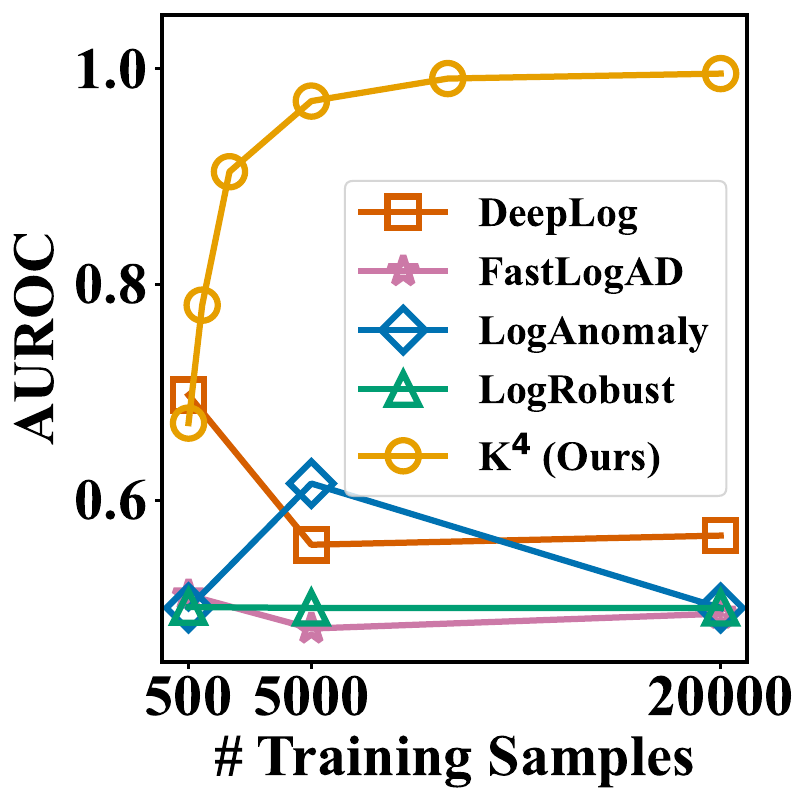}
    \subcaption{HDFS}
    \label{fig:auroc-train-hdfs}
  \end{subfigure}
  \hfill
  \begin{subfigure}{0.32\columnwidth}
    \includegraphics[width=\linewidth]{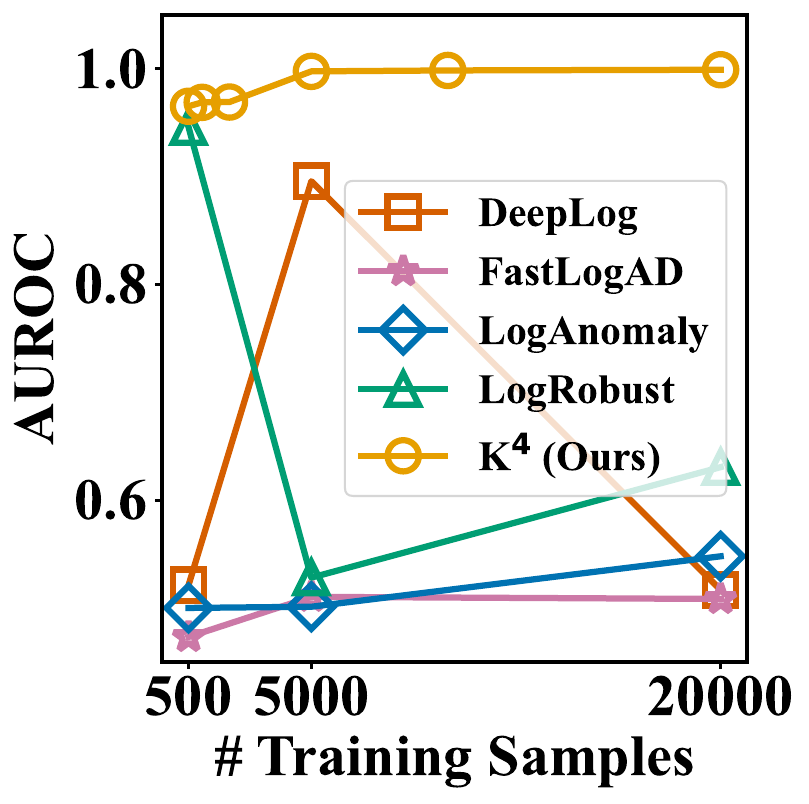}
    \subcaption{BGL}
    \label{fig:auroc-train-bgl}
  \end{subfigure}
  \hfill
  \begin{subfigure}{0.32\columnwidth}
    \includegraphics[width=\linewidth]{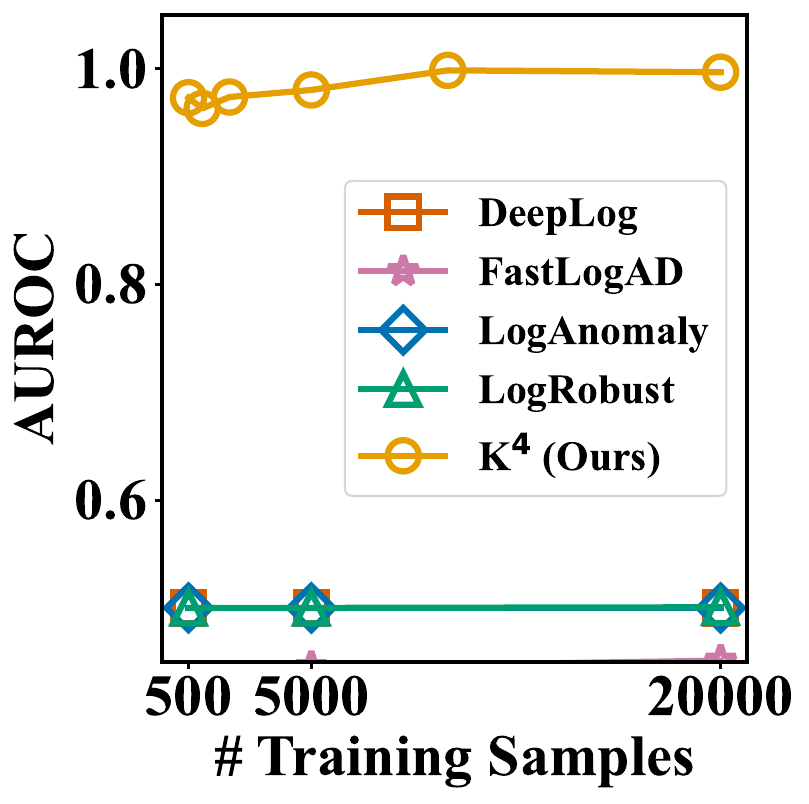}
    \subcaption{Thunderbird}
    \label{fig:auroc-train-tb}
  \end{subfigure}
  \caption{Comparison between \projname{} and baselines in AUROC vs. training sample size across three datasets, averaged over all chunks. All experiments use a window size of 320 and a stride of 5. \projname{} consistently outperforms all baselines. \textbf{Notably, in most cases, baselines perform near random (AUROC $\approx$ 0.5), highlighting significant limitations in their applicability to real-world deployments.}}
  \label{fig:auroc-train-comparison}
\end{figure}

\vspace{1ex}
\noindent\textbf{Comparison with Baselines:}
Fig.\ref{fig:auroc-train-comparison} and Table\ref{tab:detection-comparison} compare \projname{} against state-of-the-art unsupervised LogAD baselines across AUROC, AUPRC, FPR@95TRP, F1, Precision, and Recall. All studies are evaluated across varying training sample sizes (500, 5,000, 20,000) using a consistent window size of 320 and stride of 5. For \projname{}, the highest average AUROC across all chunks for each dataset is: 20,000 training samples + TF-IDF + OCSVM for HDFS and BGL, and 10,000 training samples + TF-IDF + DeepSVDD for Thunderbird.

Across all datasets and metrics, \projname{} achieves substantial performance gains. For AUROC, it reaches near-perfect scores of $0.996$ (HDFS), $0.999$ (BGL), and $0.999$ (Thunderbird), significantly outperforming all baselines. In contrast, DeepLog achieves only $0.697$ (HDFS) and $0.5$ (Thunderbird), while LogAnomaly and LogRobust display unstable performance, with AUROC often near or at chance level (e.g., $0.501$ on HDFS for LogRobust). Most baselines show AUROC values around $0.5$ on Thunderbird, highlighting limited generalization. 

\projname{} also achieves high AUPRC (up to $0.999$) and drastically lower FPR@95TRP ($<0.8\%$), indicating reliable detection with minimal false alarms. For example, on BGL, \projname{} reports nil FPR@95TRP, versus $1.0$ for LogAnomaly and $0.904$ for DeepLog. In terms of F1, Precision, and Recall, our method consistently achieves high scores across all datasets (e.g., Thunderbird F1=$0.992$, Precision=$0.987$, Recall=$0.997$), while baselines either collapse (e.g., LogRobust F1=$0.002$) or produce unstable outputs.

LLM-AD~\cite{jin2024large} and Ladle~\cite{myllari2025ladle} are omitted from Figure~\ref{fig:auroc-train-comparison} as they do not trivially support sample-dependent input (i.e., they inherently treat each log line as input). Therefore, their sole per-dataset results are reported in Table~\ref{tab:detection-comparison}.

\begin{table*}[htbp]
\centering
\caption{Detection performance of \projname{} and baselines on HDFS, BGL, and Thunderbird (TB), using their best-performing configurations (highest AUROC). \projname{}'s per-dataset settings are detailed in the text; baseline settings correspond to those shown in Figure~\ref{fig:auroc-train-comparison}.}
\label{tab:detection-comparison}
\resizebox{\textwidth}{!}{
\begin{tabular}{|l|ccc|ccc|ccc|ccc|ccc|ccc|}
\hline
\multirow{2}{*}{\textbf{Method}} 
& \multicolumn{3}{c|}{\textbf{AUROC}} 
& \multicolumn{3}{c|}{\textbf{AUPRC}} 
& \multicolumn{3}{c|}{\textbf{FPR@95TRP}} 
& \multicolumn{3}{c|}{\textbf{F1}} 
& \multicolumn{3}{c|}{\textbf{Precision}} 
& \multicolumn{3}{c|}{\textbf{Recall}} \\
& HDFS & BGL & TB & HDFS & BGL & TB & HDFS & BGL & TB & HDFS & BGL & TB & HDFS & BGL & TB & HDFS & BGL & TB \\
\hline
DeepLog~\cite{du2017deeplog}      & 0.697 & 0.896 & 0.500 & 0.506 & 0.899 & 0.508 & 0.568 & 0.208 & 1.000 & 0.093 & 0.888 & 0.031 & 0.049 & 0.799 & 0.016 & 0.963 & 1.000 & 1.000 \\
LogAnomaly~\cite{meng2019loganomaly}   & 0.615 & 0.548 & 0.500 & 0.460 & 0.738 & 0.508 & 1.000 & 0.904 & 1.000 & 0.347 & 0.645 & 0.032 & 0.664 & 0.476 & 0.017 & 0.235 & 1.000 & 1.000 \\
LogRobust~\cite{zhang2019robust}    & 0.501 & 0.946 & 0.501 & 0.612 & 0.983 & 0.509 & 0.997 & 1.000 & 1.000 & 0.367 & 0.943 & 0.002 & 0.225 & 0.999 & 1.000 & 0.999 & 0.892 & 0.001 \\
FastLogAD~\cite{lin2024fastlog}    & 0.512 & 0.510 & 0.451 & 0.031 & 0.192 & 0.020 & 0.947 & 0.945 & 0.958 & 0.059 & 0.277 & 0.049 & 0.031 & 0.188 & 0.110 & 0.771 & 0.567 & 0.031 \\
LLM-AD~\cite{jin2024large}       & 0.500 & 0.341 & 0.500 & 0.868 & 0.377 & 0.033 & 1.000 & 1.000 & 1.000 & 0.929 & 0.471 & 0.064 & 0.868 & 0.360 & 0.033 & 1.000 & 0.683 & 1.000 \\
Ladle~\cite{myllari2025ladle}        & 0.440 & 0.648 & 0.267 & 0.837 & 0.321 & 0.257 & 0.992 & 0.975 & 0.732 & 0.146 & 0.222 & 0.170 & 0.167 & 0.088 & 0.100 & 0.428 & 0.857  & 0.560 \\
\textbf{\projname{} (Ours)}  & \textbf{0.996} & \textbf{0.999} & \textbf{0.999} & \textbf{0.995} & \textbf{0.999} & \textbf{0.998} & \textbf{0.008} & \textbf{0.000} & \textbf{0.005} & \textbf{0.989} & \textbf{0.992} & \textbf{0.992} & \textbf{0.989} & \textbf{0.993} & \textbf{0.987} & \textbf{0.990} & \textbf{0.992} & \textbf{0.997} \\
\hline
\end{tabular}
}
\end{table*}

\subsection{Runtime Peformance Evaluation}
\label{ss:runtime_performance_evaluation}

\newcolumntype{C}[1]{>{\centering\arraybackslash}p{#1}}

\begin{table}[htbp]
\centering
\caption{\projname{}'s runtime breakdown for embedding and detector stages. Embedding times reflect cold-start generation for 20K training samples and per-sample inference latency for a single test window. Detector times correspond to training over 20K embeddings and per-embedding inference using cached embeddings.}
\label{tab:runtime-compact}
\renewcommand{\arraystretch}{1.2}

\begin{minipage}[b]{0.49\columnwidth}
\centering
\textbf{\projname{}'s Embedding Runtime (s)}\\[1ex]
\begin{tabular}{@{}lcc@{}}
\toprule
\textbf{Embedding} & \textbf{Train} & \textbf{Inference} \\
\midrule
TFIDF    & 162     & $8.0\!\times\!10^{-3}$ \\
Word2Vec & 299     & $1.5\!\times\!10^{-2}$ \\
SBERT    & 1,536    & $7.7\!\times\!10^{-2}$ \\
Qwen-8B     & 22,127   & 1.1 \\
\bottomrule
\end{tabular}
\end{minipage}
\hfill
\begin{minipage}[b]{0.49\columnwidth}
\centering
\textbf{\projname{}'s Detector Runtime (s)}\\[1ex]
\begin{tabular}{@{}lcc@{}}
\toprule
\textbf{Detector} & \textbf{Train} & \textbf{Inference} \\
\midrule
DSVDD   & 1.44  & $4.4\!\times\!10^{-6}$ \\
GMM     & 1.86  & $7.0\!\times\!10^{-6}$ \\
KDE     & 0.05  & $8.7\!\times\!10^{-6}$ \\
OCSVM   & 3.83  & $4.4\!\times\!10^{-6}$ \\
\bottomrule
\end{tabular}
\end{minipage}
\end{table}

\begin{table}[htbp]
\centering
\caption{Runtime performance of baselines on HDFS, BGL, and Thunderbird (TB). Training time is for 20K samples. Inference time reflects end-to-end latency per test sample.}
\label{tab:runtime}
\renewcommand{\arraystretch}{1.2}
\begin{tabular}{@{}lcccccc@{}}
\toprule
\multirow{2}{*}{\textbf{Method}} 
& \multicolumn{3}{c}{\textbf{Train Time (s)}} 
& \multicolumn{3}{c}{\textbf{Inference Time (ms)}} \\
\cmidrule(lr){2-4} \cmidrule(lr){5-7}
& HDFS & BGL & TB & HDFS & BGL & TB \\
\midrule
DeepLog~\cite{du2017deeplog}         & 21,528 & 15,228 & 13,212 & 0.77 & 0.23 & 3.20 \\
LogAnomaly~\cite{meng2019loganomaly} & 19,332 & 19,152 & 21,816 & 6.47 & 3.30 & 8.92 \\
LogRobust~\cite{zhang2019robust}     & 18,792 & 15,732 & 19,368 & 5.84 & 3.22 & 7.12 \\
FastLogAD~\cite{lin2024fastlog}      & 10,583     & 33,998     & 212,802     & 43.9   & 53.9   & 16,229 \\
LLM-AD~\cite{jin2024large}           & 10,440 & 12,240 & 18,360 & 182  & 216  & 182 \\
Ladle~\cite{myllari2025ladle}        & 1,520     & 1,587     & 1,633     &  72.3 & 81.5   & 83.8  \\
\bottomrule
\end{tabular}
\end{table}

Table~\ref{tab:runtime-compact} breaks down \projname{}’s embedding and detector runtimes. For embedding, TF-IDF and Word2Vec are highly efficient, requiring only $162$ and $299$ seconds to generate 20,000 training embeddings, and under $0.02$ ms per test embedding. SBERT is more expensive at $1536$ seconds and $0.077$ seconds. Qwen-8B is the slowest, taking over $6$ hours and $1.1$ seconds. Notably, these training times reflect the worst-case costs of cold-starting (i.e., no historical embeddings are available). Since \projname{} uses pre-trained-only embedding models, previously seen samples do not need to be re-embedded. Moreover, for streaming deployment, a new log window is embedded once and reused for both inference and future training if the new window is classified as normal, which enables efficient continual learning without repeated embedding overhead. Importantly, \ul{TF-IDF not only offers the most efficient runtime profile, but also delivers the highest detection accuracy across datasets} (see Table~\ref{tab:detection-comparison}), making it the ideal choice in this study.

Detectors run efficiently: KDE is the fastest in training with $0.05$ seconds and slowest in inference with $8.7$\,$\mu$s. DeepSVDD and OCSVM are $1.44$ and $3.83$ seconds for training, with inference latency under $5$\,$\mu$s. GMM takes both times the middle ground.

Table~\ref{tab:runtime} compares baseline frameworks. DeepLog, LogAnomaly, and LogRobust each require over $3$-$6$ hours to train on 20,000 samples, and inference latency ranges from $0.23$ ms (BGL, DeepLog) to $8.92$ ms (Thunderbird, LogAnomaly). LLM-AD, despite avoiding windowing, still incurs significant runtime: supervised fine-tuning takes $2.9$-$5.1$ hours (per dataset), and each test sample inference takes over $180$ ms. \ul{All baselines are orders of magnitude slower than \projname{} in training and inference speed.} 

\section{Discussion}
As discussed in ``Gaps and Needs'' of Section~\ref{sec:introduction}, deploying LogAD in real-world online scenarios requires well-defined operational strategies to ensure detection effectiveness and robustness. Here, we outline key considerations and recommended practices to maximize the practical usability and reliability of \projname{} in production environments.

\subsection{How to ensure clean normal logs for training?}

In real-world online LogAD scenarios, logs arrive as continuous streams without labels. This makes it essential to avoid contaminating training data with anomalous messages, as doing so can distort the estimation of the ``normal’’ manifold which \projname{} relies on. A conservative strategy for selecting training samples is therefore critical.

One practical approach is keyword-based filtering. Many production systems and mature software platforms (e.g., operating systems, databases, or web servers) adopt standardized logging conventions with severity annotations. Excluding log entries containing keywords such as \texttt{error}, \texttt{fatal}, or even \texttt{warning} (which may also precede or accompany anomalous behavior) can substantially reduce the risk of including anomalies in training. To further increase robustness, surrounding log entries within a tunable time window or sequence neighborhood of flagged messages can also be conservatively excluded.
        
Another strategy is to include a small-scale manual verification step during the initial deployment. In high-stakes applications, investing minimal human effort to curate a few hundred clean log sequences can be highly beneficial. As demonstrated in Figure~\ref{fig:num_train_samples_results}, \projname{} achieves strong performance with as few as 500 training samples. Given that many production systems exhibit stable log patterns over time, this one-time labeling effort can remain valid across long periods, making it a practical and cost-effective solution.

These strategies can also be combined hierarchically, such as using keyword filtering as a first-pass filter, then human verification for high-assurance deployment scenarios. 

\subsection{How to set threshold without ground truth?}

In the absence of labeled test data, setting an appropriate anomaly threshold is inherently challenging for any score-based LogAD method, as scores must be binarized to trigger decisions. Many prior methods sidestep this by tuning thresholds on held-out labels~\cite{le2022log, guo2021logbert, meng2019loganomaly, zhang2019robust}, which is an unrealistic luxury in real deployments. 

The straightforward solution is to classify a test sample as anomalous if its score  $s_i$ exceeds a fixed percentile (e.g., the $95^\text{th}$) of the training score distribution. 
An alternative is parametric thresholding, which assumes the anomaly score follows a known distribution, allowing thresholds to be selected based on analytically derived tail probabilities. In practice, since anomalies typically reside in sparse, low-density regions, the shape of the score distribution depends on the underlying detection model. Density estimators (e.g., GMM, KDE) often yield scores resembling exponential or Gamma-like distributions, reflecting the sharp drop in likelihoods for outliers. One-class classifiers (e.g., OCSVM, DeepSVDD) tend to produce score distributions with a sharp boundary near zero (as illustrated in Figure~\ref{fig:5000-score} and~\ref{fig:20000-score}) and a long right tail. Therefore, understanding the empirical score distribution induced by the chosen model can inform the selection of a more robust and principled thresholding strategy.

Ultimately, the best thresholding strategy depends on the log dataset's inherent attribute (such as the rarity of anomalies) and the operator's emphasis on precision vs. recall, among other factors. \projname{} can be tailored on a case-by-case basis for all these strategies.
\section{Conclusion}
We present \projname{} (Knowing the Unkown by Knowing only the Known), a novel log anomaly detection framework that operates directly on raw log sequences without requiring log parsing, supervised labels, or retraining of embedding models. By decoupling embedding from detection and leveraging compact representation-level PRDC statistics, \projname{} supports limitless pretrained embeddings and offers four efficient detector backends. 
Our extensive evaluation under a realistic online LogAD setting demonstrates that \projname{} consistently delivers near-perfect detection performance while achieving detection latency as low as 4\,$\mu$s, which is several orders of magnitude faster than existing methods. These results underscore the practical viability of \projname{} as a high-performance, modular, and robust solution for modern log anomaly detection workloads.


\bibliographystyle{IEEEtran}
\bibliography{references}

\end{document}